\documentclass{article}

\usepackage[nonatbib,preprint]{neurips_2025}
\usepackage{shortex} 
\usepackage[utf8]{inputenc} % allow utf-8 input
\usepackage[T1]{fontenc}    % use 8-bit T1 fonts
\usepackage{hyperref}       % hyperlinks
\usepackage{url}            % simple URL typesetting
\usepackage{booktabs}       % professional-quality tables
\usepackage{amsfonts}       % blackboard math symbols
\usepackage{nicefrac}       % compact symbols for 1/2, etc.
\usepackage{microtype}      % microtypography
\usepackage{xcolor}         % colors
\usepackage{algorithm}
\usepackage{algpseudocode}
\usepackage{pifont}
\usepackage{accents}
\usepackage{bibentry}
\nobibliography* % allows us to do inline citations in supplement without changing bibliography
\usepackage{enumitem}

\newcommand{\anonymize}[1]{}
\newcommand{\anonymizewithtext}[1]{\textit{\textcolor{gray}{(removed to preserve anonymity)}}}

\definecolor{julia1}{rgb}{0, 0.60, 0.98}
\definecolor{julia2}{rgb}{0.89, 0.43, 0.28}
\definecolor{julia3}{rgb}{0.24, 0.64, 0.30}
\definecolor{julia4}{rgb}{0.76, 0.44, 0.82}
\definecolor{julia5}{rgb}{0.67, 0.55, 0.09}
\definecolor{julia6}{HTML}{00aaae}
\definecolor{julia7}{HTML}{ed5e93}
\definecolor{julia8}{HTML}{c68125}
\definecolor{julia9}{HTML}{00a98d}

\newcounter{xxx}
\title{AutoSGD: Automatic Learning Rate Selection for Stochastic Gradient Descent}

\author{%
  Nikola Surjanovic \\
  Department of Statistics \\
  University of British Columbia \\
  Vancouver, BC V6T 1Z4\\
  \texttt{nikola.surjanovic@stat.ubc.ca} \\
  \And  
  Alexandre Bouchard-C\^{o}t\'{e} \\
  Department of Statistics \\
  University of British Columbia \\
  Vancouver, BC V6T 1Z4\\
  \texttt{bouchard@stat.ubc.ca} \\
  \And
  Trevor Campbell \\
  Department of Statistics \\
  University of British Columbia \\
  Vancouver, BC V6T 1Z4\\
  \texttt{trevor@stat.ubc.ca} \\
}

\begin{document}

\maketitle

\begin{abstract}
  The learning rate is an important tuning parameter for stochastic gradient descent (SGD)
  and can greatly influence its performance. 
  However, appropriate selection of a learning rate schedule across all iterations 
  typically requires a non-trivial amount of user tuning effort.
  To address this, we introduce AutoSGD: an SGD method that 
  automatically determines whether to increase or decrease the learning 
  rate at a given iteration and then takes appropriate action.  
  We introduce theory supporting the convergence of AutoSGD, along with its 
  deterministic counterpart for standard gradient descent. 
  Empirical results suggest strong performance 
  of the method on a variety of traditional optimization problems and machine learning tasks.
\end{abstract}
\section{Introduction}
\label{sec:intro}

Stochastic gradient descent (SGD) \cite{robbins1951stochastic} and its many popular variants---such as 
AdaGrad \cite{duchi2011adagrad}, 
RMSProp \cite{hinton2012rmsprop}, and 
Adam \cite{kingma2014adam}---are indispensable tools for solving optimization
problems using cheap, noisy gradient evaluations. 
For instance, such optimization problems are frequently encountered in machine learning and 
statistics when the amount of training data is large and each stochastic gradient evaluation only 
involves a small subset of the training data.   

An important goal of modern optimization algorithms is to provide efficient convergence 
to an optimum of the objective function with as little tuning effort required by the user as possible. 
Gradient descent algorithms each differ in their set of tuning parameters, but the 
learning rate (sequence) is common among almost all such methods and is critical to performance.  
If the learning rate is chosen to be too large, SGD may become unstable or diverge; whereas 
if the learning rate is too small, the iterates may converge but at a painstakingly 
slow pace. Further, there is usually no ``one size fits all'' learning rate for a given 
optimization problem as different regions of the parameter space during optimization may benefit 
from larger or smaller learning rates. 

In this work we introduce AutoSGD, a new SGD algorithm that 
adaptively selects appropriate learning rates on the fly by comparing the 
performance of neighbouring (larger and smaller) learning rates. 
The iterates are divided into episodes, during which we assess which of three learning rates
is the winner. We also allow to reset to a previous point and consider a collection 
of smaller learning rates if we realize all learning rates are too large.
We establish that both AutoSGD and its deterministic counterpart (i.e., based on exact gradients) converge 
under appropriate assumptions on the objective function. 
In our experiments we find that AutoSGD outperforms or
is comparable to other SGD methods with minimal or no tuning effort.

\noindent \textbf{Related work.} 
Several authors have developed diagnostics
for detecting when SGD switches from the transient phase to the stationary phase.
For instance, if stationarity is detected, 
the learning rate can be decreased (e.g., halved).
The work of \cite{pflug1983determination,pflug1990confidence} 
considers inner products of consecutive gradients to detect when stationarity is
reached; this work and follow-up methods by \cite{chee2017convergence,chee2020convergence}
apply to SGD both with and without momentum.
Similar approaches are considered by \cite{pesme2020convergence,yaida2018fluctuation,lang2019sasa},
using different diagnostics to detect stationarity, and a 
related approach specific to variational inference is presented by 
\cite{welandawe2024rabvi}.
However, a drawback of such methods is that they do not allow for an 
increase in the learning rate as they do not prescribe what 
action should be taken during the transient phase of SGD.

Another line of work considers line-search variants in the stochastic setting 
\cite{galli2023linesearch,vaswani2019painlesssgd}. These methods require the 
specification of a maximum learning rate $\gamma_\text{max}$ from which they backtrack 
until a stochastic version of standard Armijo conditions are satisfied. 
The methods and theoretical results are primarily designed for the interpolation setting. 
Also, it is not clear how the stochastic Armijo conditions perform when combined 
with control variates for variance reduction in gradient estimates \cite{gower2020variance} 
(e.g., such as the ``sticking the landing'' estimator,
commonly used for black-box variational inference \cite{kim2024stl,roeder2017stl}). 
Furthermore, such methods are restricted by their choice of $\gamma_\text{max}$, which can 
hinder performance if it is chosen to be too small. 

Recently, hyperparameter-free methods similar in spirit to our proposed AutoSGD 
algorithm have emerged in both the deterministic and stochastic optimization settings 
\cite{defazio2023dadaptation,ivgi2023dog,khaled2023dowg,orabona2017coin,kreisler2024parameterfree,mishchenko2023prodigy,malitsky2019adgd,khaled2024tuningfree,loizou2021stochasticpolyak,orvieto2022polyak,attia2024parameterfree}.
In particular, the distance over gradients (DoG) algorithm 
\cite{ivgi2023dog}, and a corresponding weighted gradient version (DoWG) \cite{khaled2023dowg}, 
have proven to be empirically successful with theoretical convergence guarantees. 
Inspired by a result of \cite{carmon2022bisection}, they consider 
an estimate of the distance to the optimum divided by a square root of a (weighted) sum 
of squared gradient norms. 
The DoWG method \cite{khaled2023dowg} modifies the original DoG \cite{ivgi2023dog} 
algorithm and establishes convergence guarantees primarily in the deterministic setting. 
For optimizers such as D-Adaptation \cite{defazio2023dadaptation} (and DoG \cite{ivgi2023dog}), 
\cite{mishchenko2023prodigy} remark that ``the methods might still be slow because 
the denominator in the step size might grow too large over time.'' 

\section{Setup} 
\label{sec:background}

The goal in this work is to find a minimum of a differentiable objective function $f: \reals^d \to \reals$. 
Without loss of generality, throughout we assume that $f$ is nonnegative and attains its minimum at
$\min_x f(x) = 0$, as none of our proposed methods require knowledge of the minimum value attained by $f$.
We also assume that $f$ and its gradient $\nabla f$ take the form
\[
  f(x) = \E[f(x, u)], \qquad 
  \nabla f(x) = \E[g(x, u)], \qquad 
  u \sim \phi,
\]
for a noise distribution $\phi$. For the purposes of our algorithms,
we assume the ability to evaluate $g(x,u)$ exactly, and $f(x,u)$ up to an unknown constant
independent of $x$ but which may depend on $u$, i.e., we are able to evaluate $\stf$ where $f(x,u) = \stf(x,u) + C_u$.
For simplicity in the rest of the paper, we will simply refer to $f(x,u)$.
This setup and model of computation is typical of stochastic optimization problems where $f(x)$ and $\nabla f(x)$ are 
expensive or intractable to compute, e.g., 
the common empirical risk minimization setting  where $f(x) = n^{-1} \sum_{i=1}^n f_i(x)$ 
if we set $u \sim \text{Unif}(\{1, 2, \ldots, n\})$ and $f(x,u) = f_u$.
However, note that it also generalizes the case of classical deterministic optimization problems
where $f$ and $\nabla f$ are tractable, as we can 
simply set $f(x,u) = f(x)$ and $g(x,u) = \nabla f(x)$ (i.e., ignore the noise variable).

Starting with an initial iterate $y_0 \in \reals^d$,
a sequence of learning rates $(\gamma_k)_{k \geq 0}$,
and a noise sequence $(u_k)_{k \geq 0}$,
the SGD algorithm is defined by the sequence $(y_k)_{k \geq 0}$ produced by the iteration
\[
\label{eq:SGD}
  y_{k+1} = y_k - \gamma_k g(x_k, u_k). 
\]

\section{Automatic learning rate selection} 
\label{sec:auto_methods}

In this section we introduce AutoSGD, a stochastic gradient descent algorithm 
with automatic learning rate selection. 
To build intuition, we first present AutoGD, which is a deterministic version of AutoSGD useful for problems 
where evaluation of the full gradient is computationally feasible.

\subsection{AutoGD} 
\label{sec:autogd} 

\begin{figure*}[!t]
  \centering
  \begin{subfigure}{0.32\textwidth}
    \centering
    \includegraphics[width=\textwidth]{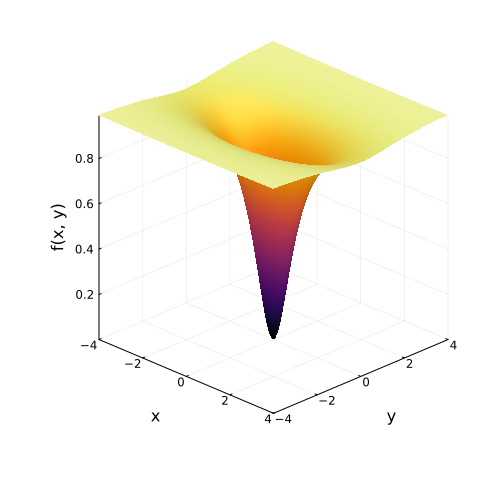}
  \end{subfigure}
  \begin{subfigure}{0.32\textwidth}
    \centering
    \includegraphics[width=\textwidth]{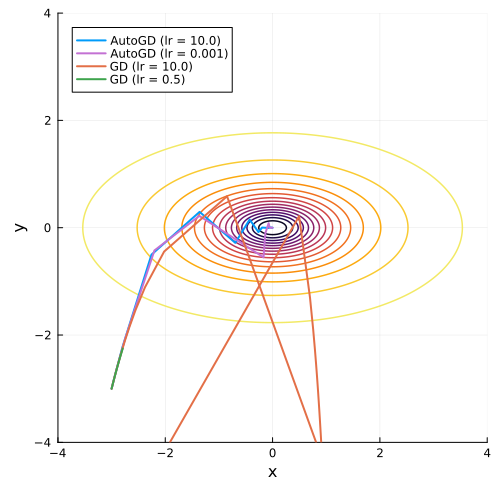}
  \end{subfigure}
  \begin{subfigure}{0.32\textwidth}
    \centering
    \includegraphics[width=\textwidth]{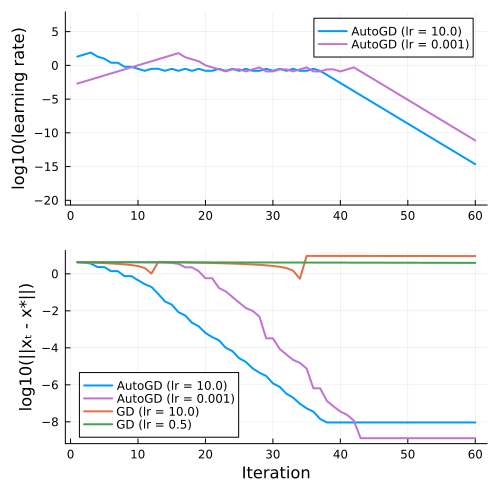}
  \end{subfigure}
  \caption{Performance of deterministic optimizers on the 
  non-convex objective function $f(x, y) = 1 - 1/(1 + x^2 + 4y^2)$. 
  \textbf{Left:} Surface plot of the objective function. 
  \textbf{Middle:} Trajectories of AutoGD with initial learning rates $\gamma_0 \in \{0.001, 10.0\}$ 
  and GD with learning rates $\gamma \in \{0.5, 10.0\}$ over 100 iterations. 
  Here, GD with $\gamma = 0.5$ converges very slowly, while $\gamma = 10.0$ is unstable.
  AutoGD is stable as it approaches the minimum for different learning rate values.
  \textbf{Top right:} Automatically selected learning rates (on log scale) for each of the first 60 iterations.
  AutoGD automatically learns to anneal the learning rate in the initial phase, and then 
  decreases the learning rate upon convergence. 
  \textbf{Bottom right:} Distance to optimum (log scale) for AutoGD and GD iterates.}
  \label{fig:autogd}
\end{figure*}

The AutoGD algorithm has the following inputs: 
an initial iterate $x_0$, 
initial learning rate $\gamma_0$, 
and learning rate expansion/contraction factors $0 < c < 1 < C < \infty$.
\cref{result:autogd_convergence2} shows that AutoGD works with a wide 
range of reasonable values of $C$ and $c$, but we recommend a default choice of $C = 1/c = 2$, 
which corresponds to doubling/halving the learning rate. 
We also note that even if $\gamma_0$ is initially too small or too large, we should reach an 
appropriate learning rate $\gamma$ within $\abs{\log_2(\gamma/\gamma_0)}$ iterations,
and so $\gamma_0$ can be set anywhere within a wide acceptable neighbourhood.

Suppose that at time $t \geq 0$ our current iterate and learning rate are $x_t$ and $\gamma_t$, 
respectively. We define the \emph{proposed} new learning rate and iterate, 
\[
  \gamma_{t+1}' &= \argmin_{\gamma \in \{c\gamma_t, \gamma_t, C\gamma_t\}} f(x_t - \gamma \nabla f(x_t)), \qquad
  x_{t+1}'       = x_t - \gamma_{t+1}' \nabla f(x_t). 
\]
(In case the $\argmin$ over $\gamma$ is not unique, we choose the largest $\gamma$.) 
Then we either accept or reject the proposal as follows: 
\[
  (x_{t+1}, \gamma_{t+1}) = 
  \begin{cases}
    (x_{t+1}', \gamma_{t+1}'), & f(x_{t+1}') < f(x_t) \\ 
    (x_t, c \cdot \gamma_t), & f(x_{t+1}') \geq f(x_t)
  \end{cases}.
\]
That is, we look forward one step to see which learning rate in $\{c\gamma_t, \gamma_t, C\gamma_t\}$ 
would perform best, and then use the best choice to obtain $x_{t+1}$. 
Because there is no noise in the deterministic setting, 
if all learning rates in the grid fail to decrease the objective, we decrease 
the learning rate by a factor of $c$ to consider a new grid of learning rates.
(In the common case where $C = 1/c$, we can decrease by $c^2$ to try a fresh 
collection of smaller learning rates.)

\cref{fig:autogd} shows the performance of AutoGD on a non-convex objective function 
and how the chosen learning rate varies as the algorithm proceeds.
In terms of the computational complexity of AutoGD compared to standard 
gradient descent, the main difference between the two algorithms is that the former 
evaluates three learning rates at each iteration instead of one. 
However, these evaluations can be performed in parallel and independently from one 
another. Therefore, with an efficient implementation, 
each iteration can have the same average runtime as GD, but requires a constant factor 
of three times as much memory.
Appendix~\ref{sec:autogd_experiments} contains additional experimental results that show excellent
performance of AutoGD on various tasks, along with a comparison to backtracking line search.

\subsection{AutoSGD}
\label{sec:autosgd}

\begin{figure*}[!t]
	\centering
	\includegraphics[width=0.9\textwidth]{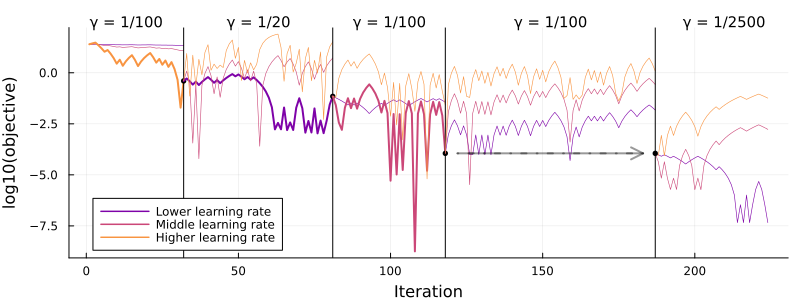}
	\caption{
		Example of the AutoSGD learning rate selection procedure with $C = 1/c = 5$
		(a larger value of $C$ is used for better visualization). 
		Episode endpoints are indicated as black vertical lines and the middle learning 
		rate within the episode is indicated at the top of each section. 
		The exact objective function value is known at episode endpoints here, but is 
		typically estimated in practice. Bold lines indicate winning trajectories.
		\textbf{Episode 0:} The highest learning rate is selected for the next episode. 
		\textbf{Episode 1:} The smallest learning rate is selected. 
		\textbf{Episode 2:} The middle learning rate is selected. 
		\textbf{Episode 3:} Evidence of function increase at all learning rates. Decrease by $1/C^2$
		and restart at the previous episode's starting point $x_{t-1}$.
		\textbf{Episode 4:} In progress.}
	\label{fig:autosgd_notation}
\end{figure*}

\begin{figure*}[!t]
  \centering
  \includegraphics[width=0.5\textwidth]{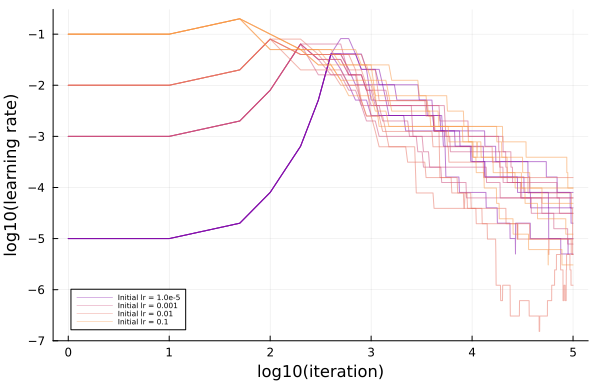}
  \caption{
    Performance of the new learning rate selection procedure within AutoSGD 
    (plotted on log-log scale) on the ``sum of quadratics'' problem. AutoSGD is initialized with four different learning rates 
    $\gamma \in \{10^{-1}, 10^{-2}, 10^{-3}, 10^{-5}\}$. 
    All three initializations learn to automatically 
    warm up the learning rate and eventually converge and decay at a rate of 
    approximately $O(1/t)$ in this example.
  }
  \label{fig:averaging_and_learning_rates}
\end{figure*}

AutoSGD generalizes AutoGD to the setting where we have access only to noisy estimates of the gradient and objective function.
Given a learning rate grid $\{c\gamma, \gamma, C\gamma\}$, 
the goal is to determine which learning rate yields the lowest objective 
value. Stochasticity in $f$ and $\nabla f$ must be handled, and so we introduce a 
decision process. This results in random lengths of time, called \emph{episodes}, 
where the learning 
rate grid is constant until a decision is made, at which point the grid may shift. 

To assess the performance of the three different learning rates, we keep track of 
three streams of SGD iterates (\cref{fig:autosgd_notation}). 
Consider an initial point $x_t$ and learning rate $\gamma_t$ at the start of episode $t$. 
We run three standard streams $k=1,2,\dots$ of SGD during the episode:
\[
  \slx_{t,k+1} &= \slx_{t,k}-c\gamma_t g(\slx_{t,k}, u_{t,k}) \\ 
  x_{t,k+1}    &= x_{t,k}-\gamma_t g(x_{t,k}, u_{t,k}) \\
  \sbx_{t,k+1} &= \sbx_{t,k}-C\gamma_t g(\sbx_{t,k}, u_{t,k}),
\]
which are each initialized according to $\slx_{t, 0} = x_{t,0} = \sbx_{t,0} = x_t$.
These three streams all stop at a (random) time $\tau_t$ when we have confidently identified what to do for the next episode (e.g., shift the grid to smaller learning rates if the 
stream $ \slx_{t,k}$ clearly outperforms the others).

We now describe the four valid moves at the end of an episode.
Let $I_t$, $S_t$, $D_t$, and $R_t$ be binary indicators denoting the decisions to 
increase $\gamma$, stay at $\gamma$, decrease $\gamma$, 
or restart to the previous episode's starting point while decreasing 
the learning rate. 
More precisely, at the end of episode $t$ with initial learning rate $\gamma_t$, we may either: 
increase the learning rate (event $I_t = 1$), with 
$(x_{t+1}, \gamma_{t+1}) = (\sbx_{t,\tau_t}, C \gamma_t)$;
decrease the learning rate ($D_t = 1$), with 
$(x_{t+1}, \gamma_{t+1}) = (\slx_{t,\tau_t}, c \gamma_t)$; 
stay at the same learning rate ($S_t = 1$), with 
$(x_{t+1}, \gamma_{t+1}) = (x_{t,\tau_t}, \gamma_t)$; or 
detect that all learning rates result in an increase in the objective 
($R_t = 1$), in which case $(x_{t+1}, \gamma_{t+1}) = (x_t, c \gamma_t)$.
\cref{fig:autosgd_notation} displays all four possible scenarios on a simple example.

The indicators $(I_t, D_t, S_t, R_t)$ are defined through a decision process. 
We leave them in a general form for our initial presentation and define one possible 
approach that can be computed efficiently with constant memory in the 
following section.
The AutoSGD algorithm is presented in \cref{alg:autosgd}. 

\begin{algorithm}[t]
	\begin{algorithmic}[1]
    \Require Initial state $x_0$, 
      initial learning rate $\gamma_0$, 
      number of episodes $T$,
      expansion coefficient $C > 1$ (default: $C = 2$), 
      contraction coefficient $0 < c < 1$ (default: $c = 1/2$).
    \For{$t$ {\bf in} 0, 1, \dots, $T-1$}
      \State $\tau_t \gets 0$,\hspace{.3cm} $\slx_{t,0}, x_{t,0}, \sbx_{t,0} \gets x_t$, \hspace{.3cm}$I_t, D_t, S_t, R_t \gets 0$
      \While{$I_t = D_t = S_t = R_t = 0$} \Comment{run three SGD streams in parallel}
        \State $\slx_{t,\tau_t+1} \gets \slx_{t,\tau_t} - c\gamma_t g(\slx_{t,\tau_t}, u_{t,\tau_t})$ 
        \State $x_{t,\tau_t+1} \gets x_{t,\tau_t} - \gamma_t g(x_{t,\tau_t}, u_{t,\tau_t})$ 
        \State $\sbx_{t,\tau_t+1} \gets \sbx_{t,\tau_t} - C\gamma_t g(\sbx_{t,\tau_t}, u_{t,\tau_t})$
        \LineComment{\cref{alg:autosgd_decisions}, see supplement for constant-memory/time \texttt{decision()} implementation}
        \State $I_t, D_t, S_t, R_t \gets \texttt{decision}((\slx_{t,k})_{k=0}^{\tau_t}, (x_{t,k})_{k=0}^{\tau_t}, (\sbx_{t,k})_{k=0}^{\tau_t})$
        \State $\tau_t \gets \tau_t+1$
      \EndWhile
      \State \textbf{if} $I_t = 1$, set $x_{t+1}, \gamma_{t+1} \gets \sbx_{t,\tau_t}, C \gamma_t$\Comment{increase learning rate}
      \State \textbf{elseif} $D_t = 1$, set $x_{t+1}, \gamma_{t+1} \gets \slx_{t,\tau_t}, c \gamma_t$\Comment{decrease learning rate}
      \State \textbf{elseif} $S_t = 1$, set $x_{t+1}, \gamma_{t+1} \gets x_{t,\tau_t}, \gamma_t$\Comment{stay at the same learning rate} 
      \State \textbf{else} set $x_{t+1}, \gamma_{t+1} \gets x_t, c \gamma_t$ \Comment{restart: objective increase detected at all learning rates}
      \State \textbf{endif}
    \EndFor
    \State \Return $x_T$
	\end{algorithmic}
  \caption{AutoSGD}
  \label{alg:autosgd}
\end{algorithm}

\subsubsection{A decision process} 
\label{sec:autosgd_decisions}

The $t^\text{th}$ episode length $\tau_t$ is random and
given the three SGD streams 
$(\slx_{t,j})_{j=0}^k, (x_{t,j})_{j=0}^k, (\sbx_{t,j})_{j=0}^k$
up to time $k$, we must decide whether to keep running the streams. 
If not, we must decide whether to increase the learning rate, decrease the learning rate, 
stay at the same learning rate, or reject all learning rates in the grid and restart 
at the previous episode's iterate.

We specify one possible stopping time that can be easily computed online with 
constant memory cost.
We introduce two new groups of mutually independent sources of noise for learning rate 
assessment: $u_{t,k}^{(1)}, u_{t,k}^{(2)}$, which are also independent of the 
gradient sources of noise $u_{t,k}$. 
Within episode $t$ at sub-episode iteration $k$, for $m=1,2$ define the differences 
\[
  \slDelta_{t,k}^{(m)} &= f(x_t, u_{t,k}^{(m)}) - f(\slx_{t,k}, u_{t,k}^{(m)}) \\
  \Delta_{t,k}^{(m)}   &= f(x_t, u_{t,k}^{(m)}) - f(x_{t,k}, u_{t,k}^{(m)}) \\ 
  \sbDelta_{t,k}^{(m)} &= f(x_t, u_{t,k}^{(m)}) - f(\sbx_{t,k}, u_{t,k}^{(m)}).
\]
Intuitively, a learning rate should perform well if the long-run averages of 
$\Delta_{t,k}^{(m)}$ are large relative to their estimated variance. 
To perform this assessment, we define
\[
\label{eq:z_statistic}
  Z_{t,k}
  = \frac{\sum_{j=0}^k \frac{1}{2} \left(\Delta_{t,j}^{(1)} + \Delta_{t,j}^{(2)}\right) }
    {\sqrt{\sum_{j=0}^k \max\left(\eps, \frac{1}{2} \left(\Delta_{t,j}^{(1)} - \Delta_{t,j}^{(2)}\right)^2\right)}},
\]
with $\eps = 10^{-12}$, to avoid division by zero in the rare cases where 
$\Delta_{t,k}^{(1)} \approx \Delta_{t,k}^{(2)}$ up to floating point error. 
The statistics $\slZ_{t,k}$ and $\sbZ_{t,k}$ for the 
lower and upper streams are defined analogously using $\slDelta_{t,k}^{(m)}$ and $\sbDelta_{t,k}^{(m)}$.
Crucially, we note that the $Z_{t,k}$ can be computed online with constant memory cost. 

We use these statistics inside 
\cref{alg:autosgd_decisions}, detailed in the supplement,
where we also describe the constant-memory online implementation
along with the exact conditions that a decision process should satisfy in order to guarantee 
AutoSGD convergence. 

\section{Theory} 
\label{sec:theory} 

In this section we prove various properties of AutoGD and AutoSGD.
Standard definitions, such as (strong) convexity and $L$-smoothness are deferred to 
the supplementary material.

\subsection{AutoGD}
\label{sec:autogd_theory}

The function evaluations of iterates produced by AutoGD must always converge.  
This means that AutoGD will never create unstable or divergent iterations, even 
under very weak assumptions on $f$.

\bprop 
\label{result:autogd_always_converges}
Let $f$ be differentiable. The iterates $f(x_t)$ of AutoGD 
converge (not necessarily to a local minimum): 
$\lim_{t \to \infty} f(x_t) = \slf$, for some $\slf \geq 0$.
\eprop

The following result highlights why it is important to consider a 
``no movement'' step in the algorithm, where we decrease the learning rate 
and restart from the previous value of $x_t$.
Suppose we were to set 
\[
\label{eq:autogd_forced_move}
  x_{t+1} = x_t - \gamma \nabla f(x_t), \quad \text{for some $\gamma \in \{c\gamma_t, \gamma_t, C\gamma_t\}$}
\]
at every iteration, even if objective function increase is detected at all $\gamma$ values.
Then the algorithm would diverge even on some simple polynomials.
The following results assumes that in our AutoGD algorithm we shrink the middle 
learning rate by $c$ if we detect an increase in the objective at all learning rates 
in the grid.
Recall that in practice, if $C = c^{-1}$, we can shrink by $c^2$. 
A similar result to the one below, with slightly modified constants, holds also 
in that special case. 

\bprop 
\label{result:autogd_diverges}
Consider iterates $(x_t)_{t \geq 0}$ of AutoGD with initial learning rate 
$\gamma_0$ and where at each iteration we are forced to update our iterates according 
to \eqref{eq:autogd_forced_move} (i.e., without the ``no movement'' option).
Suppose $\abs{x_0} > 1$ and that $f(x) = x^{2p}$ for $p \geq 2$ with 
\[
  p > \frac{c+1}{2 c^2 \gamma_0\abs{x_0}}.
\]
Then the iterates diverge exponentially: $\abs{x_t} = \Omega(c^{-t})$.
\eprop

In contrast, we know that our proposed version of AutoGD (with a ``no movement'' option)
always converges by \cref{result:autogd_always_converges}. Next, we establish 
conditions under which AutoGD converges to a minimum with a bound on the rate of convergence. 
For $x\in\reals^d$ with $\|\nabla f(x)\| \neq 0$ and $\gamma > 0$, define
\[
\label{eq:G}
  G(x, \gamma) &= \frac{\grad f(x)}{\|\grad f(x)\|}^T \int_0^1 2(1-t) \grad^2 f(x - t\gamma \grad f(x))\d t \frac{\grad f(x)}{\|\grad f(x)\|},
\]
which is used to control the rate of descent.
To state the result, we introduce a few assumptions.

\bassum 
\label{assum:unimodal_descent}
For all $x\in\reals^d$ with $\|\nabla f(x)\| \neq 0$, 
$f(x-\gamma \grad f(x))$ is unimodal (with a unique minimum) 
for $\gamma > 0$.
\eassum

\bassum
\label{assum:G_bound}
$f$ is $L$-smooth, twice continuously differentiable, and there exists 
$\sbgamma > 1/L$ such that for all $x$ with $\|\nabla f(x)\| \neq 0$, we have 
$\sbgamma - \frac{\sbgamma^2}{2} G(x, \sbgamma) \leq \frac{1}{2L}$.
\eassum

We remark that \cref{assum:unimodal_descent} is satisfied in the case where $f$ is 
$\mu$-strongly convex by definition. In this case, then the technical condition in  
\cref{assum:G_bound} also follows by noting that $G \geq \mu$, by strong-convexity, and setting  
$\sbgamma \geq (1+\sqrt{1-1/\kappa})/\mu$, where $\kappa = L/\mu$.

\bthm
\label{result:autogd_convergence2}
Suppose that $f$ is $\mu$-Polyak-\L{}ojasiewicz and 
\cref{assum:unimodal_descent,assum:G_bound} hold. Let
\[
  \tau  &= \lt\lceil \log_{c^{-1}} \frac{Lc^{-1}\sbgamma}{\sqrt{2}-1}\rt\rceil + 
    \max\lt\{\lt\lceil \log_{C} \frac{C^{-1}}{c} \rt\rceil, 
      \lt\lceil \log_{c^{-1}}\frac{C}{c^{-1}}\rt\rceil\rt\} + 1\\
  t_0 &= \max\lt\{
  0,
  \log_{c^{-1}}\frac{\gamma_0}{\sbgamma},
  \log_{C}\frac{(\sqrt{2}-1)}{L\gamma_0}
  \rt\}.
\]
Then
\[
  f(x_t) 
  &\leq f(x_0) \cdot \lt(1 - \frac{\mu c(\sqrt{2}-1)\lt(2C - c(\sqrt{2}-1)\rt)}{C^2 L}\rt)^
    {\lfloor{\frac{t-t_0}{\tau}}\rfloor}. 
\]
\ethm

From this result, we can see that regardless of the choice of initial learning rate, 
we can obtain a linear convergence rate of approximately $1/(\kappa \log \kappa)$. 
In contrast, optimally-tuned gradient descent has rate 
$1/\kappa$, but requires selecting the learning rate depending on the unknown quantity $L$.

\subsection{AutoSGD}
\label{sec:autosgd_theory}

We present a result that establishes AutoSGD convergence under some assumptions.
The exact conditions are described precisely in \cref{sec:autosgd_proof} 
and a statement of the following theorem with exact constants is also provided there.
The conditions roughly stipulate that: 
the indicators $I_t,S_t,D_t$ are more likely to trigger when their respective objective function is lower;
if the learning rate is too large, there is a high probability of restarting ($R_t = 1$); 
if the learning rate is too small, there is a high probability of increasing the learning rate quickly ($I_t=1$);
and if the learning rate is adequate, the test takes some time to trigger in order to 
guarantee a certain amount of descent.
\cref{result:autosgd_convergence} demonstrates that AutoSGD converges under
these assumptions, and that the rate is linear at a sequence of random 
episode iterations.
It is worth noting that the proof in \cref{sec:autosgd_proof} uses techniques drawn from the Markov chain
Monte Carlo literature that do not commonly appear in stochastic optimization, which may be useful for future study of similar adaptive algorithms.
\bthm 
\label{result:autosgd_convergence}
Fix a function $f:\reals^d \to \reals$ and suppose 
that the conditions listed in Appendix~\ref{sec:autosgd_proof} hold. 
Then there exists a random sequence of times $t_n$, with $t_n< \infty$ almost surely, 
and constants $0 \leq \nu < 1$ and $0 \leq f_0,a,b < \infty$, such that   
\[
  \forall n\in\nats_0, \quad 
  \E\lt[f(x_{t_{n}})\rt] &\leq \nu^n f_0  \quad\text{and}\quad 
  \E[t_n] \leq a + bn.
\]
\ethm
Recall that each $x_t$ is the start of a new \emph{episode}, whose length (in terms of individual SGD iterations) is 
$\tau_t$. While the above result demonstrates linear convergence in episode iterations,
it does not make any claim about the total number of SGD iterations. Indeed, the technical assumptions
in \cref{sec:autosgd_proof} indicate that $\tau_t$ should increase as $\gamma_t$ decreases,
which results in an overall sublinear convergence as expected for SGD with a 
decaying learning rate in the presence  of noise.

\section{Experiments} 
\label{sec:experiments}

We assess the performance of AutoSGD on both classical optimization objectives and 
various machine learning (ML) training experiments.  
We consider five classes of optimizers: SGD with various 
learning rates held constant, SGD with various initial learning rates and decay, 
Distance over Gradient (DoG) \cite{ivgi2023dog}, 
schedule-free SGD \cite{defazio2024road}, 
non-monotone line-search (NMLS) \cite{galli2023linesearch},
and AutoSGD. 
For NMLS, we use backtracking with the modified termination condition in \cite{galli2023linesearch}, 
but do not modify the initial learning rate as it requires knowing quantities 
such as $\inf f_i$.
The classical optimization experiments are performed in Julia 
\cite{bezanson2017julia}, while the remaining ML experiments are done in 
Python within the PyTorch \cite{paszke2019pytorch} framework.
All experiments are performed on the ARC Sockeye compute cluster at the University 
of British Columbia.
Experiments with AutoGD performed on deterministic optimization 
 problems \cite{surjanovic2013virtual} 
are presented in the supplement. 
In all plots, we define one \emph{iteration} (x-axis)  such that all methods incur roughly the same computational cost per iteration 
(i.e., we do not plot $x_t$ but rather all iterations within all episodes).

\subsection{Classical optimization} 
\label{sec:classic_optimization} 

We consider five classical optimization settings: 
linear regression, approximate matrix factorization, 
logistic regression, sums of quadratic functions, and multiclass logistic regression.
The last four settings are based on 
\cite{vaswani2019painlesssgd} and \cite{nutini2022block}.
Detailed descriptions of the experimental settings and additional figures 
are provided in the supplement.
In this section we present figures for the first three groups of experiments.

\cref{fig:averaging_and_learning_rates} shows a representative example of the learning rates 
selected by AutoSGD. 
We find that AutoSGD is able to successfully adapt the 
learning rate on various problems, automatically introducing an initial warmup 
and decay of the learning rate (more examples shown in the supplement). 

\begin{figure*}[!t]
  \centering
  \begin{subfigure}{0.32\textwidth}
    \centering
    \includegraphics[width=\textwidth]{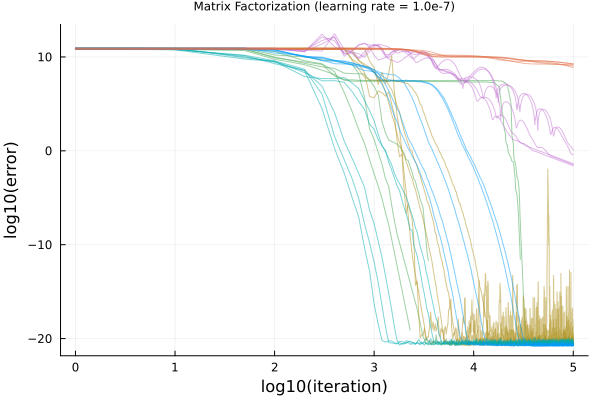}
  \end{subfigure} 
  \begin{subfigure}{0.32\textwidth}
    \centering
    \includegraphics[width=\textwidth]{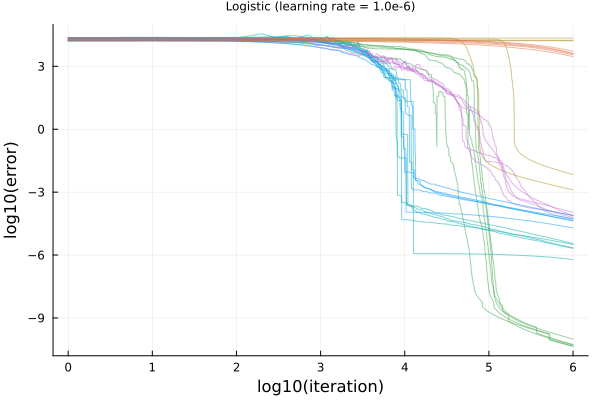}
  \end{subfigure}
  \begin{subfigure}{0.32\textwidth}
    \centering
    \includegraphics[width=\textwidth]{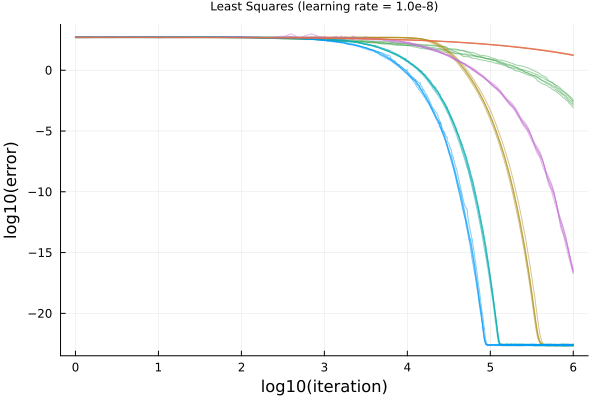}
  \end{subfigure}
  \caption{Classical optimization settings with learning rates chosen that are favourable 
  to the majority of optimizers (excluding AutoSGD and DoG). 
  Five different seeds are presented for each optimizer.
  The optimizers are labeled with the following colours:
  \textcolor{julia3}{AutoSGD}, 
  \textcolor{julia5}{DoG},
  \textcolor{julia4}{SFSGD},
  \textcolor{julia1}{SGD (constant)},
  \textcolor{julia2}{SGD (invsqrt)}, 
  \textcolor{julia6}{NMLS}. 
  The matrix factorization example corresponds to $k=10$.}
  \label{fig:classical_tuned}
\end{figure*}

\begin{figure*}[!t]
  \centering
  \begin{subfigure}{0.32\textwidth}
    \centering
    \includegraphics[width=\textwidth]{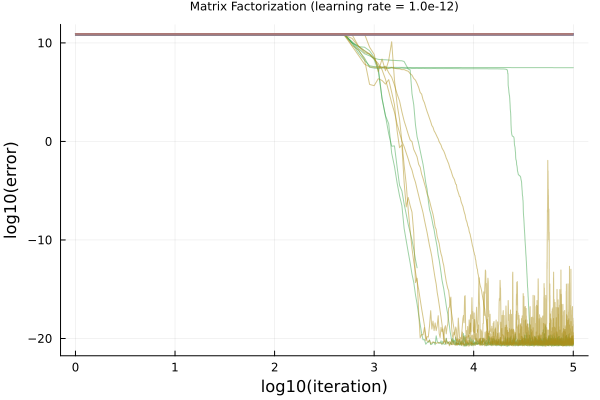}
  \end{subfigure}
  \begin{subfigure}{0.32\textwidth}
    \centering
    \includegraphics[width=\textwidth]{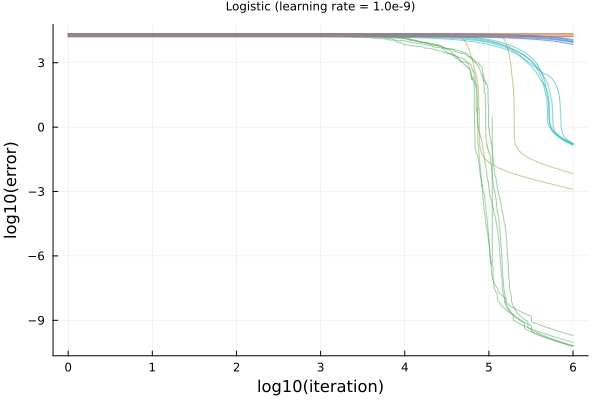}
  \end{subfigure}
  \begin{subfigure}{0.32\textwidth}
    \centering
    \includegraphics[width=\textwidth]{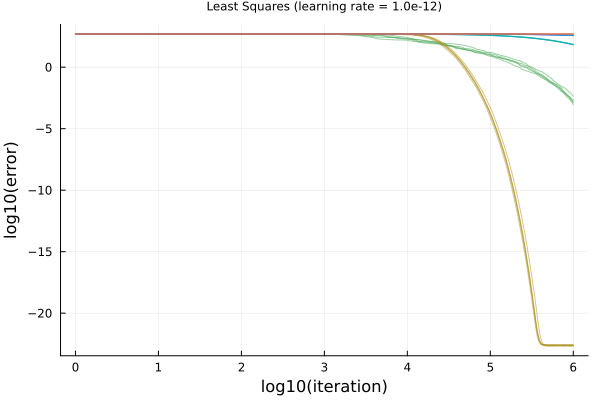}
  \end{subfigure}
  \caption{Classical optimization settings with initially small learning rates for the different optimizers 
  in order to assess robustness.}
  \label{fig:classical_small}
\end{figure*}

In \cref{fig:classical_tuned} we present results for choices of initial learning rates that 
result in best loss decay for the majority of methods (excluding AutoSGD and DoG). 
To highlight the robustness of AutoSGD to a specification of initial learning rate, 
in \cref{fig:classical_small} we present results for all optimizers with a 
smaller initial learning rate outside of its approximately optimal magnitude.
Additional experimental results are presented in \cref{fig:classical_tuned_additional,fig:classical_small_additional,fig:matrixfactor_additional}.
In all settings, AutoSGD and DoG perform well without the need for any tuning.
Generally, we find that AutoSGD has more stable performance once it reaches 
an optimum (even without any averaging), 
whereas DoG can exhibit some noise possibly due to the gradient noise. 
This is seen, for instance, in the 
matrix factorization example, where the DoG iterates display a considerable amount of noise 
towards the end of optimization.

In \cref{fig:classical_tuned} with tuned learning rates, we find that NMLS can 
perform quite well on certain problems. However, because its maximum learning rate is capped, 
if an initial guess is too low, its performance is severely hindered, 
as shown in \cref{fig:classical_small}.
We note that in all experiments we set the maximum learning rate for NMLS to 10 
times the amount specified for other optimizers in order for it to be able to explore a wider range 
of learning rates. However, even with this additional room, its performance still suffers
if the initial guess is too small.
We note that on the figures the iteration count is not scaled with respect 
to the number of NMLS backtracks, and so we would expect NMLS to perform 
\emph{a constant factor worse} than displayed here.

In the multiclass logistic regression setting 
(\cref{fig:classical_tuned_additional,fig:classical_small_additional}), we found that 
AutoSGD can sometimes momentarily be unstable before reverting back to a desirable loss.
Somewhat surprisingly, even with this temporary instability, the final training loss is 
still better than all other methods for most of the seeds. 

Finally, we find that on some of these problems, such as the matrix factorization 
and sum of quadratics problems, schedule-free SGD can exhibit oscillatory behaviour. 
Interestingly, for the ML training tasks in the following section, 
this oscillatory behaviour does not occur, presumably due to higher levels of gradient noise.

\subsection{ML training} 
\label{sec:ML_training}

For our ML experiments, we consider both natural language processing (NLP) and vision tasks. 
We fine-tune base models on various datasets, allowing us to 
remain within an academic compute budget.
For the NLP tasks, we fine-tune a RoBERTa-base model \cite{liu2019roberta} with different datasets from 
from the GLUE benchmark \cite{wang2019glue}: 
QNLI (question answering) \cite{rajpurkar2016qnli}, 
MRPC (semantic equivalence) \cite{dolan2005mrpc}, 
and SST-2 (sentiment analysis) \cite{socher2013sst2}.
For the vision tasks, we fine-tune a ResNet \cite{he2016resnet} on the 
CIFAR-10 and CIFAR-100 data sets \cite{krizhevsky2009cifar}. 
This experimental setup is similar to some of those considered in \cite{ivgi2023dog}.
Based on the results from \cref{sec:classic_optimization}, we exclude optimizers 
that are not robust to the choice of initial learning rate, with the exception of 
SGD and schedule-free SGD as they are both widely used.
For these experiments, we consider learning rates 
$\gamma \in \{10^{-1}, 10^{-2}, 10^{-3}\}$, which seem to be an acceptable range 
for this problem set, and then keep 
the best and worst performer for each optimizer that has a learning rate parameter.

\begin{figure*}[!t]
  \centering
  \begin{subfigure}{0.32\textwidth}
    \centering
    \includegraphics[width=\textwidth]{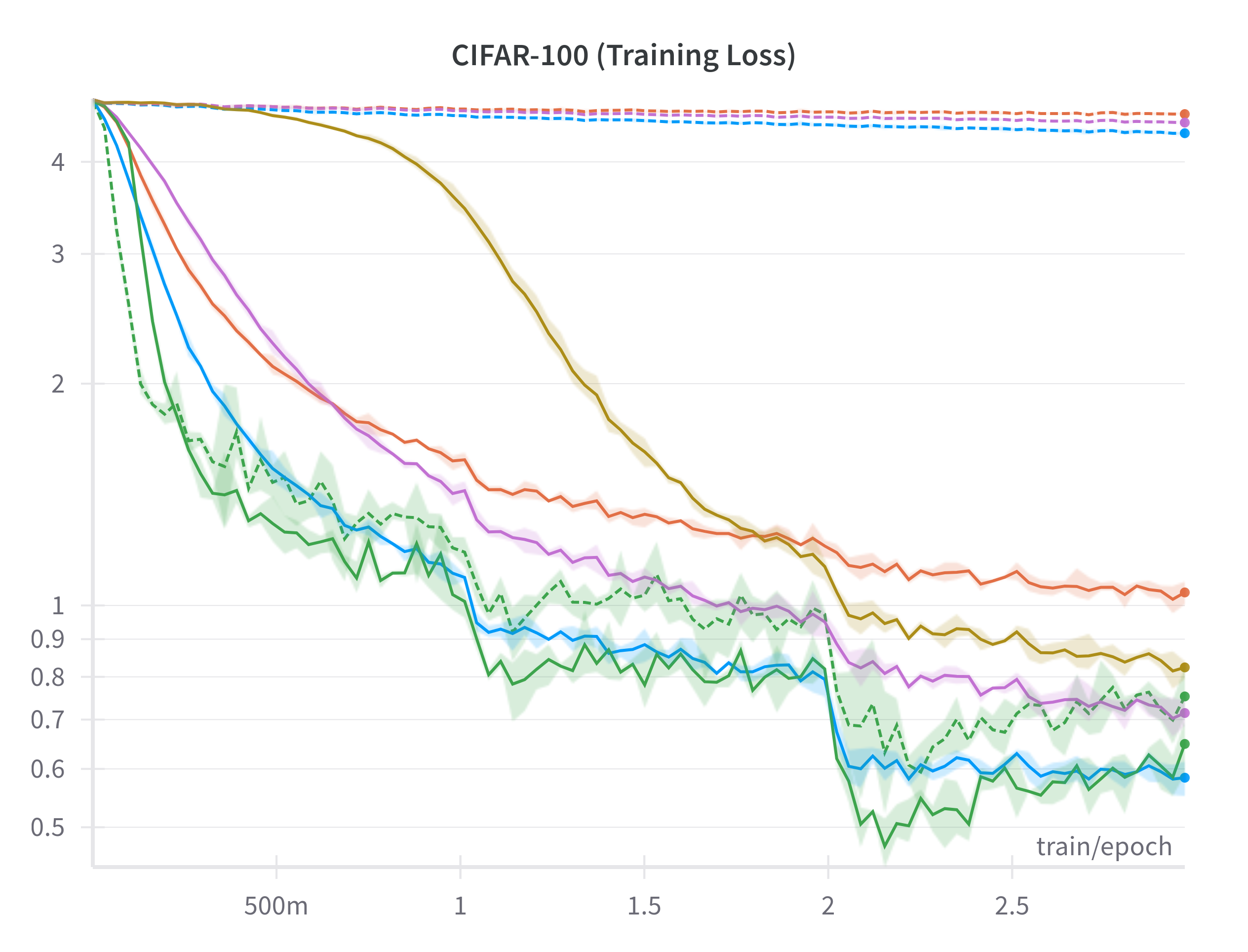}
  \end{subfigure}
  \begin{subfigure}{0.32\textwidth}
    \centering
    \includegraphics[width=\textwidth]{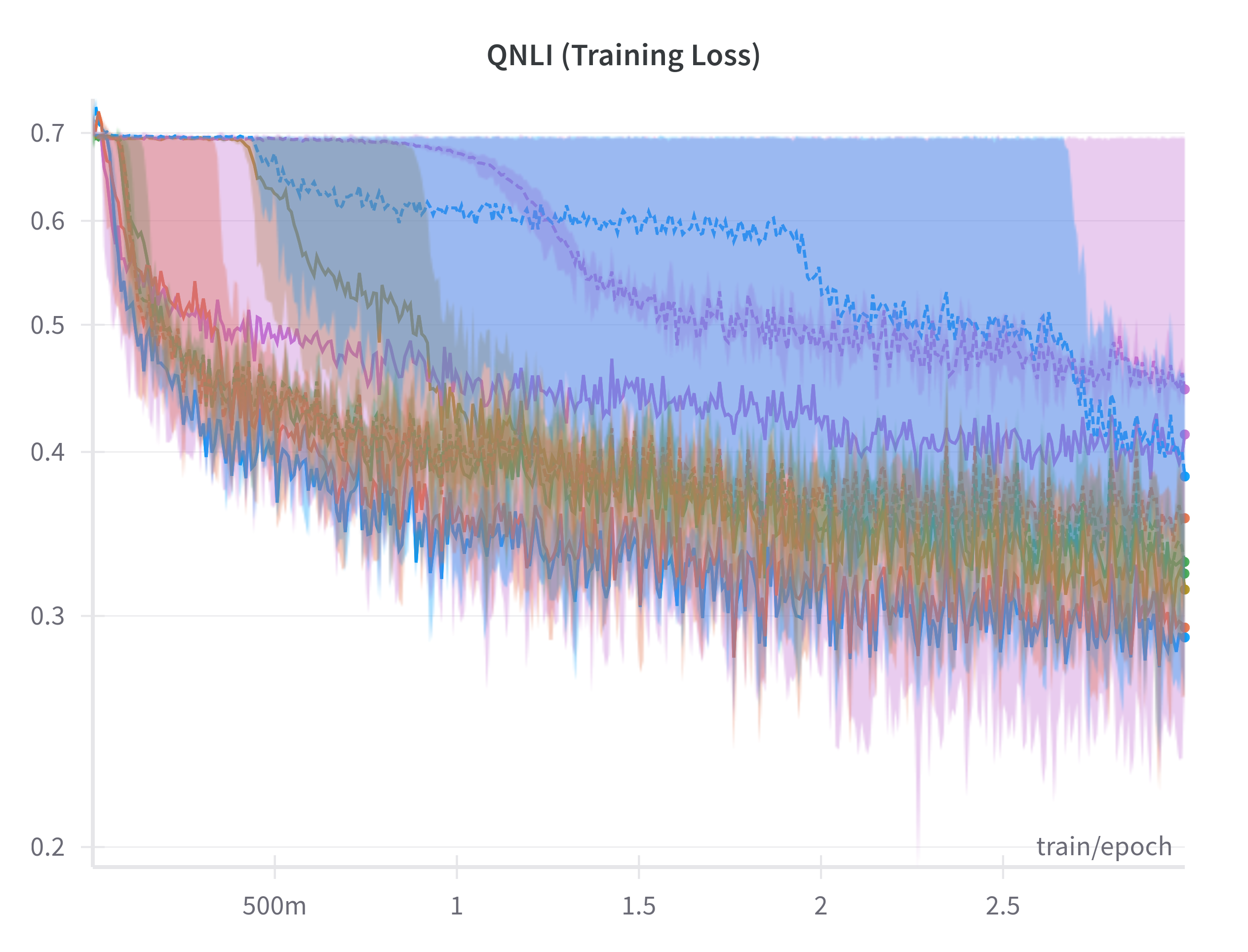}
  \end{subfigure}
  \begin{subfigure}{0.32\textwidth}
    \centering
    \includegraphics[width=\textwidth]{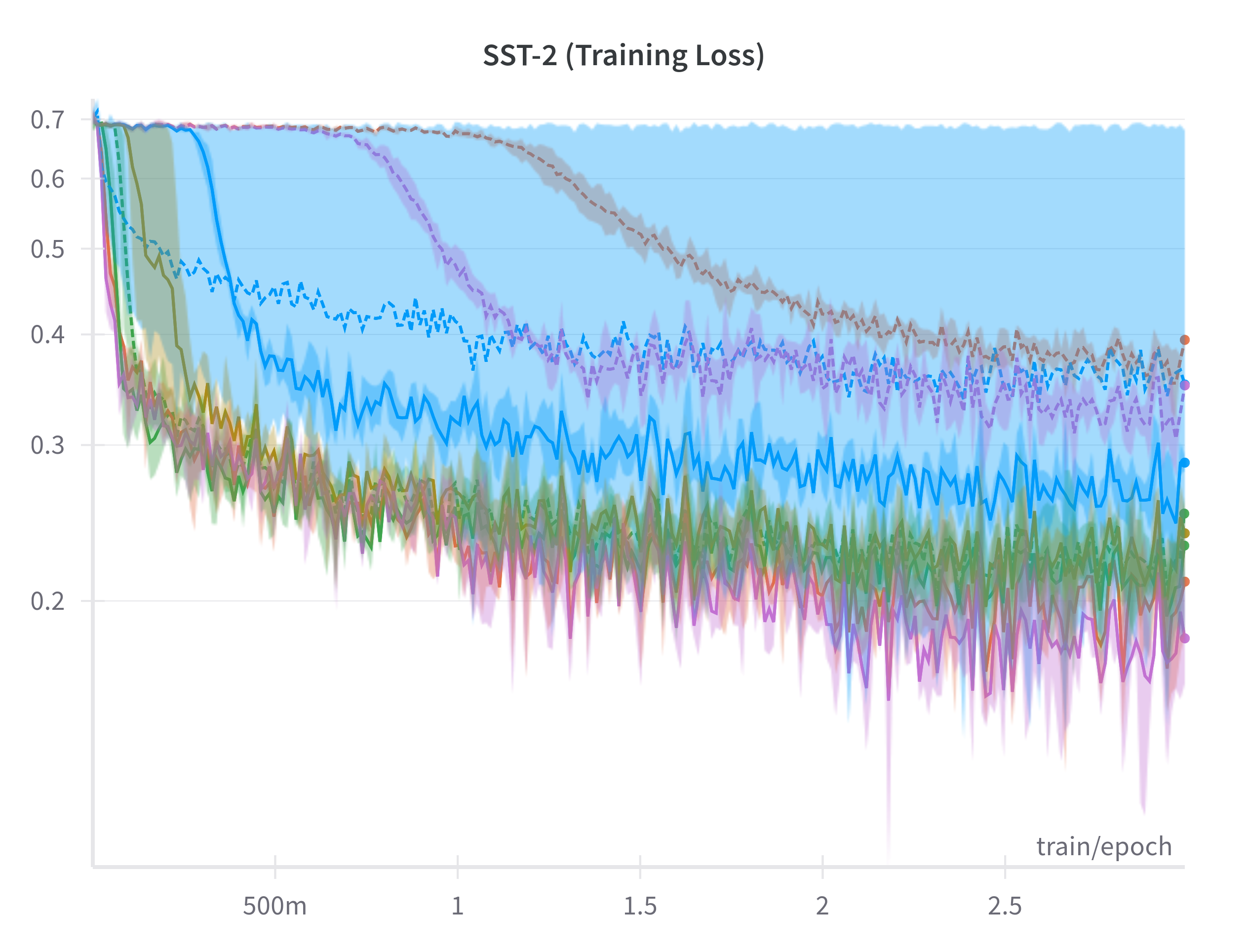}
  \end{subfigure}
  \caption{Training loss for various optimizers on three out of five ML data sets. Solid lines represent averages across seeds, 
  and shaded regions are the maximum and minimum across three seeds for a given optimizer. 
  For each data set and optimizer configuration, we present the best (solid line) 
  and worst (dashed line) learning rate 
  selections from a predefined grid. The colour scheme is the same as 
  in \cref{fig:classical_tuned}: 
  \textcolor{julia3}{AutoSGD}, 
  \textcolor{julia5}{DoG},
  \textcolor{julia4}{SFSGD},
  \textcolor{julia1}{SGD (constant)},
  \textcolor{julia2}{SGD (invsqrt)}.}
  \label{fig:ML_training}
\end{figure*}

\begin{figure*}[!t]
  \centering
  \begin{subfigure}{0.32\textwidth}
    \centering
    \includegraphics[width=\textwidth]{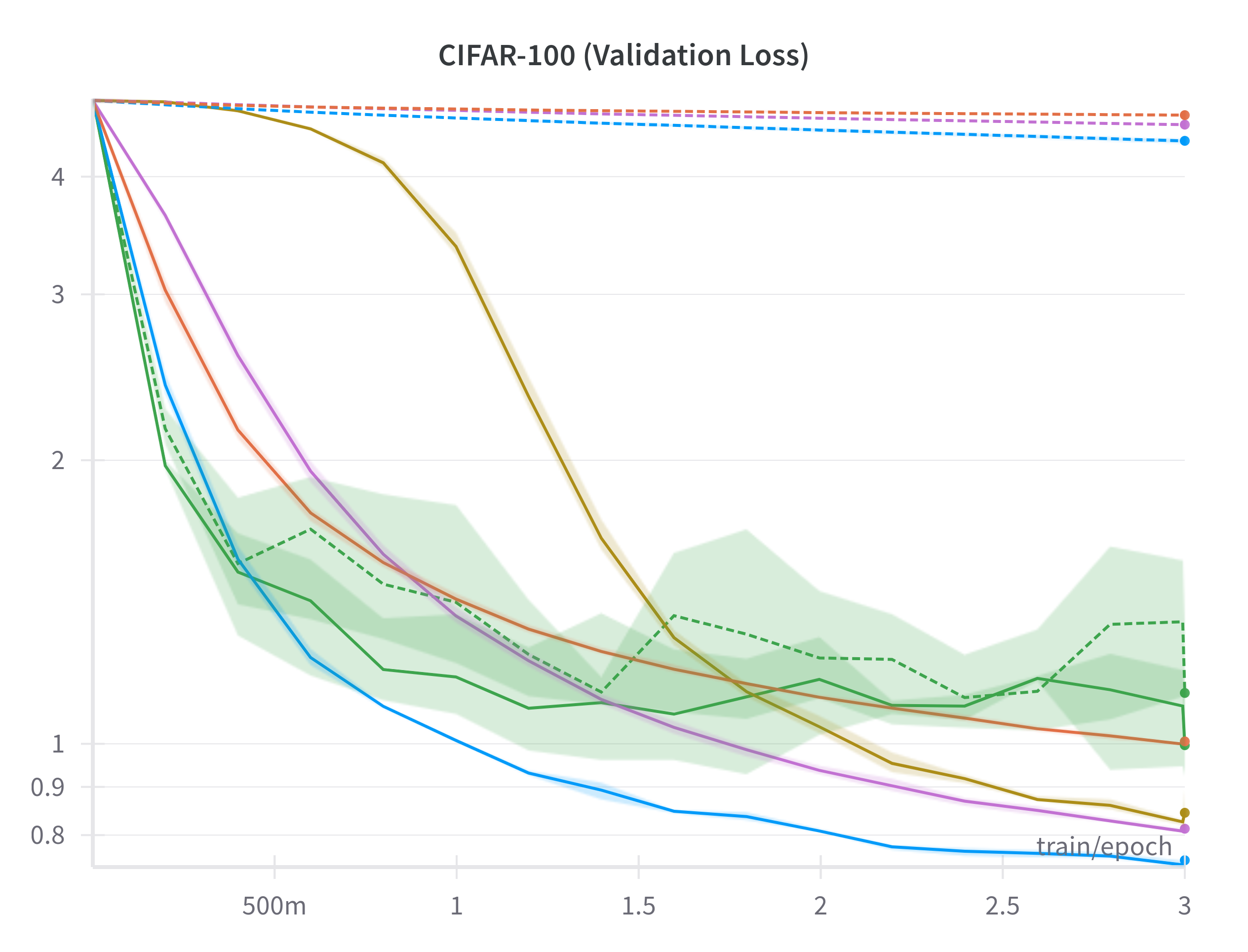}
  \end{subfigure}
  \begin{subfigure}{0.32\textwidth}
    \centering
    \includegraphics[width=\textwidth]{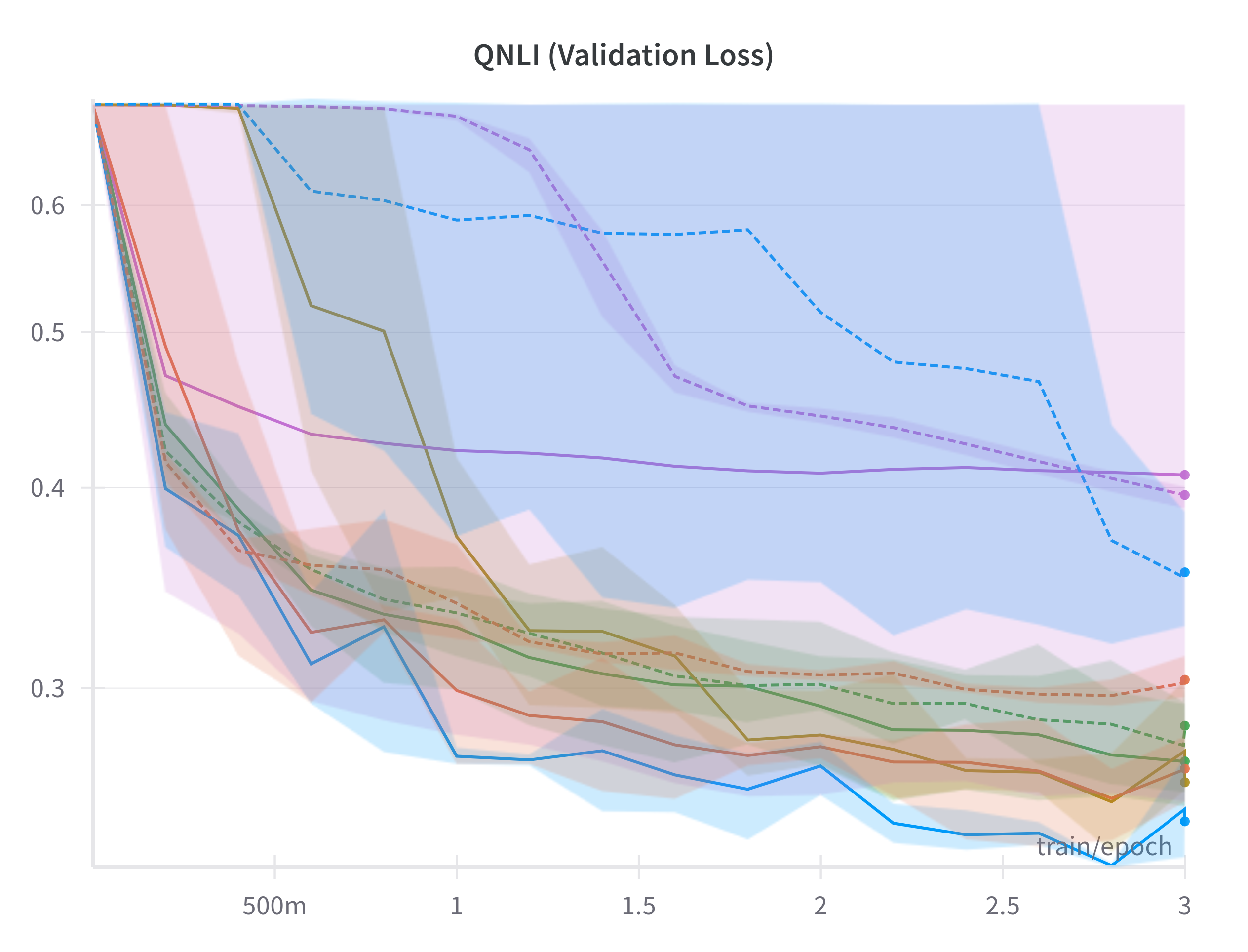}
  \end{subfigure}
  \begin{subfigure}{0.32\textwidth}
    \centering
    \includegraphics[width=\textwidth]{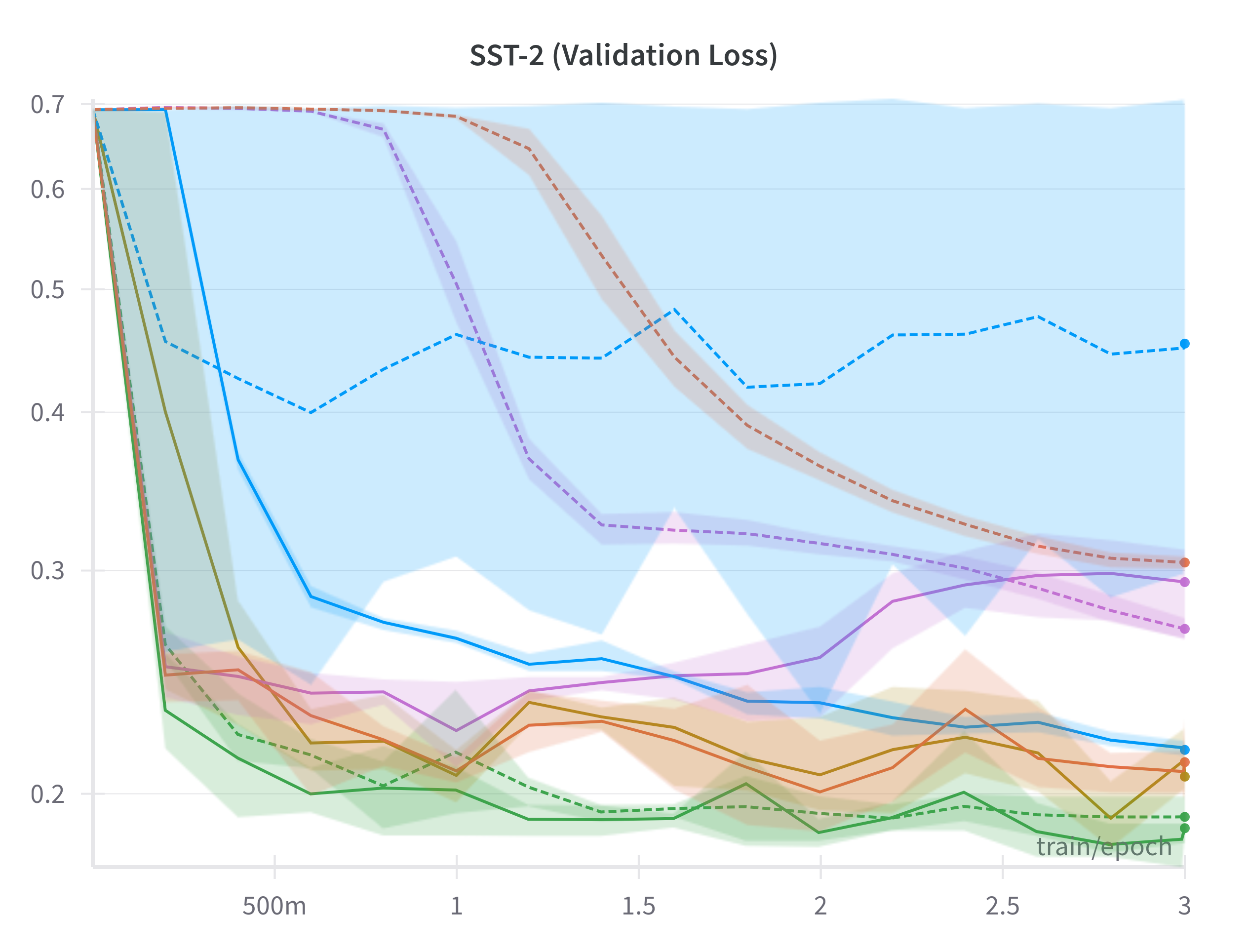}
  \end{subfigure}
  \caption{Validation loss for various optimizers on three out of five ML data sets.}
  \label{fig:ML_validation}
\end{figure*}

\cref{fig:ML_training,fig:ML_validation} present the training and validation loss, respectively, 
for three out of the five settings (CIFAR-100, QNLI, SST-2), with the remaining 
two (CIFAR-10, MRPC) presented in \cref{fig:ML_training_additional,fig:ML_validation_additional}.
From these results we take away the following: AutoSGD and DoG are both 
robust and perform well across different settings with minimal or no tuning. 
Between these two, which one performs better on a given task is problem-specific, 
and so we view both as useful optimizers in the ML training toolkit.
We find that AutoSGD can sometimes be slightly sensitive to the initial learning 
rate parameter choice, but generally performs quite well. This is in contrast 
to schedule-free SGD, which can be highly sensitive to the choice of learning rate 
parameter. However, when the learning rate is chosen carefully for 
schedule-free SGD, it can sometimes significantly outperform 
both AutoSGD and DoG, while a poorly chosen learning rate can lead to 
considerably weaker training and validation loss.

\section{Discussion} 
\label{sec:discussion}

In this paper we presented AutoSGD, an algorithm for stochastic gradient descent 
with automatic selection of learning rates. 
In our experiments, we found that AutoSGD is robust to the choice of initial 
learning rate and outperforms or is on par with its competitors
on many optimization problems.
These experimental results are also supported by convergence theory 
for both AutoSGD and its deterministic counterpart, AutoGD. 
We note that the convergence proof for AutoSGD used techniques drawn from the Markov chain
Monte Carlo literature, which may be useful for future developments and analyses.

Naturally, some directions for future work remain.  
Extensions to a learning rate proposal grid with more than three elements 
would raise interesting questions about an optimal number of streams and the choice 
of spacing between learning rates. 
Further, in this work we present the AutoSGD algorithm with a general 
decision process and consider one possible implementation of such a process. 
Exploring other decision processes and studying their properties would be 
a useful component of future work.

\begin{ack}
ABC and TC acknowledge the support of an NSERC Discovery Grant. 
NS acknowledges the support of a Four Year Doctoral Fellowship from the University of British Columbia. 
We additionally acknowledge use of the ARC Sockeye computing platform from the 
University of British Columbia.
\end{ack}

% \clearpage

\bibliographystyle{alpha}
\bibliography{main.bib}

%%%%%%%%%%%%%%%%%%%%%%%%%%%%%%%%%%%%%%%%%%%%%%%%%%%%%%%%%%%%

\appendix
\clearpage

\section{Common definitions} 
\label{sec:supp_definitions} 

Some standard definitions are listed below for completeness.

\bdefn{(Convex function)} 
\label{def:convex}
A function $f : \reals^d \to \reals$ is convex if for any $x,y \in \reals^d$ and $t \in [0,1]$ 
we have 
\[
  f(tx + (1-t)y) \leq t f(x) + (1-t) f(y).
\]
\edefn

\bdefn{(Strongly convex function)} 
\label{def:strongly_convex}
A function $f : \reals^d \to \reals$ is $\mu$-strongly convex for some $\mu > 0$ 
if for any $x,y \in \reals^d$ and $t \in (0,1)$ we have 
\[
  f(tx + (1-t)y) + \mu \frac{t(1-t)}{2} \|x - y\|^2 \leq t f(x) + (1-t) f(y).
\]
\edefn

\bdefn{(Polyak--\L{}ojasiewicz function)} 
\label{def:PL}
A differentiable function $f : \reals^d \to \reals$ is $\mu$-Polyak--\L{}ojasiewicz (P\L{}) 
for some $\mu > 0$ if it is bounded from below and for any $x\in \reals^d$ we have 
\[
  f(x) - \inf f \leq \frac{1}{2\mu} \|\nabla f(x)\|^2.
\]
\edefn

\clearpage
\section{AutoGD proofs} 
\label{sec:autogd_proofs}

\bprfof{\cref{result:autogd_always_converges}}
The sequence $(f(x_t))_{t \geq 0}$ satisfies $f(x_{t+1}) \leq f(x_t)$ for all 
$t$, and $f(x_t) \geq 0$. By the monotone convergence theorem, 
$\lim_{t \to \infty} f(x_t) = \underline{f} \geq 0$.
\eprfof

%%%%%%%%%%%%%%%%%%%%%%%%%%%%%%%%%%%%%%%%%%%%%%%%%%%%%%%%%%%%%%%%%%%%%%%%%%%%%%%%

\bprfof{\cref{result:autogd_diverges}}
We show $\abs{x_{t+1}} > c^{-1} \abs{x_t}$ and $\gamma_t = c^t \gamma_0$ 
for all $t \geq 0$. 
We proceed by induction. 
By assumption, we have that 
\[
  \abs{x_0} > \max\left\{1, \frac{1}{2p c \gamma_0}, \frac{1}{2p c \gamma_0} \cdot \frac{c+1}{c}\right\}.
\]
Then note that
\[
  x_1 
  = x_0 - \gamma \nabla f(x_0) 
  = x_0 \left(1 - 2p \gamma x_0^{2(p-1)}\right)
\]
for some $\gamma \in \{c\gamma_0, \gamma_0, C\gamma_0\}$.
Therefore, for any $\gamma \in \{c\gamma_0, \gamma_0, C\gamma_0\}$,
\[
  \abs{x_1} 
  = \abs{x_0} \cdot \left(2p \gamma x_0^{2(p-1)} - 1\right) 
  \geq \abs{x_0} \cdot \left(2p \gamma \abs{x_0} - 1\right) 
  > \abs{x_0} \left( \frac{c+1}{c} - 1\right) 
  = c^{-1} \abs{x_0}. 
\]
Therefore, we also have $\gamma_1 = c \gamma_0$. This concludes the base case. 

Consider now any $t \geq 1$ and suppose that $\abs{x_i} > c^{-1} \abs{x_{i-1}}$ and that $\gamma_i = c^i \gamma_0$
for all $0 \leq i \leq t$. 
This implies that $\abs{x_i} > c^{-i} \abs{x_0}$ for any $i \leq t$.
Therefore, for any $\gamma \in \{c\gamma_t, \gamma_t, C\gamma_t\}$,
\[
  2p \gamma x_t^{2(p-1)} 
  &\geq 2 p c^{t+1} \gamma_0 x_t^{2(p-1)} \\
  &\geq 2 p c^{t+1} \gamma_0 \abs{x_0}^{2(p-1)} c^{-2t(p-1)} \\
  &= 2 p c \gamma_0 \abs{x_0} \cdot c^t \abs{x_0}^{2p-3} c^{-2t(p-1)} \\
  &> \frac{c+1}{c} \cdot c^t c^{-2t(p-1)} \\
  &> \frac{c+1}{c}, 
\]
and so 
\[
  \abs{x_{t+1}}
  = \abs{x_t} \cdot \left( 2p\gamma x_t^{2(p-1)} - 1 \right) 
  > c^{-1} \abs{x_t},
\]
and $\gamma_{t+1} = c \gamma_t = c^{t+1} \gamma_0$.
This completes the proof.
\eprfof

%%%%%%%%%%%%%%%%%%%%%%%%%%%%%%%%%%%%%%%%%%%%%%%%%%%%%%%%%%%%%%%%%%%%%%%%%%%%%%%%
\newpage

In the following two sections we present two different proofs of convergence 
for AutoGD. 

\subsection{First proof of AutoGD convergence}

\bprfof{\cref{result:autogd_convergence2}}
Since $f$ is $\mu$-P\L{}, if $\|\nabla f(x)\| = 0$, we have $f(x) = 0$ 
and the bound trivially holds. It therefore suffices to check the bound for 
sequences $x_t$ where $\|\nabla f(x_t)\| \neq 0$. 

By the Taylor remainder theorem, for any $x, x' \in \reals^d$,
\[
f(x') = f(x) + \grad f(x)^T(x' - x) + (x'-x)^T \int_0^1 (1-t) \grad^2 f((1-t)x+tx')\d t (x'-x).
\]
If $x' = x - \gamma \grad f(x)$ for some $\gamma > 0$, then
\[
f(x') = f(x) - \gamma \|\grad f(x)\|^2 + \gamma^2 \grad f(x)^T \int_0^1 (1-t) \grad^2 f(x - t\gamma \grad f(x))\d t \grad f(x),
\]
and so if $\|\nabla f(x)\| \neq 0$,
\[
f(x') = f(x) - \|\grad f(x)\|^2\lt(\gamma -  \frac{\gamma^2}{2} G(x, \gamma)\rt).
\]

Note that by smoothness,
\[
\label{eq:smoothness_bound}
f(x') \leq f(x) - \|\grad f(x)\|^2\lt(\gamma -  \frac{\gamma^2}{2} L\rt). 
\]
Therefore, for $\gamma = 1/L$,
\[
f(x') \leq f(x) - \frac{1}{2L}\|\grad f(x)\|^2.
\]
By smoothness again,
\[
f(x') \geq f(x) - \|\grad f(x)\|^2\lt(\gamma +  \frac{\gamma^2}{2} L\rt). 
\]
Therefore, for $\gamma \leq (-1+\sqrt{2})/L$,
\[
f(x')
&\geq f(x)- \frac{1}{L}\|\grad f(x)\|^2\lt(-1+\sqrt{2} + \frac{(-1+\sqrt{2})^2}{2}\rt)\\
&\geq f(x) - \frac{1}{2L}\|\grad f(x)\|^2.
\]
By \cref{assum:G_bound}, there exists an $\sbgamma > 1/L$ such that
\[
 \sbgamma - \frac{\sbgamma^2}{2}G(x,\sbgamma) \leq \frac{1}{2L},
\]
and hence, for $x' = x-\sbgamma \nabla f(x)$,
\[
f(x') &\geq f(x) - \frac{1}{2L}\|\grad f(x)\|^2.
\]
Now, by unimodality (\cref{assum:unimodal_descent}), we have that the previous inequality 
holds for any $\gamma \geq \sbgamma$.
Therefore, for all $x$, the position-dependent optimal learning rate $\gamma^\star(x)$ satisfies
\[
\gamma^\star(x) \in \lt[ \frac{\sqrt{2}-1}{L}, \sbgamma\rt], \qquad 
f(x-\gamma^\star(x) \grad f(x))  \leq f(x) - \frac{1}{2L}\|\grad f(x)\|^2.
\]

We first examine the time that it takes for the learning rate to reach a desirable set  
$\lt(C^{-1}\frac{\sqrt{2}-1}{L}, c^{-1}\sbgamma\rt)$.
Suppose $\gamma_0 \geq c^{-1}\sbgamma$. Then by \cref{assum:unimodal_descent}, 
the learning rate must initially shrink for $k$ iterations, until $\gamma < c^{-1} \sbgamma$, 
where
\[
k &\leq \lt\lfloor \log_{c^{-1}}\frac{\gamma_0}{c^{-1}\sbgamma} \rt\rfloor + 1
= \lt\lfloor \log_{c^{-1}}\frac{\gamma_0}{\sbgamma}-1 \rt\rfloor + 1
\leq \log_{c^{-1}}\frac{\gamma_0}{\sbgamma}.
\]
Similarly, if $\gamma_0 \leq C^{-1}\frac{\sqrt{2}-1}{L}$, the learning rate must 
initially grow for $k$ iterations until $\gamma > C^{-1} \frac{\sqrt{2}-1}{L}$, where
\[
k &\leq \lt\lfloor \log_{C}\frac{C^{-1}(\sqrt{2}-1)}{L\gamma_0}\rt\rfloor + 1
= \lt\lfloor \log_{C}\frac{(\sqrt{2}-1)}{L\gamma_0}-1\rt\rfloor + 1
\leq \log_{C}\frac{(\sqrt{2}-1)}{L\gamma_0}.
\]
Therefore, for the first $k \leq t_0$ iterations, where 
\[
  t_0 &= \max\lt\{0, \log_{c^{-1}}\frac{\gamma_0}{\sbgamma}, 
  \log_{C}\frac{(\sqrt{2}-1)}{L\gamma_0} \rt\},
\]
we have $f(x_{t+1}) \leq f(x_t)$. We then also have 
$\gamma_k \in \lt(C^{-1}\frac{\sqrt{2}-1}{L}, c^{-1}\sbgamma\rt)$.

For the remainder of the proof, we therefore assume $\gamma_t \in \lt(C^{-1}\frac{\sqrt{2}-1}{L}, c^{-1}\sbgamma\rt)$.
By \cref{assum:unimodal_descent}, since $\gamma^\star(x_t) \in \lt[\frac{\sqrt{2}-1}{L}, \sbgamma\rt]$, 
the only case in which it is possible for the learning rate to decrease is if 
$\gamma_t > \gamma^\star(x_t) \geq \frac{\sqrt{2}-1}{L}$.
Similarly, the only case in which it is possible for the learning to increase is if 
$\gamma_t < \gamma^\star(x_t) \leq \sbgamma$. Therefore,
\[
\min\{c, C^{-1}\} \frac{\sqrt{2}-1}{L}< \gamma_{t+1} < \max\{C, c^{-1}\} \sbgamma.
\]
If $\gamma_{t+1} > c^{-1}\sbgamma$ (which is only possible if $C > c^{-1}$), 
then the learning rate will decrease for 
\[
k \leq \lt\lceil \log_{c^{-1}}\frac{C}{c^{-1}}\rt\rceil
\]
iterations until $\gamma_t < c^{-1}\sbgamma$ again.
Likewise, if $\gamma_{t+1} < C^{-1}\frac{\sqrt{2}-1}{L}$ (which is again only possible 
if $C > c^{-1}$), the learning rate will increase for 
\[
k \leq \lt\lceil \log_{C} \frac{C^{-1}}{c} \rt\rceil
\]
iterations until $\gamma_t > C^{-1}\frac{\sqrt{2}-1}{L}$.
Also, if $\gamma_t > c^{-1}\gamma^\star(x_t)$, then 
$\gamma_{t+1} = c\gamma_t$ and $f(x_{t+1}) \leq f(x_t)$.
Note that the maximum number of iterations where this can occur in a row (starting from iteration $t$)
is 
\[
k \leq \lt\lceil \log_{c^{-1}} \frac{\gamma_t}{\frac{-1+\sqrt{2}}{L}} \rt\rceil 
= \lt\lceil \log_{c^{-1}} \frac{L\gamma_t}{\sqrt{2}-1}\rt\rceil
\leq \lt\lceil \log_{c^{-1}} \frac{Lc^{-1}\sbgamma}{\sqrt{2}-1}\rt\rceil.
\]

Finally, suppose $\gamma_t \leq c^{-1}\gamma^\star(x_t)$. Therefore, $c \gamma_t \leq \gamma^\star(x_t)$. 
By \cref{assum:unimodal_descent}, the function is decreasing in $\gamma$ until $\gamma^\star$; therefore, 
we can bound the descent by assuming we pick the lower learning rate $c\gamma_t$,
and assume $\gamma_t$ takes its smallest value $\gamma_t = \frac{\sqrt{2}-1}{CL}$ 
(although it may actually be the case that we pick $C \gamma_t$ or $\gamma_t$).
From \cref{eq:smoothness_bound}, and the fact that $f$ is $\mu$-P\L{},
\[
f(x_{t+1}) 
&\leq f(x_t)  - \|\grad f(x)\|^2\lt(c\frac{\sqrt{2}-1}{CL} - \lt(c\frac{\sqrt{2}-1}{CL}\rt)^2 \frac{L}{2} \rt)\\
&= f(x_t)  - \frac{c(\sqrt{2}-1)}{2C^2 L}\lt(2C - c(\sqrt{2}-1)\rt)\|\grad f(x)\|^2\\
&\leq f(x_t) \lt(1 - \frac{\mu c(\sqrt{2}-1)\lt(2C - c(\sqrt{2}-1)\rt)}{C^2 L}\rt).
\]
Therefore, from entry in $\lt(C^{-1}\frac{\sqrt{2}-1}{L}, c^{-1}\sbgamma\rt)$, 
we may perform  
\[
  k \leq \lt\lceil \log_{c^{-1}} \frac{Lc^{-1}\sbgamma}{\sqrt{2}-1}\rt\rceil + \max\lt\{\lt\lceil \log_{C} \frac{C^{-1}}{c} \rt\rceil, \lt\lceil \log_{c^{-1}}\frac{C}{c^{-1}}\rt\rceil\rt\} + 1
\]
iterations until reentry into the set, of which at least one iteration guarantees
strict descent. Combining the above bounds yields the final result.
\eprfof

%%%%%%%%%%%%%%%%%%%%%%%%%%%%%%%%%%%%%%%%%%%%%%%%%%%%%%%%%%%%%%%%%%%%%%%%%%%%%%%%

\subsection{Second proof of AutoGD convergence}

An alternate proof of the convergence of AutoGD, using 
slightly different assumptions and a different proof technique, is presented below.
We assume that we shrink by a factor of $c$ if we detect a function increase 
at the smallest learning rate in the grid.

\bassum 
\label{assum:gamma_bar}
For all $x\in\reals^d$ with $\|\nabla f(x)\| \neq 0$, there exists $\sbgamma(x)$ such that 
$f(x - \gamma \grad f(x)) > f(x)$ if and only if $\gamma > \sbgamma(x)$.
\eassum

For $L$-smooth $f$, define $\slgamma = \frac{2(C-1)}{L C^2}$.
Recall the definition of $G$ given by \cref{eq:G}, and that for 
convenience we assume (without loss of generality) that $\min_x f(x) = 0$.

\bassum 
\label{assum:epsilon_decrease}
There exists $\epsilon \in (0, 1]$ such that for all 
$x\in\reals^d$ with $\|\nabla f(x)\| \neq 0$ 
and $\gamma \in [\slgamma, \sbgamma(x)]$, we have
$c\gamma G(x,c\gamma) \leq 2(1-\epsilon)$.
\eassum

\bassum 
\label{assum:G_lower_bound}
There exists $\slG > 0$ such that for all $x \in \reals^d$ with $\|\nabla f(x)\| \neq 0$, 
we have $G(x, \sbgamma(x)) \geq \slG$.
\eassum

\cref{assum:epsilon_decrease,assum:G_lower_bound} are technical, but are satisfied 
for reasonable quadratic functions. In particular,
consider an arbitrary quadratic function $f(x) = x^T A x$ where $A$ is positive definite. 
The condition number of this function is $\kappa = \lambda_\text{max}(A) / \lambda_\text{min}(A)$, 
where $\lambda_\text{min}, \lambda_\text{max}$ are the minimum and maximum eigenvalues.
In this case, $L = \lambda_\text{max}(A)$ and $\lambda_\text{min}(A) = \mu$.
We show that in this case, \cref{assum:gamma_bar} holds with 
$\sbgamma = x^T A^2 x/(x^T A^3 x)$.
Also, \cref{assum:epsilon_decrease} holds with any 
$\eps \leq 1-2c$, which is independent of $\kappa$,
and \cref{assum:G_lower_bound} holds with $\slG = 4 \lambda_\text{min}(A) = 4 \mu$. 
Therefore, the main restriction is that $c < 1/2$, to allow for $\eps > 0$.

To see this, note that 
\[
  f(x - \gamma \nabla f(x))
  = x^T (I - 2\gamma A)^T A (I - 2\gamma A) x
  = x^T (I - 2\gamma A) A (I - 2\gamma A) x,
\]
and so, for $x \neq 0$, we have 
$f(x - \gamma \nabla f(x)) - f(x) > 0$ if and only if 
$\gamma > \bar\gamma(x)$, where 
\[
  \bar\gamma(x) = \frac{x^T A^2 x}{x^T A^3 x}.
\]
Now, because $\nabla f(x) = 2Ax$ and $\nabla^2 f = 2A$, we have, for $x \neq 0$,
\[
  G(x,\gamma) 
  = \frac{16 x^T A^3 x}{4 x^T A^2 x} 
  = \frac{4 x^T A^3 x}{x^T A^2 x} 
  = \frac{4}{\bar\gamma(x)}.
\]
Therefore, for any $x \in \reals^d$ ($x \neq 0$) and $\gamma \in [\slgamma, \sbgamma(x)]$, we have 
\[
  c \gamma G(x, c\gamma) 
  \leq c \sbgamma(x) \cdot \frac{4}{\sbgamma(x)} 
  = 4c.
\]
Thus, condition \cref{assum:epsilon_decrease} is guaranteed to hold if $\eps \leq 1-2c$.
Further, for \cref{assum:G_lower_bound}, note that 
\[
  \inf_{x \in \reals^d} G(x, \bar\gamma(x)) 
  = \inf_{x \in \reals^d} \frac{4 x^T A^3 x}{x^T A^2 x} 
  = \inf_{y \in \reals^d : \|y\| = 1} 4 y^T A y 
  \geq 4 \lambda_\text{min}(A) 
  = 4 \mu.
\]

\bthm
\label{result:autogd_convergence1}
Suppose $f$ is $L$-smooth, $\mu$-Polyak-\L{}ojasiewicz, twice continuously differentiable, and that 
\cref{assum:gamma_bar,assum:epsilon_decrease,assum:G_lower_bound} hold.
Let
\[
\tau  &= \lt\lceil \log_{c^{-1}} \frac{LC}{\slG}\rt\rceil + 
  \lt\lceil -\log_{C}(c)\rt\rceil + 1 \\
t_0 &= \max\lt\{0,
  \lt\lceil \log_C \frac{2(C-1)}{L C^2 \gamma_0}\rt\rceil, 
  \lt\lceil \log_{c^{-1}}\frac{L\gamma_0}{2}\rt\rceil + \max\lt\{0, \lt\lceil -1-\log_{C}(c)\rt\rceil \rt\}
\rt\}
\]
Then
\[
f(x_t) \leq f(x_0) (1 - 2c\slgamma\epsilon \mu)^{\lfloor \frac{t-t_0}{\tau} \rfloor}.
\]
\ethm

\bprfof{\cref{result:autogd_convergence1}}
As before, since $f$ is $\mu$-P\L{}, if $\|\nabla f(x)\| = 0$, we have $f(x) = 0$ 
and the bound trivially holds. It therefore suffices to check the bound for 
sequences $x_t$ where $\|\nabla f(x_t)\| \neq 0$. 

By the Taylor remainder theorem, for any $x, x' \in \reals^d$,
\[
f(x') = f(x) + \grad f(x)^T(x' - x) + (x'-x)^T \int_0^1 (1-t) \grad^2 f((1-t)x+tx')\d t (x'-x).
\]
If $x' = x - \gamma \grad f(x)$ for some $\gamma > 0$, then
\[
f(x') = f(x) - \gamma \|\grad f(x)\|^2 + \gamma^2 \grad f(x)^T \int_0^1 (1-t) \grad^2 f(x - t\gamma \grad f(x))\d t \grad f(x),
\]
and so 
\[
\label{eq:taylor}
f(x') = f(x) - \|\grad f(x)\|^2\lt(\gamma -  \frac{\gamma^2}{2} G(x, \gamma)\rt). 
\]
Suppose $\|\nabla f(x)\| \neq 0$. 
Note that by continuity of $f$ and \cref{assum:gamma_bar}, we have 
\[
f(x - \sbgamma(x) \nabla f(x)) = f(x).
\]
From \cref{eq:taylor}, this must mean that
if $\|\nabla f(x)\| \neq 0$ and $\sbgamma(x) \neq 0$, we have 
\[
\label{eq:gamma_G_relation}
  \sbgamma(x) = \frac{2}{G(x, \sbgamma)}.
\]
This property will come in useful in the remainder of the proof.
In fact, at any point where $\|\nabla f(x)\| \neq 0$, it must necessarily also be 
the case that $\sbgamma(x) \neq 0$ by differentiability of $f$.
Therefore, it suffices to check that $\|\nabla f(x)\| \neq 0$ in order to use
\cref{eq:gamma_G_relation}.
Since $\sbgamma(x)$ is undefined when $\|\nabla f(x)\| = 0$, in what follows we set 
$\sbgamma(x) = \infty$, forcing \cref{assum:gamma_bar} to hold in this case as well.

Suppose the state of the optimizer is $x_t$ with learning rate $\gamma_t$. Denote the three potential next states
\[
\sbx_{t+1} &= x_t - C \gamma_t \grad f(x_t) \quad \stx_{t+1} = x_t - \gamma_t \grad f(x_t) \quad  \slx_{t+1} = x_t - c \gamma_t \grad f(x_t).
\]
We split the proof into three cases: 
$\gamma_t$ too large ($\gamma_t > \sbgamma(x_t)$), 
$\gamma_t$ too small ($\gamma_t < \slgamma$), 
and $\gamma_t$ within bounds.
If $\gamma_t$ is too small, we show that $\gamma_{t+1} = C\gamma_t$ and $f(x_{t+1}) \leq f(x_t)$.
If $\gamma_t$ is too large, we show that $\gamma_{t+1} = c\gamma_t$ and $f(x_{t+1}) \leq f(x_t)$.
If $\gamma_t$ is within bounds, we show that $f(x_{t+1}) \leq a f(x_t)$ for some 
$a \in [0,1)$, and that $\gamma_{t+1}$ is not too far from being within bounds.

\textbf{$\gamma_t$ too small $(\gamma_t < \slgamma$):}
Note that by $L$-smoothness, $\sup_{x,\gamma}|G(x,\gamma)| \leq L$.
Substituting the three learning rates into \cref{eq:taylor},
\[
f(\sbx_{t+1}) &\leq f(x_t) - \|\grad f(x_t)\|^2\lt(C\gamma - \frac{C^2\gamma^2L}{2}\rt)\\
f(\stx_{t+1}) &\geq f(x_t) - \gamma \|\grad f(x_t)\|^2\\
f(\slx_{t+1}) &\geq f(x_t) - c\gamma \|\grad f(x_t)\|^2.
\]
Therefore, the minimum is guaranteed to occur at $\sbx_{t+1}$ with learning rate $C \gamma_t$ if
\[
C - \frac{C^2\gamma L}{2} > 1 \iff  \gamma < \slgamma = \frac{2(C - 1)}{L C^2}.
\]
In this case, using the fact that $f$ is $\mu$-P\L{},
\[
f(x_{t+1})  
&= f(\sbx_{t+1}) \\
&\leq f(x_t) - \|\grad f(x_t)\|^2\lt(C\gamma - \frac{C^2\gamma^2L}{2}\rt)\\
&\leq f(x_t) \lt(1 - 2\mu\lt(C\gamma - \frac{C^2\gamma^2L}{2}\rt)\rt).
\]

Note that if $\gamma_t < \slgamma$, using the relation from \cref{eq:gamma_G_relation}, 
we have
\[
\gamma_{t+1} 
= C\gamma_t 
< C\slgamma 
= \frac{2C(C-1)}{L C^2} 
\leq \frac{2}{L} 
\leq \inf_{\gamma > 0} \frac{2}{G(x_{t+1}, \gamma)} 
\leq \sbgamma(x_{t+1}),
\]
and hence $\gamma_{t+1}$ will never exceed $\sbgamma(x_{t+1})$ when 
$\gamma_t$ increases from below $\slgamma$.

It takes 
\[
k 
\leq \lt\lceil \log_C \frac{\slgamma}{\gamma_t}\rt\rceil 
= \lt\lceil \log_C \frac{2(C-1)}{L C^2 \gamma_t}\rt\rceil
\]
iterations until $C^k\gamma_t \geq \slgamma$.

\textbf{$\gamma_t$ too large ($\gamma_t > \sbgamma(x_t)$):}
By assumption, we have that $f(x - \gamma \grad f(x)) > f(x)$ iff $\gamma > \sbgamma(x)$.
Therefore if $\gamma_t > \sbgamma(x_t)$, then $C\gamma_t > \sbgamma(x_t)$ as well.
There are two cases to consider: $c\gamma_t > \sbgamma(x_t)$ and $c\gamma_t \leq \sbgamma(x_t)$.
If $c\gamma_t > \sbgamma(x_t)$, then all three choices increase the objective, and so the algorithm stands still and
shrinks the learning rate: $x_{t+1} = x_t$ and $\gamma_{t+1} = c\gamma_t$.
If $c\gamma_t \leq \sbgamma(x_t)$, then the minimum occurs at $\slx_{t+1}$, and $f(x_{t+1}) \leq f(x_t)$ with $\gamma_{t+1} = c\gamma_t$.

Since $G(x,\sbgamma(x)) \leq L$, and either 
$\sbgamma(x) = \frac{2}{G(x,\sbgamma(x))}$ or $\sbgamma(x) = \infty$, 
we have that $\sbgamma(x) \geq 2/L$. Therefore, it takes 
\[
k \leq \lt\lceil \log_{c^{-1}}\frac{L\gamma_t}{2}\rt\rceil
\]
iterations until $c^k \gamma_t \leq \sbgamma(x_{t+k})$.
Note that even if $c^{k-1}\gamma_t > \sbgamma(x_{t+k-1})$,
it is still possible to have $c^k\gamma_t < \slgamma$, in which case we need to increase 
the learning rate before $\gamma_t \in [\slgamma, \sbgamma(x)]$.
Since $\sbgamma(x) \geq 2/L$, this takes 
\[
k 
\leq \max\lt\{0, \lt\lceil \log_{C}\frac{\slgamma}{{2c/L}}\rt\rceil \rt\} 
= \max\lt\{0, \lt\lceil \log_{C}\frac{(C-1)}{c C^2}\rt\rceil \rt\}
\leq \max\lt\{0, \lt\lceil -1-\log_{C}(c)\rt\rceil \rt\}.
\]

\textbf{$\gamma_t$ within bounds ($\slgamma \leq \gamma_t \leq \sbgamma(x_t)$):}
Suppose $\gamma_t \in \lt[\slgamma, \sbgamma(x_t)\rt]$.
By \cref{eq:taylor},
\[
f(x_{t+1}) 
\leq f(\slx_{t+1}) 
= f(x_t) - c\gamma_t\|\grad f(x_t)\|^2\lt(1 - \frac{c\gamma_tG(x_t, c\gamma_t)}{2}\rt).
\]
By \cref{assum:epsilon_decrease}, and the fact that $f$ is $\mu$-P\L{}, if $\|\nabla f(x)\| \neq 0$,
\[
f(x_{t+1}) 
&\leq f(x_t) - c\gamma_t\epsilon\|\grad f(x_t)\|^2 \\
&\leq f(x_t) - c\slgamma\epsilon\|\grad f(x_t)\|^2 \\
&\leq f(x_t) - c\slgamma\epsilon 2\mu f(x_t) \\
&= f(x_t) (1 - 2c\slgamma\epsilon \mu).
\]
That is, $f(x_{t+1}) \leq a f(x_t)$ with $0 \leq a = 1-2c\slgamma \epsilon \mu < 1$.
(We know $a \geq 0$ when $f$ is not identically equal to zero everywhere, or else $\min_x f(x) < 0$, for a contradiction.)
Otherwise, if $\|\nabla f(x)\| = 0$, since $f$ is $\mu$-P\L{}, we know that 
$f(x_{t+1}) = f(x_t) = 0$ and so $f(x_{t+1}) \leq a f(x_t)$ in this case, as well. 
Further, note that
\[
\gamma_{t+1} \geq c\gamma_t \geq c\slgamma
\]
and
\[
\gamma_{t+1} 
\leq C\gamma_t 
\leq C \sbgamma(x_t) 
= \frac{2C}{G(x_t, \sbgamma(x_t))} 
\leq \frac{2C}{\slG}.
\]
Therefore, using the earlier bounds on the number of iterations in the too large 
and too small cases, it takes
\[
k 
&\leq \max\lt\{ \lt\lceil\log_C \frac{2(C-1)}{L C^2 c\slgamma}\rt\rceil, 
  \lt\lceil \log_{c^{-1}} \frac{2LC}{2\slG}\rt\rceil + 
    \max\lt\{0, \lt\lceil -1-\log_{C}(c)\rt\rceil \rt\} \rt\} \\
&= \max\lt\{ \lceil -\log_C(c)\rceil, \lt\lceil \log_{c^{-1}} \frac{LC}{\slG}\rt\rceil 
  + \max\lt\{0, \lt\lceil -1-\log_{C}(c)\rt\rceil \rt\} \rt\} \\
&= \lt\lceil \log_{c^{-1}} \frac{LC}{\slG}\rt\rceil + \lt\lceil -\log_{C}(c)\rt\rceil 
\]
iterations to return to $\gamma_t \in \lt[\slgamma, \sbgamma(x_t)\rt]$.
Combining the behaviour in the three regions yields the stated result.
\eprfof

\clearpage
\allowdisplaybreaks
\section{AutoSGD implementation details and theory} 
\label{sec:supp_autosgd}

\subsection{Online test statistic calculation and decision process} \label{subsec:onlineteststat}

Within episode $t$,
the sequence of test statistics $\slZ_{t,k}, Z_{t,k}, \sbZ_{t,k}$
can be computed online without storing the traces 
$(x_{t,k})_{k=0}^{\tau_t}$. Recall that
\[
  Z_{t,k}
  = \frac{\sum_{j=0}^k \frac{1}{2} \left(\Delta_{t,j}^{(1)} + \Delta_{t,j}^{(2)}\right) }
    {\sqrt{\sum_{j=0}^k \max\left(\eps, \frac{1}{2} \left(\Delta_{t,j}^{(1)} - \Delta_{t,j}^{(2)}\right)^2\right)}}.
\]
It suffices to store only two floating point numbers for
$\sum_{j=0}^k \frac{1}{2} \left(\Delta_{t,j}^{(1)} + \Delta_{t,j}^{(2)}\right)$
and
$\sum_{j=0}^k \max\left(\eps, \frac{1}{2} \left(\Delta_{t,j}^{(1)} - \Delta_{t,j}^{(2)}\right)^2\right)$.
These two quantities can be updated online with rolling sums.
Our proposed AutoSGD decision process is given in \cref{alg:autosgd_decisions}.
For the description of the decision algorithm, we define $Z_{t,k}^{(\gamma)}$ 
to be the statistic among $\slZ_{t,k}, Z_{t,k}, \sbZ_{t,k}$ corresponding to 
the stream with learning rate $\gamma$. 

\begin{algorithm}
	\begin{algorithmic}[1]
    \Require Streams $(\slx_{t,j})_{j=0}^k, (x_{t,j})_{j=0}^k, (\sbx_{t,j})_{j=0}^k$ 
    with learning rates in $E_t := \{c \gamma_t, \gamma_t, C \gamma_t\}$,
    statistic threshold $z^\star$ (default: 1.96), 
    minimum number of samples $M$ (default: 30).
    \If{$k+1 < M$}  \Comment{not enough samples}
      \State $I_t, D_t, S_t, R_t \gets 0$
    \Else
      \State $A \gets \{\gamma \in E_t : Z^{(\gamma)}_{t,k} > z^\star, 
        Z_{t,k}^{(\gamma')} \geq -z^\star \text{ for any $\gamma' < \gamma$, $\gamma' \in E_t$}\}$
      \If{$A \neq \emptyset$}  \Comment{significant objective function reduction} 
        \State $\gamma' \gets \max A$
        \If{$\gamma' = c\gamma_t$} \Comment{decrease the learning rate} 
          \State $I_t, D_t, S_t, R_t \gets 0, 1, 0, 0$ 
        \ElsIf{$\gamma' = \gamma_t$} \Comment{stay at the current learning rate} 
          \State $I_t, D_t, S_t, R_t \gets 0, 0, 1, 0$ 
        \Else \Comment{increase the learning rate} 
          \State $I_t, D_t, S_t, R_t \gets 1, 0, 0, 0$
        \EndIf
      \Else 
        \State $A \gets \{\gamma \in \{c\gamma_t, \gamma_t\} : Z_{t,k}^{(\gamma)} \geq -z^\star, 
          Z_{t,k}^{(\gamma')} \geq -z^\star \text{ for any $\gamma' < \gamma$}, \gamma' \in \{c\gamma_t, \gamma_t\}\}$
        \If{$A \neq \emptyset$}
          \State $\gamma' \gets \max A$
          \If{$\gamma' = c\gamma_t$} \Comment{decrease the learning rate} 
            \State $I_t, D_t, S_t, R_t \gets 0, 1, 0, 0$ 
          \Else \Comment{stay at the current learning rate} 
            \State $I_t, D_t, S_t, R_t \gets 0, 0, 1, 0$ 
          \EndIf
        \Else  \Comment{evidence of objective increase at all learning rates}
          \State $I_t, D_t, S_t, R_t \gets 0, 0, 0, 1$ 
        \EndIf
      \EndIf
    \EndIf
    \State \Return $I_t, D_t, S_t, R_t$ 
	\end{algorithmic}
  \caption{\texttt{decision()}}
  \label{alg:autosgd_decisions}
\end{algorithm}

%%%%%%%%%%%%%%%%%%%%%%%%%%%%%%%%%%%%%%%%%%%%%%%%%%%%%%%%%%%%%%%%%%%%%%%%%%%%%%%%

\subsection{Stochastic convergence theory}
\label{sec:autosgd_proof} 

\cref{thm:appendixfinalconvergence} presents the main AutoSGD convergence result and its proof.
Note that \cref{thm:appendixfinalconvergence} is just a more specific version of \cref{result:autosgd_convergence}
from the main text with all the constants and value constraints specified.

The proof of convergence of AutoSGD relies on the following high-level strategy with 3 steps.
\begin{enumerate}[label=Step \arabic*:]
  \item We first consider the ``not bad \emph{sub}sequence'' of episode iterations $x_t,\gamma_t$ such that
  $a \gamma_t \leq f(x_t)$ (for some proportionality constant $a > 0$). At these episodes,
  the step size is small enough that the inner SGD streams should result in an 
  expected descent $\E\lt[f(x_{t+1}) | x_t,\gamma_t\rt] \leq f(x_t)$ (though it may be a vanishingly small amount of descent).
  In between these iterations, the algorithm can take excursions into the ``bad'' region
  where $a\gamma_t > f(x_t)$ (the step size is too large) for some random number of episodes. \cref{lem:tourdescent} shows that these excursions
  are not \emph{too} bad, and only result in at most a small increase after the initial one-step descent $\E\lt[f(x_{t+1}) | x_t,\gamma_t\rt] \leq f(x_t)$.

  \item We next consider the ``good \emph{subsub}sequence'' of episode iterations $x_t, \gamma_t$ such that
  $\ell f(x_t) \leq a\gamma_t \leq f(x_t)$ (for some other constant $\ell>0$). In these episodes,
  the step size is both small enough to guarantee descent, and large enough to guarantee \emph{non-negligible} descent.
  \cref{lem:tourtime} uses a stochastic Lyapunov function technique (i.e., a drift function, in the terminology of the 
  MCMC literature \cite{douc2018markov})
  to bound the amount of time on excursions away from this set.

  \item Finally, \cref{thm:abstractconvergence} puts these two results together.
  If we assume that the one-step descent in the ``not bad'' region where $a\gamma_t \leq f(x_t)$
  satisfies $\E\lt[f(x_{t+1}) | x_t,\gamma_t\rt] \leq (1-\eta \frac{a\gamma_t}{f(x_t)})f(x_t)$ (i.e., vanishing descent as $\gamma_t \to 0$),
  then the overall descent on the ``not bad subsequence'' of episodes forms a nonnegative supermartingale.
  At that point, the optional stopping theorem shows that the ``good subsubsequence'' exhibits geometric descent,
  and the drift function bounds from \cref{lem:tourtime} guarantee that the good subsubsequence occurs regularly in the original episode sequence.
  The remainder of this section after presenting \cref{lem:tourdescent,lem:tourtime,thm:abstractconvergence} is devoted
  to posing specific assumptions about the decision process (that determines $I_t,S_t,D_t,R_t,\tau_t$) 
  to enable verification of the conditions of these three abstract results for AutoSGD, resulting in \cref{thm:appendixfinalconvergence}.
\end{enumerate}

Note that the proofs of all results in this section are admittedly rather technical and not very illuminating; to clarify the exposition, we will simply present and
discuss results here, and defer the proofs to \cref{sec:stochasticproofs}.

We begin with \cref{lem:tourdescent}, which analyzes the expected change in $f$ 
on the ``not bad subsequence'' in between excursions to the ``bad'' region. Here $a$ is the aforementioned proportionality constant for the ``not bad'' region.
The constant $b>0$ characterizes the amount of increase of $f$ during excursions into the ``bad'' region; in AutoSGD, this will be
proportional to the probability of $R_t$ \emph{not} triggering in the bad region (which should be very small). The proportionality to $\gamma_t$ in
terms with $b$ is crucial for later showing that the ``not bad subsequence'' forms a nonnegative supermartingale; in particular, the closer $\gamma_t$ is 
to 0 (the farther from the bad region it is), the less increase we expect, because the episode sequence is most likely to simply stay in the ``not bad'' region.
The constant $v\in[0,1)$ characterizes the expected decrease of the step size while in the bad region; in AutoSGD, $v$ should be just slightly greater than $c < 1$, since 
$R_t$ is very likely in that region (at which point $\gamma_{t+1} \gets c \gamma_t$).

\blem\label{lem:tourdescent}
Consider a discrete-time Markov chain $(x_t,\gamma_t)$, $t\in\nats_0$ on $\reals^{d+1}$ and a nonnegative function $f : \reals^d\to\reals$, $f\geq 0$.
Suppose there exist constants $a, b, C \geq 0$ and $v\in[0,1)$ such that 
for all $t\in\nats_0$, we have $\gamma_{t+1}\leq C\gamma_t$ \as and
\[
a\gamma_t > f(x_t) \implies \E\lt[\gamma_{t+1}|x_t,\gamma_t\rt] &\leq v \gamma_t \quad\text{and}\quad\E\lt[f(x_{t+1})|x_t,\gamma_t\rt] \leq f(x_t) + b\gamma_t.
\]
Fix a particular $t\in\nats_0$, and let $\tau\in\nats$ be the smallest index such that $\tau > t$ and $a\gamma_\tau \leq f(x_\tau)$, or $\infty$ if no such index exists.
Then
\[
\E\lt[f(x_{\tau}) | x_{t},\gamma_{t}\rt] &\leq \E\lt[f(x_{t+1})|x_{t},\gamma_{t}\rt] + \gamma_t\frac{C b}{1-v},
\]
where $f(x_\tau) = \lim_{k\to\infty} f(x_{\tau\wedge k})$ 
and $\E\lt[f(x_{\tau}) | x_{t},\gamma_{t}\rt] = \lim_{k\to\infty}\E\lt[f(x_{\tau \wedge k}) | x_{t},\gamma_{t}\rt]$ are well-defined random variables.
\elem

Next, given an appropriate drift function $V$, \cref{lem:tourtime} characterizes the amount of time spent
away from the region where $V \leq 1$. 
Note that \cref{lem:tourtime} is not novel, and similar results appear commonly in the MCMC convergence literature
in the analysis of geometric ergodicity \cite{douc2018markov}.
Later in \cref{thm:abstractconvergence}, we will assume that 
$V$ is designed such that the region where $V\leq 1$ corresponds to the region where $\ell f(x_t) \leq a\gamma_t \leq f(x_t)$
(i.e., the ``good'' region where sufficient decrease occurs). 
In AutoSGD, the constant $\xi$ will be related to the probability of $R_t$ in the ``bad'' region $a\gamma_t > f(x_t)$, and the probability 
of $I_t$ when $\gamma_t$ is far too small; in other words, how likely the algorithm is to shrink the step size when it is too large,
and how likely it is to grow it when it is too small. The constant $\zeta$ can take any finite value, and is just a technical constant
in the application to AutoSGD below.

\blem\label{lem:tourtime}
Given a discrete-time Markov chain $x_t,\gamma_t$, $t\in\nats_0$ on $\reals^{d+1}$,
suppose there exist constants $\zeta\geq 0$, $\xi\in[0,1)$ and a measurable function $V:\reals^{d+1} \to \reals$ such that
\[
\E\lt[V(x_{t+1},\gamma_{t+1}) | x_t,\gamma_t\rt] &\leq \xi V(x_t,\gamma_t) + \zeta\1[V(x_t,\gamma_t) \leq 1] \quad a.s.
\]
Fix any $t\in\nats_0$, and let $\tau\in\nats$ be the smallest index such that $\tau > t$ and $V(x_\tau,\gamma_\tau) \leq 1$.
Then
\[
\forall k\in\nats, \qquad \P(\tau > k | x_t,\gamma_t) \leq \xi^{k-1}\lt(\xi V(x_t,\gamma_t) + \zeta\1[V(x_t,\gamma_t)\leq 1]\rt) \quad a.s.
\]
\elem

Finally, \cref{thm:abstractconvergence} makes some slight additional assumptions on the one-step descent of $f$ and
the design of the drift function $V$ to put the two above lemmas together for the main abstract result about convergence.
Here $\ell$ is the earlier-mentioned constant for the lower bound on the ``good'' region. We will see below (in \cref{lem:innerdescent}) that
due to the use of SGD for inner iterations, $\E\lt[f(x_{t+1}) | x_t,\gamma_t\rt]$ is always a mixture of $f(x_t)$ and a quantity proportional to $\gamma_t$,
and so the constant $B$ captures that proportionality constant. Finally, $\eta$ captures the one-step descent of $f$ 
in the ``not bad'' region in a manner that vanishes as $\gamma_t \to 0$.

\bthm\label{thm:abstractconvergence}
Consider a discrete-time Markov chain $x_t,\gamma_t$, $t\in\nats_0$ on $\reals^{d+1}$,
functions $f : \reals^d\to\reals$ and $V:\reals^{d+1}\to\reals$, and constants $a,b,C,v,\zeta,\xi$
all satisfying the conditions of \cref{lem:tourdescent,lem:tourtime}.
Suppose additionally that there exist constants $B,\ell > 0$, $\eta \in (0, 1]$ such that
\[
\begin{aligned}
  &\E\lt[f(x_{t+1})|x_t,\gamma_t\rt] \leq \max\{f(x_t), B\gamma_t\} \\ 
  &V(x_t,\gamma_t)\leq 1 \implies \ell \leq \frac{a\gamma_t}{f(x_t)} \leq 1 \\
  &a\gamma_t\leq f(x_t) \implies \E\lt[f(x_{t+1})|x_t,\gamma_t\rt] \leq f(x_t)\lt(1-\eta\frac{a\gamma_t}{f(x_t)}\rt) \\
  &\eta - \frac{Cb}{a(1-v)} > 0.
\end{aligned}
\label{eq:thmassump}
\]
Let $t_n$ be the $n^\text{th}$ index $t$ at which $V(x_t, \gamma_t) \leq 1$.
Then
\[
\forall n\in\nats, \quad \E\lt[f(x_{t_{n}})\rt] &\leq \nu^{n-1} f_0 \quad\text{and}\quad \E[t_n] \leq \frac{V_0}{1-\xi} + (n-1)\frac{\xi+\zeta}{1-\xi},
\]
where
\[
\nu &= \lt(1-\ell\lt(\eta - \frac{Cb}{a(1-v)}\rt)\rt) < 1\\
f_0 &= \max\{f(x_0), B\gamma_0\}+ \gamma_{0}\frac{Cb}{1-v} 
\qquad V_0 = \xi V(x_0,\gamma_0) + \zeta\1[V(x_0,\gamma_0)\leq 1].
\]
\ethm

The remainder of this section is devoted to applying \cref{lem:tourdescent,lem:tourtime,thm:abstractconvergence} to the 
special case of AutoSGD. 
We begin with technical assumptions \cref{assum:f,assum:testpreferlower,assum:irprobs,assum:sufficientdecrease,assum:boundeddecrease}, using the notation and basic setup established in the main text. 
Note that while $I_t,S_t,D_t,R_t$ are binary indicator variables for the various decision events in episode $t$,
we use a slight abuse of notation below in that the indicators are treated
also as events in probability statements; for example, $\P(I_t)$ is used as a shorthand for $\P(I_t=1)$.
We use these assumptions to prove  \cref{lem:innerdescent,lem:outerdescent,lem:outerlyapunov},
and finally conclude the section with the proof of \cref{thm:appendixfinalconvergence} (the more detailed version of \cref{result:autosgd_convergence} from the main text), which amounts to translating
the constants from \cref{assum:f,assum:testpreferlower,assum:irprobs,assum:sufficientdecrease,assum:boundeddecrease} into the
constants needed for \cref{lem:tourdescent,lem:tourtime,thm:abstractconvergence}.

Note that the below set of assumptions and theoretical analysis are designed to illustrate the  
properties necesssary for a \emph{general} decision process to yield convergent AutoSGD. 
We do not attempt to apply these results to our specific recommended decision process from \cref{subsec:onlineteststat}
in the interest of avoiding significant amounts of uninteresting computation, although that decision process is 
designed to intuitively follow the below desiderata.

\cref{assum:f} stipulates some basic properties of $f$. The technical condition in the display is satisfied in fairly general settings,
e.g., when $f$ has Lipschitz gradients and $g(x,u)$ has a variance bound proportional to $\|\grad f(x)\|^2 + b$.
\bassum
\label{assum:f}
$f:\reals^d\to\reals$ is twice continuously differentiable, $\mu$-P\L{}, attains its minimum with value $\min_{x\in\reals^d}f(x) = 0$,
and there exist $\alpha,\beta \geq 0$ such that $\forall 0\leq\gamma \leq \alpha^{-1}$,
\[
\lt| \E\lt[g(x,u)^T \int_0^1 2(1-s) \grad^2 f(x - s \gamma g(x,u))\d s  g(x,u)\rt]\rt| \leq \alpha \|\grad f(x)\|^2 + \beta, \quad u\sim \phi.
\]
\eassum

\cref{assum:testpreferlower} captures the desideratum that the decision process tends to prefer lower objective values.
For example, in particular, it says that $I_t \1[\tau_t = k]$ is negatively correlated with $f(\sbx_{t,k})$. In other words,
it says that on events where $I_t$ triggers at inner iteration $\tau_t = k$, $f(\sbx_{t,k})$ tends to be smaller than it 
would be absent any decision process occurring. We expect that this assumption is satisfied  by reasonable decision processes.
\bassum\label{assum:testpreferlower}
For all $k\in\nats$,
\[
\begin{aligned}
\E\lt[I_t\1[\tau_t = k]f(\sbx_{t,k}) |x_t,\gamma_t\rt] &\leq \P(I_t, \tau_t=k | x_t,\gamma_t) \E[f(\sbx_{t,k})|x_t,\gamma_t]\\
\E\lt[S_t\1[\tau_t = k]f(x_{t,k}) |x_t,\gamma_t\rt] &\leq \P(S_t, \tau_t=k | x_t,\gamma_t) \E[f(x_{t,k})|x_t,\gamma_t]\\
\E\lt[D_t\1[\tau_t = k]f(\slx_{t,k}) |x_t,\gamma_t\rt] &\leq \P(D_t, \tau_t=k | x_t,\gamma_t) \E[f(\slx_{t,k})|x_t,\gamma_t].
\end{aligned}\label{eq:isdassump}
\]
\eassum
\cref{assum:irprobs} captures the need for the decision process to reset and
decrease the step size when it is too large ($R_t$), and to increase the step
size when it is too small ($I_t$). In particular, the decision process should
have a very small probability $\delta$ of making the wrong decision in either
case. It also guarantees that $\gamma_t \leq \alpha^{-1}$ \as

This assumption, although not particularly intricate, has one clear departure from the
actual decision process we use. In particular, the condition that $\forall
\gamma_t > (C\alpha)^{-1}$, $\P(I_t|x_t,\gamma_t) = 0$ is not one we can
realistically enforce in practice, because we do not know the value of the
smoothness constant $\alpha>0$. However, this is equivalent to asking for a
hard upper bound on the step size $\gamma_t$ proportional to an inverse
Lipschitz constant, which is a very common assumption in SGD analysis \cite{garrigos2023handbook}. 
In practice, we expect that this departure is not a major issue;
the probability of $I_t$ in the case where the objective function
is growing is not exactly zero, but should be close enough to zero (and decaying as the objective function increases)
so as to not modify the below results meaningfully.
\bassum\label{assum:irprobs}
\cref{assum:f} holds,$\gamma_0 \leq \alpha^{-1}$,
and there exist constants $\delta,\ell \in (0,1)$ such that 
\[
\frac{C\gamma_t\beta}{2\mu} > f(x_t) \implies \P(R_t|x_t,\gamma_t) \geq 1-\delta, \qquad
\frac{C\gamma_t\beta}{2\mu} \leq f(x_t) \implies \P(R_t|x_t,\gamma_t) \leq \delta,
\]
and
\[
\gamma_t > (C\alpha)^{-1} \implies \P(I_t|x_t,\gamma_t) = 0, \qquad \gamma_t\leq (C\alpha)^{-1}\text{ and }\frac{C\gamma_t\beta}{2\ell\mu} < f(x_t) \implies \P(I_t|x_t,\gamma_t) \geq 1-\delta.
\]
\eassum
\cref{assum:sufficientdecrease} below has three parts: \cref{eq:gamassump,eq:tauassump1,eq:tauassump2}.
\cref{eq:gamassump} requires that we are likely to pick a step size that results in a non-negligible descent,
which again is reasonably expected to be satisfied by the decision process in \cref{subsec:onlineteststat} in practice.
\cref{eq:tauassump1} enforces that when the step size $\gamma_t$ is small enough (in the ``not bad'' region mentioned earlier),
the decision process takes \emph{at least} an amount of time roughly proportional to $1/f(x_t)$ to make a decision.
This should be satisfied for some small enough $u \in (0,1)$ by most reasonable decision processes that are motivated by streaming statistical hypothesis tests;
if $f(x_t)$ is small, the difference between $f(x_t)$ and subsequent descending iterations must also be small (since $f \geq 0$), and so distinguishing which stream
is the ``winner'' should take more samples and hence take longer. Our suggested process in \cref{subsec:onlineteststat} should
adhere to this behaviour. \cref{eq:tauassump2} simply asserts that when the learning rate is close to the upper threshold of $\alpha^{-1}$
and the function value $f$ is also large, that the decision process guarantees reasonable descent; this again should 
be satisfied in practice for some large enough $w$. Note that in the decision process in \cref{subsec:onlineteststat}, we enforce a minimum
number of samples $\tau_t \geq M$, which also incidentally helps enforce \cref{eq:tauassump2}. For example, 
if $M > (\log (c/2)) / \log(1-c\mu/(\alpha C))$, \cref{eq:tauassump2} is automatically satisfied with $w=1/2$.

\bassum\label{assum:sufficientdecrease}
\cref{assum:f} holds
and there exist constants $\epsilon,u,w \in (0,1)$, $u \leq c\mu/C\alpha$
such that
 if $C\gamma_t\beta/2\mu \leq f(x_t)$,
\[
  \frac{\gamma_t\beta}{2\mu}\E[\1[I_t]C + \1[S_t] + \1[D_t]c | x_t,\gamma_t] \leq (1-\delta)\epsilon f(x_t), \label{eq:gamassump}
\]
if $C\gamma_t\beta/2\mu \leq f(x_t)$ and $\gamma_t\leq(C\alpha)^{-1}$,
\[
  &\max\lt\{\E[(1-C\mu\gamma_t)^{\tau_t} | I_t, x_t, \gamma_t],
  \E[(1-\mu\gamma_t)^{\tau_t} | S_t, x_t, \gamma_t],
  \E[(1-c\mu\gamma_t)^{\tau_t} | D_t, x_t, \gamma_t]\rt\} \\
  &{\quad}\leq \lt(1-\frac{u\gamma_tC\beta}{2\mu f(x_t)}\rt),\label{eq:tauassump1}
\]
and if $C\gamma_t\beta/2\mu \leq f(x_t)$ and $(C\alpha)^{-1} < \gamma_t < \alpha^{-1}$,
\[
  &\max\lt\{
  \E[\lt(1-\mu/(\alpha C)\rt)^{\tau_t} | S_t, x_t, \gamma_t],
  \E[\lt(1-c\mu/(\alpha C)\rt)^{\tau_t} | D_t, x_t, \gamma_t]\rt\} \leq w c.\label{eq:tauassump2}
\]

\eassum
The final assumption in \cref{assum:boundeddecrease} is the most technical of the assumptions, but is 
generally quite weak; it simply asserts that it is unlikely that any inner SGD stream terminates at a vanishingly small value of
$f(x_t)$ in a single episode. This assumption is likely to hold for almost any reasonable decision process given
known results about the geometric ergodicity of the uncorrected Langevin algorithm\anonymize{\cite{xx}}.
Unlike the previous assumptions, this assumption is probably not even necessary, and a more careful analysis might be able
to avoid it entirely. In practice, in the off-chance $f(x_t)$ happens to be much smaller than expected,
the algorithm will simply try to shrink the step sizes
as much as possible until either it successfully returns to the set $a\gamma_t \leq f(x_t)$, or it will accidentally not trigger $R_t$ at some point and
immediately $f(x_t)$ will jump larger to a value roughly $\propto \gamma_t$.
\bassum\label{assum:boundeddecrease}
There exists $\sigma \geq 0$ such that for all $t$,
\[
\sum_{(y,A)\in\{(\slx_{t, \tau_t},D_t),(x_{t, \tau_t},S_t),(\sbx_{t, \tau_t},I_t)\}} 
  \int_0^1 s^{-2}\P(f(y) < saC\gamma_t, A | x_t,\gamma_t)\d s 
  \leq \sigma\frac{aC\gamma_t}{f(x_t)} \mathbb{P}(R_t^c | x_t, \gamma_t).
\]
\eassum

We now use these assumptions in \cref{lem:innerdescent,lem:outerdescent,lem:outerlyapunov} to verify the conditions
of \cref{lem:tourdescent,lem:tourtime,thm:abstractconvergence} and obtain specific values for the constants $a,b,v,\xi,\zeta,\ell,\eta,B$
in terms of the constants $\alpha,\beta,\delta,\epsilon,\mu,u,w,\sigma$ appearing in \cref{assum:f,assum:testpreferlower,assum:irprobs,assum:sufficientdecrease,assum:boundeddecrease}.
We begin with \cref{lem:innerdescent}, which is a fairly standard result capturing the expected descent for an individual inner stream of SGD iterates.

\blem\label{lem:innerdescent}
Suppose \cref{assum:f} holds, $0 \leq \gamma \leq \alpha^{-1}$, and $y_0\in\reals^d$.
Consider the iteration $y_{k+1} = y_k - \gamma g(y_k, u_k)$, where $u_k \distiid \phi$.
Then for all $j,k\in\nats_0$, $j\leq k$, and for all $z\in\reals$, $0 < z \leq 2\mu\lt(1-\frac{1}{2}\gamma\alpha\rt)$,
\[
\E\lt[f(y_{k})| y_j\rt]
&\leq f(y_j)(1 - \gamma z)^{k-j} + (1- (1-\gamma z)^{k-j})\frac{\gamma\beta}{2z}.
\]
\elem

\cref{lem:outerdescent} is used to verify the conditions of \cref{lem:tourdescent},
as well as the $f(x_t)(1-\eta a\gamma_t/f(x_t))$ and $\max\{f(x_t),B\gamma_t\}$ bound conditions in \cref{thm:abstractconvergence},
with constant values
\[
a = \frac{C\beta}{2\mu},\quad B=\frac{(1+\delta)C\beta}{2\mu}, \quad b = \frac{\delta C\beta}{2\mu}, \quad \eta = u(1-\delta)(1-\epsilon), \quad v = (1-\delta)c + \delta C.\label{eq:constants1}
\]

\blem\label{lem:outerdescent}
Suppose \cref{assum:f,assum:testpreferlower,assum:irprobs,assum:sufficientdecrease} hold.
Then if $C\gamma_t\beta/2\mu > f(x_t)$,
\[
\E\lt[ f(x_{t+1}) | x_t, \gamma_t\rt] &\leq f(x_t) +\gamma_t\frac{\delta C \beta}{2\mu} \quad\text{and}\quad \E\lt[\gamma_{t+1} | x_t,\gamma_t\rt] \leq \gamma_t((1-\delta)c+\delta C),\label{eq:c2highresult}
\]
and otherwise,
\[
\E\lt[ f(x_{t+1}) | x_t, \gamma_t\rt]  &\leq 
\lt(1 -\frac{u C\beta\gamma_t}{2\mu f(x_t)}(1-\delta)(1-\epsilon)\rt)f(x_t).\label{eq:c2lowresult}
\]
\elem
Let $V:\reals^{d+1} \to \reals$ be defined as
\[
\label{eq:V_definition}
  V(x,\gamma) 
  = \frac{\ell f(x)}{a\gamma}\1\lt[\frac{\ell f(x)}{a \gamma} > 1\rt] 
  + \frac{a\gamma}{ f(x)}\1\lt[\frac{a\gamma}{f(x)}>1\rt],
\]
where $\ell$ is the constant from \cref{assum:irprobs}, and $a = \beta C/2\mu$ as per \cref{eq:constants1}.
The final technical lemma is \cref{lem:outerlyapunov}, which is used to verify the conditions relating to the drift function $V$ in \cref{lem:tourtime,thm:abstractconvergence}
with constant values
\[
\begin{aligned}
\ell &= \text{the }\ell\text{ from \cref{assum:irprobs}}\\
\xi &= \max\lt\{\lt(w + \frac{(1-wc)\ell}{C} + \delta \rt)\vee \lt(\frac{1-\delta}{C} + \frac{\delta}{c}\rt)+\delta c \ell + \sigma C\ell,\ell c^{-1}\delta +c + \sigma\delta C\rt\}\\
\zeta &=  c^{-1} + \delta c + \sigma C.
\end{aligned}\label{eq:constants2}
\]
\blem\label{lem:outerlyapunov}
Suppose \cref{assum:f,assum:testpreferlower,assum:irprobs,assum:sufficientdecrease,assum:boundeddecrease} hold with $\ell \leq c$.
Then
\[
  \E\lt[V(x_{t+1},\gamma_{t+1}) | x_t,\gamma_t\rt] &\leq \xi V(x_t,\gamma_t) + \zeta\1[V(x_t,\gamma_t) \leq 1],
\]
where
\[
\xi &= \max\lt\{\lt(w + \frac{(1-wc)\ell}{C} + \delta \rt)\vee \lt(\frac{1-\delta}{C} + \frac{\delta}{c}\rt)+\delta c \ell + \sigma C\ell,\ell c^{-1}\delta +c + \sigma\delta C\rt\}\\
\zeta &= c^{-1} + \delta c + \sigma C,
\]
and $V$ is given by \cref{eq:V_definition}.
\elem
\cref{thm:appendixfinalconvergence} presents the final convergence result, 
which is just a more specific version of \cref{result:autosgd_convergence}
from the main text with all constants and value constraints specified.
Note that the theorem only holds when $\ell,\delta$ are small enough, but this is not a major requirement.
The constant $\ell$ is an abstract value not involved in any aspect of the AutoSGD algorithm, and so can be taken as large as possible
while still satisfying the bounds in \cref{eq:finalconstraints}.
The constant $\delta$ can be tuned to be as small as needed by making the decision process more conservative (e.g., increasing the 
value of $z^\star$ in \cref{alg:autosgd_decisions}).

\bthm\label{thm:appendixfinalconvergence}
Suppose \cref{assum:f,assum:testpreferlower,assum:irprobs,assum:sufficientdecrease,assum:boundeddecrease} 
hold with $\ell,\delta$ small enough such that
\[
\begin{aligned}
\ell &\leq c\\
(1-\delta)c + \delta C &< 1\\
\max\lt\{\lt(w + \frac{(1-wc)\ell}{C} + \delta \rt)\vee \lt(\frac{1-\delta}{C} + \frac{\delta}{c}\rt)+\delta c \ell + \sigma C\ell,\ell c^{-1}\delta +c + \sigma\delta C\rt\} &< 1\\
\text{and}\quad u(1-\delta)(1-\epsilon) - \frac{C\delta}{\lt(1-(1-\delta)c-\delta C\rt)} &> 0.
\end{aligned}\label{eq:finalconstraints}
\]
Let $t_n$ be the $n^\text{th}$ episode index $t$ at which $\ell f(x_t) \leq \frac{C\beta}{2\mu} \gamma_t \leq f(x_t)$. Then
$\forall n\in\nats$,
\[
\E\lt[f(x_{t_n})\rt] &\leq \nu^{n-1}\lt(\max\lt\{f(x_0),\frac{C\beta (1+\delta)}{2\mu}\gamma_0\rt\} + \gamma_0 \frac{C^2\delta\beta }{2\mu(1-(1-\delta)c-\delta C)}\rt)\\
\E\lt[t_n\rt] &\leq \frac{\xi V(x_0,\gamma_0)+\zeta}{1-\xi} + (n-1)\frac{\xi+\zeta}{1-\xi},
\]
where
\[
\nu &= 1 - \ell\lt(u(1-\delta)(1-\epsilon) - \frac{C\delta}{\lt(1-(1-\delta)c-\delta C\rt)}\rt) < 1\\
\xi &= \max\lt\{\lt(w + \frac{(1-wc)\ell}{C} + \delta \rt)\vee \lt(\frac{1-\delta}{C} + \frac{\delta}{c}\rt)+\delta c \ell + \sigma C\ell,\ell c^{-1}\delta +c + \sigma\delta C\rt\}\\
\zeta &= c^{-1}+\delta c + \sigma C.
\]
\ethm
\bprf
\cref{assum:f,assum:testpreferlower,assum:irprobs,assum:sufficientdecrease,assum:boundeddecrease}
with the constant constraints given by \cref{eq:finalconstraints}
imply that \cref{lem:innerdescent,lem:outerdescent,lem:outerlyapunov} hold. 
These then verify the conditions of
\cref{lem:tourdescent,lem:tourtime,thm:abstractconvergence} with the constants specified in \cref{eq:constants1,eq:constants2}.
(We also use the intermediate bound \cref{eq:f_t_bound_intermediate} from the 
proof of \cref{lem:tourtime}.)
Substitution yields the stated result.
\eprf

Note finally that \cref{thm:appendixfinalconvergence} does not say anything about the total number of SGD iterations;
in fact, none of the assumptions above preclude the possibility that $\tau_t = \infty$ for some $t\in\nats$.
That being said, any reasonable decision process should indeed be designed to satisfy $\tau_t < \infty$ \as to avoid this pathology.
We leave a more detailed analysis of the expected total number of SGD iterations to future work.

\subsection{Stochastic convergence proofs}\label{sec:stochasticproofs}

\bprfof{\cref{lem:tourdescent}}
We have that for $j \ge 1$,
\[
f(x_{\tau\wedge (t+ j)}) = f(x_{t+1}) + \sum_{k=2}^j \1[\tau \geq t+k](f(x_{t+k}) - f(x_{t+k-1})).
\]

Note that since $f\geq 0$, if
\[
\E\lt[\sum_{k=2}^{\infty}\1[\tau \geq t+k] (f(x_{t+k}) + f(x_{t+k-1})) | x_t,\gamma_t\rt] < \infty\quad a.s.,
\]
then
$\sum_{k=2}^{\infty}\1[\tau \geq t+k] (f(x_{t+k}) + f(x_{t+k-1})) < \infty$ \as, which implies
the limit $f(x_\tau) = \lim_{k\to\infty}f(x_{\tau\wedge k})$ is a well-defined random variable,
and the dominated convergence theorem guarantees that 
$\E\lt[f(x_{\tau}) | x_{t},\gamma_{t}\rt] = \lim_{k\to\infty}\E\lt[f(x_{t\wedge k}) | x_{t},\gamma_{t}\rt]$ is similarly well-defined.

We now show that the sum is finite. By the definition of $\tau$ and Fubini's theorem,
\[
&\E\lt[ \sum_{k=2}^{\infty}\1[\tau \geq t+k] (f(x_{t+k}) + f(x_{t+k-1})) | x_{t}, \gamma_{t}\rt] \\
&= \sum_{k=2}^{\infty}\E\lt[\1[\tau \geq t+k] (f(x_{t+k}) + f(x_{t+k-1})) | x_{t}, \gamma_{t}\rt]\\
&= \sum_{k=2}^{\infty}\E\lt[\1[\tau \geq t+k] f(x_{t+k}) | x_{t}, \gamma_{t}\rt] + \E\lt[\1[\tau\geq t+k]f(x_{t+k-1}) | x_{t}, \gamma_{t}\rt]\\
&\leq \sum_{k=2}^{\infty}\E\lt[\1[\tau \geq t+k] \lt(f(x_{t+k-1}) + b \gamma_{t+k-1}\rt) | x_{t}, \gamma_{t}\rt] + \E\lt[\1[\tau\geq t+k]f(x_{t+k-1}) | x_{t}, \gamma_{t}\rt].
\]
Now, since $\tau \geq t+k$, we have $f(x_{t+k-1}) < a\gamma_{t+k-1}$, and so
\[
\E\lt[ \sum_{k=2}^{\infty}\1[\tau \geq t+k] (f(x_{t+k}) + f(x_{t+k-1})) | x_{t}, \gamma_{t}\rt] 
&\leq \lt(2a + b\rt)\sum_{k=2}^{\infty}\E\lt[\1[\tau \geq t+k] \gamma_{t+k-1} | x_{t}, \gamma_{t}\rt].
\]
The sum on the right hand side can be bounded by noting that $\gamma_{t+1} \leq C\gamma_t$ \as, 
extracting the first term, shifting indices, and then bounding the decay in $\gamma_{t+k}$ using the definition of $\tau$
and the law of total expectation:
\[
\sum_{k=2}^{\infty}\E\lt[\1[\tau \geq t+k] \gamma_{t+k-1} | x_{t}, \gamma_{t}\rt]
&\leq C\gamma_{t}\P(\tau > t+1|x_t,\gamma_t) + \sum_{k=3}^{\infty}\E\lt[\1[\tau \geq t+k] \gamma_{t+k-1} | x_{t}, \gamma_{t}\rt]\\
&\leq C\gamma_{t}\P(\tau > t+1|x_t,\gamma_t) + \sum_{k=3}^{\infty}\E\lt[\1[\tau \geq t+k-1] \gamma_{t+k-1} | x_{t}, \gamma_{t}\rt]\\
&= C\gamma_{t}\P(\tau > t+1|x_t,\gamma_t) + \sum_{k=2}^{\infty}\E\lt[\1[\tau \geq t+k] \gamma_{t+k} | x_{t}, \gamma_{t}\rt]\\
&\leq C\gamma_{t}\P(\tau > t+1|x_t,\gamma_t) + v\sum_{k=2}^{\infty}\E\lt[\1[\tau \geq t+k] \gamma_{t+k-1} | x_{t}, \gamma_{t}\rt].
\]
Noting that the sums on the left and right hand sides are the same, we have that
\[
\E\lt[ \sum_{k=2}^{\infty}\1[\tau \geq t+k] (f(x_{t+k}) + f(x_{t+k-1})) | x_{t}, \gamma_{t}\rt] 
&\leq \frac{\lt(2a + b\rt)C\gamma_t \P(\tau > t+1)}{1-v} < \infty \quad a.s.
\]
Therefore, $f(x_\tau)$ and $\E\lt[f(x_\tau)|x_t,\gamma_t\rt]$ are well-defined 
(as stated earlier) and
\[
\E\lt[f(x_{\tau})|x_t,\gamma_t\rt]&= \E\lt[f(x_{t+1})|x_t,\gamma_t\rt]+ \sum_{k=2}^{\infty}\E\lt[\1[\tau \geq t+k] (f(x_{t+k}) - f(x_{t+k-1})) | x_t,\gamma_t\rt].
\]
To obtain the bound in the result, we now apply the same technique as in the 
earlier analysis, except that we maintain the negative sign on the second $f$ 
term in the sum:
\[
& \E\lt[f(x_{\tau})|x_t,\gamma_t\rt]\\
&\leq  \E\lt[f(x_{t+1})|x_{t},\gamma_{t}\rt] + \sum_{k=2}^{\infty}\E\lt[\1[\tau \geq t+k] (f(x_{t+k-1})+ b \gamma_{t+k-1}  - f(x_{t+k-1})) | x_{t}, \gamma_{t}\rt]\\
&=  \E\lt[f(x_{t+1})|x_{t},\gamma_{t}\rt] + b\sum_{k=2}^{\infty}\E\lt[\1[\tau \geq t+k]\gamma_{t+k-1} | x_{t}, \gamma_{t}\rt]\\
&\leq  \E\lt[f(x_{t+1})|x_{t},\gamma_{t}\rt] +  b\lt(C\gamma_{t}\P(\tau>t+1|x_{t},\gamma_{t}) + \sum_{k=3}^{\infty}\E\lt[\1[\tau \geq t+k]\gamma_{t+k-1} | x_{t}, \gamma_{t}\rt]\rt)\\
&\leq  \E\lt[f(x_{t+1})|x_{t},\gamma_{t}\rt] +  \frac{bC\gamma_{t}\P(\tau>t+1|x_{t},\gamma_{t})}{1-v}\\
&\leq  \E\lt[f(x_{t+1})|x_{t},\gamma_{t}\rt] +  \frac{bC\gamma_{t}}{1-v}.
\]
\eprfof

\bprfof{\cref{lem:tourtime}} 
For notational brevity in the proof, let $z_t = (x_t,\gamma_t)$ and $V_t = V(z_t) = V(x_t,\gamma_t)$.
By definition of $\tau$,
\[
\P(\tau > k |z_t) &=\E\lt[\1\lt[\min_{j \in \{1,\dots,k\}} V_{t+j} > 1\rt]|z_t\rt]\\
&= \E\lt[\E\lt[\1\lt[V_{t+k} > 1\rt] | z_{t+k-1}\rt] \1\lt[\min_{j \in \{1,\dots,k-1\}} V_{t+j} > 1\rt]|z_t\rt]\\
&\leq \E\lt[\E\lt[V_{t+k}| z_{t+k-1}\rt] \1\lt[\min_{j \in \{1,\dots,k-1\}} V_{t+j}> 1\rt]|z_t\rt]\\
&\leq \xi\E\lt[V_{t+k-1} \1\lt[\min_{j \in \{1,\dots,k-1\}} V_{t+j} > 1\rt]|z_t\rt]\\
&\leq \xi\E\lt[V_{t+k-1} \1\lt[\min_{j \in \{1,\dots,k-2\}} V_{t+j}> 1\rt]|z_t\rt].
\]
Continuing this logic yields
\[
\P(\tau > k |z_t,\gamma_t) &\leq \xi^{k-1}\E\lt[V_{t+1} | z_t\rt] \leq \xi^{k-1}(\xi V_t + \zeta\1[V_t \leq 1]).
\]
\eprfof

\bprfof{\cref{thm:abstractconvergence}}
Let $t'_i$, $i\in\nats$ be the $i^\text{th}$ index $t\in\nats_0$ at which $a\gamma_t \leq f(x_t)$ (or $\infty$ if no such index exists); 
\cref{lem:tourdescent,eq:thmassump} imply that
\[
\E\lt[f(x_{t'_{1}}) | x_{0},\gamma_{0}\rt] &\leq \max\{f(x_0), B\gamma_0\}+ \gamma_{0}\frac{Cb}{1-v}\\
\forall i\in\nats, \quad \E\lt[f(x_{t'_{i+1}}) | x_{t'_i},\gamma_{t'_i}\rt] &\leq 
f(x_{t'_i})\lt(1-\frac{a\gamma_{t'_i}}{f(x_{t'_i})}\lt(\eta - \frac{Cb}{a(1-v)}\rt) \rt).
\]
By \cref{eq:thmassump}, the multiplicative factor is at most 1; hence $f(x_{t'_i})$ is a nonnegative supermartingale.
Let $i_n$, $n\in\nats$ be the $n^\text{th}$ index $i$ at which $V(x_{t'_i}, \gamma_{t'_i})\leq 1$.
Then by the optional stopping theorem for nonnegative supermartingales \cite[Theorem 5.7.6]{durrett}
and \cref{eq:thmassump}, 
\[
\E\lt[f(x_{t'_{i_{n+1}}}) | x_{t'_{i_n}}, \gamma_{t'_{i_n}}\rt]
&= 
\E\lt[\E\lt[f(x_{t'_{i_{n+1}}}) | x_{t'_{i_n+1}}, \gamma_{t'_{i_n+1}}\rt] | x_{t'_{i_n}}, \gamma_{t'_{i_n}}\rt]\\
&\leq \E\lt[f(x_{t'_{i_{n}+1}}) | x_{t'_{i_n}}, \gamma_{t'_{i_n}}\rt]\\
&\leq f(x_{t'_{i_n}})\lt(1-\frac{a\gamma_{t'_{i_n}}}{f(x_{t'_{i_n}})}\lt(\eta- \frac{Cb }{a(1-v)}\rt)\rt)\\
&\leq f(x_{t'_{i_n}})\lt(1-\ell\lt(\eta- \frac{Cb }{a(1-v)}\rt)\rt).
\]
Therefore,
\[
\E\lt[f(x_{t'_{i_{n}}})\rt] &\leq \nu^{n-1}\E\lt[f(x_{t'_{i_1}})\rt] \leq \nu^{n-1}\E\lt[f(x_{t'_{1}})\rt] \leq \nu^{n-1}f_0.
\]
The first result follows by setting $t_n = t'_{i_n}$.
Next, by \cref{lem:tourtime}, we have that
\[
\E[t_n | x_0,\gamma_0] &= \sum_{k=1}^n \E[t_k - t_{k-1} | x_0,\gamma_0]\\
&= \E[t_1-t_0 | x_0,\gamma_0] + \sum_{k=2}^n \E\lt[ \E\lt[t_k - t_{k-1} | x_{t_{k-1}},\gamma_{t_{k-1}}\rt] | x_0, \gamma_0\rt]\\
&\leq \frac{\xi V(x_0, \gamma_0) + \zeta\1[V(x_0, \gamma_0) \leq 1]}{1-\xi} + \sum_{k=2}^n \E\lt[ \frac{\xi+\zeta}{1-\xi} | x_0, \gamma_0\rt]\\
&= \frac{\xi V(x_0,\gamma_0) + \zeta\1[V(x_0,\gamma_0) \leq 1]}{1-\xi} + (n-1)\frac{\xi+\zeta}{1-\xi}\\
&= \frac{V_0}{1-\xi} + (n-1)\frac{\xi+\zeta}{1-\xi}.
\]
\eprfof

\bprfof{\cref{lem:innerdescent}}
Since $f$ is twice continuously differentiable by \cref{assum:f}, the Taylor remainder theorem \cite[Theorem 7.6]{apostol} and iteration $y_{k+1} = y_k-\gamma g(y_k,u_k)$
yield
\[
f(y_{k+1}) 
&= f(y_k) + \grad f(y_k)^T(y_{k+1}-y_k) \\ 
&{\qquad}+ (y_{k+1}-y_k)^T \int_0^1 (1-u) \grad^2 f((1-u)y_k+uy_{k+1})\d u (y_{k+1}-y_k)\\
&= f(y_k) - \gamma \grad f(y_k)^Tg(y_k,u_k) + \frac{1}{2}\gamma^2 g(y_k,u_k)^T \int_0^1 2(1-s) \grad^2 f(y_k - s \gamma  g_k)\d s  g(y_k,u_k),
\]
where $g_k := g(y_k, u_k)$.
Then by the unbiasedness of $g(y_k,u_k)$ for $\grad f(y_k)$ and the moment bound in \cref{assum:f},
\[
\E\lt[f(y_{k+1})| y_k\rt]
&\leq f(y_k) - \gamma \|\grad f(y_k)\|^2 + \frac{1}{2}\gamma^2\lt(\alpha \|\grad f(y_k)\|^2 + \beta\rt)\\
&= f(y_k) - \lt(\gamma - \frac{1}{2}\gamma^2\alpha\rt) \|\grad f(y_k)\|^2 + \frac{1}{2}\gamma^2\beta\\
&\leq f(y_k)\lt(1 - 2\mu\gamma\lt(1- \frac{1}{2}\gamma\alpha\rt)\rt) + \frac{1}{2}\gamma^2\beta.
\]
Since $f(y_k) \geq 0$, for all $0< z \leq 2\mu\lt(1-\frac{1}{2}\gamma\alpha\rt)$,
\[
\E\lt[f(y_{k+1})| y_k\rt]
&\leq f(y_k)(1 - \gamma z) + \frac{1}{2}\gamma^2\beta.
\]
Recursing this bound yields, for $j\leq k$,
\[
\E\lt[f(y_{k})| y_j\rt]  &\leq f(y_j)(1 - \gamma z)^{k-j} + \sum_{i=0}^{k-j-1} (1-\gamma z)^i \frac{1}{2}\gamma^2\beta\\
&= f(y_j)(1 - \gamma z)^{k-j} + (1- (1-\gamma z)^{k-j})\frac{\gamma\beta}{2 z}.
\]
\eprfof

\bprfof{\cref{lem:outerdescent}}
By the definition of $\gamma_{t+1}$,
\[
\E\lt[\gamma_{t+1}|x_t,\gamma_t\rt] &= \E\lt[\gamma_t\lt(I_t C + S_t + (D_t+R_t) c\rt)|x_t,\gamma_t\rt]\\
&= \gamma_t\lt(C \P(I_t|x_t,\gamma_t) + \P(S_t|x_t,\gamma_t) + c\P(D_t\cup R_t|x_t,\gamma_t)\rt). \label{eq:expectedouterg}
\]
If $C\gamma_t\beta/2\mu > f(x_t)$, we have $\P(R_t|x_t,\gamma_t) \geq 1-\delta$ by \cref{assum:irprobs},
and hence
\[
\E\lt[\gamma_{t+1} | x_t,\gamma_t\rt] &\leq \gamma_t\lt(\max\{C,1\}\P(I_t\cup S_t|x_t,\gamma_t) + c\P(D_t\cup R_t|x_t,\gamma_t)\rt)\\ 
&\leq \gamma_t\lt(C\delta + c(1-\delta)\rt).
\]
By definition of $f(x_{t+1})$,
\[
&\E\lt[ f(x_{t+1}) | x_t, \gamma_t\rt] \\
&=  \E\lt[  \1[I_t] f(\sbx_{t,\tau_t}) + \1[S_t] f(x_{t,\tau_t}) + \1[D_t] f(\slx_{t,\tau_t}) + \1[R_t]f(x_t)| x_t, \gamma_t \rt]\\
& = \E\lt[\sum_{k=1}^\infty \{\1[I_t,\tau_t=k] f(\sbx_{t,k}) + \1[S_t,\tau_t=k] f(x_{t,k}) + \1[D_t,\tau_t=k] f(\slx_{t,k})\} + \1[R_t]f(x_t)| x_t, \gamma_t \rt].
\]
As the summands are nonnegative, we can interchange the expectation and sum, and by \cref{assum:testpreferlower},
\[
&\E\lt[ f(x_{t+1}) | x_t, \gamma_t\rt] \\
&\leq \sum_{k=1}^\infty \big\{ \P(I_t,\tau_t=k | x_t, \gamma_t) \E\lt[f(\sbx_{t,k}) | x_t, \gamma_t\rt]
+\P(S_t,\tau_t=k | x_t, \gamma_t) \E\lt[f(x_{t,k}) | x_t, \gamma_t\rt]\\
&{\qquad}+\P(D_t,\tau_t=k | x_t, \gamma_t) \E\lt[f(\slx_{t,k}) | x_t, \gamma_t\rt] \big\}
+\P(R_t | x_t, \gamma_t) f(x_t).
\label{eq:expectedouterx}
\]
We have that $\gamma_t \leq \alpha^{-1}$ \as by \cref{assum:irprobs}.
If $\gamma_t \leq (C\alpha)^{-1}$, then
\[
\min_{z\in\{c,1,C\}} 2\mu\lt(1-z\frac{1}{2}\gamma_t\alpha\rt) &\geq 2\mu\lt(1-\frac{1}{2}C\gamma_t\alpha\rt) \geq \mu,
\] 
and the same result holds for $\min_{z\in\{c,1\}}$ if $\gamma_t \geq (C\alpha)^{-1}$.
So applying \cref{lem:innerdescent} with $\gamma$ set to $c\gamma_t$, $\gamma_t$, and $C\gamma_t$ yields the three bounds
\[
\E\lt[f(x_{t,k}) | x_t, \gamma_t\rt]&\leq f(x_t)(1 - \mu\gamma_t)^{k} + (1- (1-\mu\gamma_t)^{k})\frac{\gamma_t\beta}{2\mu}\\
\E\lt[f(\slx_{t,k}) | x_t, \gamma_t\rt] &\leq f(x_t)(1 - c\mu\gamma_t)^{k} + (1- (1-c\mu\gamma_t)^{k})\frac{c\gamma_t\beta}{2\mu}\\
\text{if } \gamma_t\leq (C\alpha)^{-1},\quad \E\lt[f(\sbx_{t,k}) | x_t, \gamma_t\rt] &\leq f(x_t)(1 - C\mu\gamma_t)^{k} + (1- (1-C\mu\gamma_t)^{k})\frac{C\gamma_t\beta}{2\mu}.
\]
Therefore,
\[
&\E\lt[ f(x_{t+1}) | x_t, \gamma_t\rt] \\
&{\quad}\leq \P(R_t | x_t, \gamma_t) f(x_t)\\
&{\qquad}+\sum_{k=1}^\infty \P(I_t,\tau_t=k | x_t, \gamma_t)\lt(f(x_t)(1 - C\mu\gamma_t)^{k} + (1- (1-C\mu\gamma_t)^{k})\frac{C\gamma_t\beta}{2\mu}\rt)\\
&{\qquad}+\sum_{k=1}^\infty \P(S_t,\tau_t=k | x_t, \gamma_t)\lt(f(x_t)(1 - \mu\gamma_t)^{k} + (1- (1-\mu\gamma_t)^{k})\frac{\gamma_t\beta}{2\mu}\rt)\\
&{\qquad}+\sum_{k=1}^\infty \P(D_t,\tau_t=k | x_t, \gamma_t)\lt(f(x_t)(1 - c\mu\gamma_t)^{k} + (1- (1-c\mu\gamma_t)^{k})\frac{c\gamma_t\beta}{2\mu}\rt),\label{eq:expectedouterx2}
\]
where the $I_t$ term holds since $\P(I_t | x_t,\gamma_t) = 0$ when $\gamma_t \geq (C\alpha)^{-1}$ by \cref{assum:irprobs}.
First consider the case where $C\gamma_t\beta/(2\mu) > f(x_t)$. 
In this case, we have $\P(R_t|x_t,\gamma_t) \geq 1-\delta$ by \cref{assum:irprobs}, and
so \cref{eq:expectedouterx2} yields
\[
&\E\lt[ f(x_{t+1}) | x_t, \gamma_t\rt] \\
&{\quad}\leq \P(R_t | x_t, \gamma_t) f(x_t)
+\sum_{k=1}^\infty \P(I_t,\tau_t=k | x_t, \gamma_t)\max\lt\{f(x_t),\frac{C\gamma_t\beta}{2\mu}\rt\}\\
&{\qquad}+\sum_{k=1}^\infty \P(S_t,\tau_t=k | x_t, \gamma_t)\max\lt\{f(x_t),\frac{\gamma_t\beta}{2\mu}\rt\}
+\sum_{k=1}^\infty \P(D_t,\tau_t=k | x_t, \gamma_t)\max\lt\{f(x_t),\frac{c\gamma_t\beta}{2\mu}\rt\}\\
&{\quad}\leq \P(R_t | x_t, \gamma_t) f(x_t) + \P(R_t^c | x_t,\gamma_t) \max\lt\{f(x_t),\frac{C\gamma_t\beta}{2\mu}, \frac{\gamma_t\beta}{2\mu}, \frac{c\gamma_t\beta}{2\mu}\rt\}\\
&{\quad}\leq \P(R_t | x_t, \gamma_t) f(x_t) + \P(R_t^c | x_t,\gamma_t) \frac{C\gamma_t\beta}{2\mu} \label{eq:f_t_bound_intermediate} \\
&{\quad}\leq f(x_t) + \gamma_t\delta\frac{C\beta}{2\mu}.
\]
This completes the results stated in \cref{eq:c2highresult}. Finally, consider the case where $C\gamma_t\beta/2\mu \leq f(x_t)$.
We split this into two subcases: one where $\gamma_t \leq (C\alpha)^{-1}$, and one where $(C\alpha)^{-1} < \gamma_t \leq \alpha^{-1}$.
In the subcase where $\gamma_t \leq (C\alpha)^{-1}$, by \cref{eq:expectedouterx2,assum:irprobs,assum:sufficientdecrease},
\[
&\E\lt[ f(x_{t+1}) | x_t, \gamma_t\rt] \\
&{\quad}\leq \P(R_t | x_t, \gamma_t) f(x_t) + \lt(1-\frac{u C\beta\gamma_t}{2\mu f(x_t)}\rt)\P(R_t^c | x_t,\gamma_t)f(x_t) \\
&{\qquad}+ \frac{u C\beta \gamma_t}{2\mu f(x_t)}\frac{\gamma_t\beta}{2\mu}\E\lt[\1[I_t]C + \1[S_t]+\1[D_t]c|x_t,\gamma_t\rt]\\
&{\quad}\leq \delta f(x_t) + \lt(1-\frac{u C\beta\gamma_t}{2\mu f(x_t)}\rt)(1-\delta)f(x_t) 
+ \frac{u C\beta \gamma_t}{2\mu f(x_t)}\frac{\gamma_t\beta}{2\mu}\E\lt[\1[I_t]C + \1[S_t]+\1[D_t]c|x_t,\gamma_t\rt]\\
&{\quad}\leq \delta f(x_t) + \lt(1-\frac{u C\beta \gamma_t}{2\mu f(x_t)}\rt)(1-\delta)f(x_t) + \frac{u C\beta \gamma_t}{2\mu f(x_t)}(1-\delta)\epsilon f(x_t)\\
&{\quad}= \lt(1 -\frac{u C \beta\gamma_t}{2\mu f(x_t)}(1-\delta)(1-\epsilon)\rt)f(x_t).
\]
In the subcase where $(C\alpha)^{-1} < \gamma_t \leq \alpha^{-1}$, since $\tau_t \geq 1$,
\[
\max\lt\{\E[(1-\mu\gamma_t)^{\tau_t} | S_t, x_t, \gamma_t], \E[(1-c\mu\gamma_t)^{\tau_t} | D_t, x_t, \gamma_t]\rt\} &\leq \lt(1-\frac{c\mu}{C\alpha}\rt) \leq \lt(1-u\rt),
\]
and so by \cref{eq:expectedouterx2,assum:irprobs,assum:sufficientdecrease},
\[
&\E\lt[ f(x_{t+1}) | x_t, \gamma_t\rt] \\
&\leq \P(R_t | x_t, \gamma_t) f(x_t) +\lt(1-u\rt)\P(R_t^c | x_t,\gamma_t)f(x_t) + u\frac{\gamma_t\beta}{2\mu}\E\lt[\1[I_t]C + \1[S_t]+\1[D_t]c|x_t,\gamma_t\rt]\\
&\leq \delta f(x_t) +\lt(1-u\rt)(1-\delta)f(x_t) + u(1-\delta)\epsilon f(x_t)\\
&= \lt(1 -u(1-\delta)(1-\epsilon)\rt)f(x_t)\\
&\leq \lt(1 -\frac{uC\gamma_t\beta}{2\mu f(x_t)}(1-\delta)(1-\epsilon)\rt)f(x_t).
\]
This completes the result stated in \cref{eq:c2lowresult}.
\eprfof

\bprfof{\cref{lem:outerlyapunov}}
For all $t\in\nats_0$, denote $V_t = V(x_t,\gamma_t)$ for brevity.
\[
\E\lt[V_{t+1} | x_t,\gamma_t\rt] &= 
\E\lt[\frac{\ell f(x_{t+1})}{a\gamma_{t+1}}\1\lt[\frac{\ell f(x_{t+1})}{a\gamma_{t+1}}>1\rt] + \frac{a\gamma_{t+1}}{f(x_{t+1})}\1\lt[\frac{a\gamma_{t+1}}{f(x_{t+1})}>1\rt] | x_t,\gamma_t\rt].
\]
We begin by analyzing the first term. By definition of $f(x_{t+1})$,
\[
&\E\lt[\frac{\ell f(x_{t+1})}{a\gamma_{t+1}}\1\lt[\frac{\ell f(x_{t+1})}{a\gamma_{t+1}}>1\rt] | x_t,\gamma_t\rt]\\
&= \E\lt[\frac{\ell(I_t+S_t+D_t+R_t)f(x_{t+1})}{a\gamma_{t+1}}\1\lt[\frac{\ell f(x_{t+1})}{a\gamma_{t+1}}>1\rt] | x_t,\gamma_t\rt]\\
&\leq \E\lt[\frac{\ell\lt(I_t+S_t+D_t+R_t\1\lt[\frac{\ell f(x_{t+1})}{a\gamma_{t+1}}>1\rt]\rt)f(x_{t+1})}{a\gamma_{t+1}} | x_t,\gamma_t\rt]\\
&= \E\lt[\frac{I_t\ell f(\sbx_{t,\tau_t})}{a C\gamma_{t}} | x_t,\gamma_t\rt]
+\E\lt[\frac{S_t\ell f(x_{t,\tau_t})}{a \gamma_{t}} | x_t,\gamma_t\rt]
+\E\lt[\frac{D_t\ell f(\slx_{t,\tau_t})}{ac\gamma_{t}} | x_t,\gamma_t\rt] \\
&{\quad}+\E\lt[\frac{R_t\ell f(x_{t})}{a c\gamma_{t}}\1\lt[\frac{\ell f(x_{t+1})}{a\gamma_{t+1}}>1\rt] | x_t,\gamma_t\rt].
\]
Then by \cref{assum:testpreferlower,lem:innerdescent},
\[
&\leq \P\lt(R_t,\frac{\ell f(x_{t+1})}{a\gamma_{t+1}}>1|x_t,\gamma_t\rt)\frac{\ell f(x_t)}{ac\gamma_{t}}\\
&+\sum_{k=1}^\infty \frac{\ell \P(I_t,\tau_t=k | x_t, \gamma_t)}{a C\gamma_t}\lt(f(x_t)(1 - C\mu\gamma_t)^{k} + (1- (1-C\mu\gamma_t)^{k})a\gamma_t\rt)\\
&+\sum_{k=1}^\infty \frac{\ell \P(S_t,\tau_t=k | x_t, \gamma_t)}{a \gamma_t}\lt(f(x_t)(1 - \mu\gamma_t)^{k} + (1- (1-\mu\gamma_t)^{k})\frac{a}{C}\gamma_t \rt)\\
&+\sum_{k=1}^\infty \frac{\ell \P(D_t,\tau_t=k | x_t, \gamma_t)}{a c\gamma_t}\lt(f(x_t)(1 - c\mu\gamma_t)^{k} + (1- (1-c\mu\gamma_t)^{k})\frac{c a}{C}\gamma_t\rt),
\]
where the $I_t$ term holds since $\P(I_t | x_t,\gamma_t) = 0$ when $\gamma_t \geq (C\alpha)^{-1}$ by \cref{assum:irprobs}.
When $a\gamma_t < \ell f(x_t)$ and $\gamma_t \leq (C\alpha)^{-1}$,
we have $\P(I_t|x_t,\gamma_t) \geq 1-\delta$ by \cref{assum:irprobs}.
Therefore,
\[
&\E\lt[\frac{\ell f(x_{t+1})}{a\gamma_{t+1}}\1\lt[\frac{\ell f(x_{t+1})}{a\gamma_{t+1}}>1\rt] | x_t,\gamma_t\rt] \\
&{\quad}\leq 
\frac{\ell f(x_t)}{a\gamma_t}\lt(C^{-1}\P(I_t | x_t, \gamma_t) + \P(S_t | x_t, \gamma_t)
+c^{-1}\P(D_t | x_t, \gamma_t) + \P(R_t | x_t, \gamma_t)\rt)\\
&{\quad}\leq 
\frac{\ell f(x_t)}{a\gamma_t}\lt(\frac{1-\delta}{C} + \frac{\delta}{c}\rt).\label{eq:t1gamsmall1}
\]
On the other hand, when $a\gamma_t < \ell f(x_t)$ and $(C\alpha)^{-1} < \gamma_t \leq \alpha^{-1}$,
we have $\P(I_t|x_t,\gamma_t) = 0$
and $\P(R_t | x_t, \gamma_t) \leq \delta$
by \cref{assum:irprobs} and 
\[
\max\lt\{\E[(1-\mu\gamma_t)^{\tau_t} | S_t, x_t, \gamma_t], \E[(1-c\mu\gamma_t)^{\tau_t} | D_t, x_t, \gamma_t]\rt\} &\leq w c
\]
by \cref{assum:sufficientdecrease}.
Therefore,
\[
&\E\lt[\frac{\ell f(x_{t+1})}{a\gamma_{t+1}}\1\lt[\frac{\ell f(x_{t+1})}{a\gamma_{t+1}}>1\rt] | x_t,\gamma_t\rt] \\
&{\quad}\leq 
\frac{\ell f(x_t)}{a\gamma_t}\lt( (wc + (1-wc)\frac{\ell}{C})\P(S_t|x_t,\gamma_t) + c^{-1}(wc + (1-wc)\frac{c\ell}{C})\P(D_t|x_t,\gamma_t) + \delta\rt)\\
&{\quad}\leq 
\frac{\ell f(x_t)}{a\gamma_t}\lt(w + \frac{(1-wc)\ell}{C} + \delta\rt).\label{eq:t1gamsmall2}
\]

When $a\gamma_t > f(x_t)$, we have $\P(R_t|x_t,\gamma_t) \geq 1-\delta$ by \cref{assum:irprobs}. Furthermore,
if $R_t = 1$, then $\frac{\ell f(x_{t+1})}{a\gamma_{t+1}} = \frac{\ell f(x_t)}{a c\gamma_t}$; therefore, as long 
as $\ell \leq c$, $R_t = 1$ and $\frac{\ell f(x_{t+1})}{a\gamma_{t+1}} > 1$ cannot simultaneously occur when $a\gamma_t > f(x_t)$. So
when $a\gamma_t > f(x_t)$,
\[
\E\lt[\frac{\ell f(x_{t+1})}{a\gamma_{t+1}}\1\lt[\frac{\ell f(x_{t+1})}{a\gamma_{t+1}}>1\rt] | x_t,\gamma_t\rt]
&\leq 
a\gamma_t\frac{\ell}{a\gamma_t}c^{-1}\delta = \ell c^{-1}\delta \leq \ell c^{-1}\delta \frac{a\gamma_t}{f(x_t)}.\label{eq:t1gamlarge}
\]
Finally, when $\ell f(x_t) \leq a\gamma_t \leq f(x_t)$,
\[
\E\lt[\frac{\ell f(x_{t+1})}{a\gamma_{t+1}}\1\lt[\frac{\ell f(x_{t+1})}{a\gamma_{t+1}}>1\rt] | x_t,\gamma_t\rt]
\leq \frac{\ell f(x_t)}{a\gamma_t} c^{-1} \leq c^{-1}.\label{eq:t1gammed}
\]
Next, we analyze the second term. Begin by noting that
\[
&\E\lt[\frac{a\gamma_{t+1}}{f(x_{t+1})}\1\lt[\frac{a\gamma_{t+1}}{f(x_{t+1})}>1\rt] | x_t,\gamma_t\rt]\\
&= \int_0^\infty \P\lt(\frac{a\gamma_{t+1}}{f(x_{t+1})}\1\lt[\frac{a\gamma_{t+1}}{f(x_{t+1})}>1\rt] > u | x_t, \gamma_t\rt) \d u\\
&= \int_1^\infty \P\lt(\frac{a\gamma_{t+1}}{f(x_{t+1})} > u | x_t, \gamma_t\rt) \d u\\
&= \int_1^\infty \lt(\P\lt(\frac{ac\gamma_{t}}{f(x_{t})} > u,R_t | x_t, \gamma_t\rt)+\P\lt(\frac{a\gamma_{t+1}}{f(x_{t+1})} > u,R_t^c | x_t, \gamma_t\rt)\rt) \d u\\
&= \P(R_t|x_t,\gamma_t)\lt(1\vee \frac{ac\gamma_t}{f(x_t)}-1\rt) + \int_1^\infty \P\lt(\frac{a\gamma_{t+1}}{f(x_{t+1})} > u,R_t^c | x_t, \gamma_t\rt)\d u\\
&= \P(R_t|x_t,\gamma_t)\lt(1\vee \frac{ac\gamma_t}{f(x_t)}-1\rt) + \int_0^1 s^{-2} \P\lt(f(x_{t+1}) < sa \gamma_{t+1},R_t^c | x_t, \gamma_t\rt)\d s \quad \text{($u = s^{-1}$)}\\
&= \P(R_t|x_t,\gamma_t)\lt(1\vee \frac{ac\gamma_t}{f(x_t)}-1\rt) + \int_0^1 s^{-2} \P\lt(f(\sbx_{t,\tau_t}) < sa C\gamma_{t},I_t | x_t, \gamma_t\rt)\d s\\
&{\quad}+\int_0^1 s^{-2} \P\lt(f(x_{t,\tau_t}) < sa \gamma_{t},S_t | x_t, \gamma_t\rt)\d s+\int_0^1 s^{-2} \P\lt(f(\slx_{t,\tau_t}) < sa c\gamma_{t},D_t | x_t, \gamma_t\rt)\d s.
\]
By \cref{assum:boundeddecrease},
\[
\E\lt[\frac{a\gamma_{t+1}}{f(x_{t+1})}\1\lt[\frac{a\gamma_{t+1}}{f(x_{t+1})}>1\rt] | x_t,\gamma_t\rt]
&\leq \P(R_t|x_t,\gamma_t)\lt(1\vee \frac{ac\gamma_t}{f(x_t)}-1\rt) + \sigma \frac{aC\gamma_t}{f(x_t)} \P(R_t^c|x_t,\gamma_t).
\]
When $a\gamma_t < \ell f(x_t)$, by \cref{assum:irprobs} 
we have that $\P(R_t | x_t,\gamma_t) < \delta$, so
\[
\E\lt[\frac{a\gamma_{t+1}}{f(x_{t+1})}\1\lt[\frac{a\gamma_{t+1}}{f(x_{t+1})}>1\rt] | x_t,\gamma_t\rt]
&\leq \delta \frac{ac\gamma_t}{f(x_t)} + \sigma\frac{aC\gamma_t}{f(x_t)} \\
&\leq \delta c \ell + \sigma C\ell \\
&\leq \lt(\delta c \ell + \sigma C\ell\rt)\frac{\ell f(x_t)}{a\gamma_t}.\label{eq:t2gamsmall}
\]
When $a\gamma_t > f(x_t)$, we have that $\P(R_t^c|x_t,\gamma_t) < \delta$ by \cref{assum:irprobs}, so
\[
\E\lt[\frac{a\gamma_{t+1}}{f(x_{t+1})}\1\lt[\frac{a\gamma_{t+1}}{f(x_{t+1})}>1\rt] | x_t,\gamma_t\rt]
&\leq \frac{ac\gamma_t}{f(x_t)} + \sigma\delta C \frac{a\gamma_t}{f(x_t)} \\
&= \lt(c + \sigma\delta C\rt)\frac{a\gamma_t}{f(x_t)}.
\label{eq:t2gamlarge}
\]
Finally, when $\ell f(x_t) \leq a\gamma_t \leq f(x_t)$, again we have that $\P(R_t|x_t,\gamma_t) < \delta$ by \cref{assum:irprobs}, so
\[
\E\lt[\frac{a\gamma_{t+1}}{f(x_{t+1})}\1\lt[\frac{a\gamma_{t+1}}{f(x_{t+1})}>1\rt] | x_t,\gamma_t\rt]
&\leq \delta \frac{ac\gamma_t}{f(x_t)} + \sigma\frac{aC\gamma_t}{f(x_t)} \leq \delta c + \sigma C.\label{eq:t2gammed}
\]
When $a\gamma_t < \ell f(x_t)$, combining \cref{eq:t1gamsmall1,eq:t1gamsmall2,eq:t2gamsmall} yields
\[
\E\lt[V_{t+1}|x_t,\gamma_t\rt]
&\leq \lt(\lt(w + \frac{(1-wc)\ell}{C} + \delta \rt)\vee \lt(\frac{1-\delta}{C} + \frac{\delta}{c}\rt)\rt)\frac{\ell f(x_t)}{a\gamma_t}+\lt(\delta c \ell + \sigma C\ell\rt)\frac{\ell f(x_t)}{a\gamma_t}\\
&= \lt(\lt(w + \frac{(1-wc)\ell}{C} + \delta \rt)\vee \lt(\frac{1-\delta}{C} + \frac{\delta}{c}\rt)+\delta c \ell + \sigma C\ell\rt)V_t.
\]
When $a\gamma_t > f(x_t)$, combining \cref{eq:t1gamlarge,eq:t2gamlarge} yields
\[
\E\lt[V_{t+1}|x_t,\gamma_t\rt]
&\leq 
\ell c^{-1}\delta \frac{a\gamma_t}{f(x_t)} + 
\lt(c + \sigma\delta C\rt)\frac{a\gamma_t}{f(x_t)}\\
&= 
\lt(\ell c^{-1}\delta +c + \sigma\delta C\rt)V_t.
\]
When $\ell f(x_t) \leq a\gamma_t \leq f(x_t)$, combining \cref{eq:t1gammed,eq:t2gammed} yields
\[
\E\lt[V_{t+1}|x_t,\gamma_t\rt]
&\leq c^{-1} + \delta c + \sigma C.
\]
\eprfof

\clearpage
\section{Constant memory round-based averaging}
\label{sec:averaging}

In this section we present a round-based iterate averaging scheme that has two 
desirable properties. 
First, it approximately preserves the final iterate convergence rates 
\cref{result:averaging_general}. 
This is in contrast to standard averaging schemes such as 
full-trace Polyak or polynomial averaging 
\cite{shamir2013averaging}, 
which can guarantee only $O(1/t)$ convergence due to the bias of the initial terms. 
Second, the iterate average can be computed efficiently in an 
online fashion with constant memory in the length of the trace and can be thought 
of as a constant-memory implementation of an average of the last $\delta \times 100\%$ 
of iterates. 
This approach to averaging is known as suffix averaging or tail averaging in 
the literature 
\cite{jain2018averaging,rakhlin2011averaging,mucke2019averaging}.

Our contribution is most similar to that of \cite{roux2019anytimetail}, with slightly different 
weighting schemes and decision times for when to start a new round of averages.
The key idea of our method is to compute two streams of online averages of iterates
and then interpolate between them to mimic the desired behaviour.
We use a round-based procedure to discard past iterates to ensure the initial 
bias of early iterates is dropped.
Among other related work, in the context of variational inference, \cite{dhaka2020robust} consider using
the $\hat R$ statistic \cite{gelman1992Rhat,gelman2013bda} to assess when to 
start averaging, along with Monte Carlo standard error to determine when to stop averaging.

Consider an arbitrary sequence $(x_t)_{t \geq 0}$ and suppose we define
round $r$ to consist of samples $x_t$ for $T_r \leq t < T_{r+1}$, for some increasing 
sequence $(T_r)_{r \geq 0}$ (e.g., $T_r = 2^r - 1$).
Define the average of the $r^\text{th}$ round as
\[
  \bar x^{(r)} = \frac{1}{T_{r+1} - T_r} \sum_{t=T_r}^{T_{r+1}-1} x_t.
\]

The proposed average at time $t$ is denoted $z_t$, which has the form 
\[
  z_t = 
  \begin{cases}
    \bar x^{(r)}, & t = T_{r+1}-1 \text{ for some } r \geq 0 \\ 
    (1-\alpha_t) \bar x^{(r-1)} + \frac{\alpha_t}{t - T_r+1} \sum_{i=T_r}^t x_i, 
      & T_r \leq t < T_{r+1}-1 \text{ for some } r \geq 0 
  \end{cases},
\]
for some $0 \leq \alpha_t \leq 1$. 
This is a linear interpolation between the previous round's online average 
and the one of the current round.

\begin{algorithm}
	\begin{algorithmic}[1]
    \Require Stream of iterates $(x_t)_{t \geq 0}$, 
      number of rounds $R \geq 1$,
      round starting points $(T_r)_{r \geq 0}$, 
      interpolation sequence $(\alpha_t)_{t \geq 0}$.
    \State $\bar x^{(0)} \gets 0$
    \State $t \gets 0$
    \While{$t < T_1$}  \Comment{round 0}
      \State $\bar x^{(0)} \gets t \bar x^{(0)} + x_t$  \Comment{online update}
      \State $z_t \gets \bar x^{(0)}$
      \State $t \gets t+1$
    \EndWhile 
    \For{$r$ {\bf in} $1, \ldots, R$}
      \State $\bar x^{(r)} \gets 0$ 
      \While{$t < T_{r+1}$} 
        \State $\bar x^{(r)} \gets (t-T_r) \bar x^{(r)} + x_t$ 
        \If{$t = T_{r+1}-1$} 
          \State $z_t \gets \bar x^{(r)}$ 
            \Comment{end of round, use final average}
        \Else   
          \State $z_t \gets (1-\alpha_t) \bar x^{(r-1)} + \alpha_t \bar x^{(r)}$ 
            \Comment{interpolate between round $r$ and $r-1$}
        \EndIf
        \State $t \gets t+1$
      \EndWhile 
	\EndFor
    \State \Return $(z_t)_{t = 0}^{T_R}$
	\end{algorithmic}
  \caption{Constant-memory online round-based averaging}
  \label{alg:averaging}
\end{algorithm}

In our implementation, we set $T_r = 2^r-1$ and for $T_r \leq t < T_{r+1}$ use  
\[
  \alpha_t = 
  \begin{cases}
    \frac{t - T_r + 1}{t - T_{r-1}+1}, & t - T_{r-1}+1 \leq T_{r+1}-T_r \\
    \frac{t - T_r + 1}{T_{r+1} - T_r}, & t - T_{r-1}+1 > T_{r+1}-T_r
  \end{cases}.
\]
This approach roughly corresponds to online $\delta$-suffix averaging with $\delta = 0.5$.
More generally, setting $T_r \sim 1/(1-\delta)^r$ approximately corresponds to online, 
constant-memory $\delta$-suffix averaging.
An online constant-memory implementation of round-based averaging is presented in 
\cref{alg:averaging} and a theoretical result is stated in \cref{result:averaging_general}. 

The following result establishes the convergence rate of the 
proposed round-based averaging scheme.

\bthm 
\label{result:averaging_general}
Suppose $f$ is lower-bounded and convex and that there exist non-negative non-increasing sequences 
$(a_t)_{t \geq 0}$ and $(b_t)_{t \geq 0}$ such that  
\[
  \E[f(x_t) - \inf f] \leq a_t [f(x_0) - \inf f] + b_t.
\]
Then for any $T_r \leq t < T_{r+1}$ with $r \geq 1$,
\[
  \E[f(z_t) - \inf f] \leq a_{T_{r-1}} [f(x_0) - \inf f] + b_{T_{r-1}}.
\]
\ethm

\bprfof{\cref{result:averaging_general}}
By assumption,
\[
  \E[f(x_t) - \inf f] \leq a_t [f(x_0) - \inf f] + b_t.
\]
Then for any $T_r \leq t < T_{r+1}$, by convexity of $f$,
\[
  f(z_t)  
  &\leq (1-\alpha_t) f(\bar x^{(r-1)}) + \frac{\alpha_t}{t - T_r+1} \sum_{i=T_r}^t f(x_i) \\
  &\leq \frac{1-\alpha_t}{T_r - T_{r-1}} \sum_{i=T_{r-1}}^{T_r-1} f(x_i) + \frac{\alpha_t}{t - T_r+1} \sum_{i=T_r}^t f(x_i). 
\] 
Then subtracting $\inf f$ from both sides and taking expectations, noting that 
$a_{t'} \geq a_t$ and $b_{t'} \geq b_t$ for any $t' \geq t$, we have
\[
  \E[f(z_t) - \inf f] 
  \leq a_{T_{r-1}} [f(x_0) - \inf f] + b_{T_{r-1}}.
\]
\eprfof
\clearpage
\section{Additional experimental details and results} 
\label{sec:supp_experiments}

The models and datasets considered in this paper are summarized in 
\cref{tab:exp_settings}. 

\begin{table}
    \centering
    \begin{tabular}{llrr}
    	\toprule
        &Optimization problem & $n$ & $d$ \\ 
      \midrule 
	      Classical optimization & Sum of quadratics & 100 & 10 \\
	      & Matrix factorization ($k \in \{1, 4, 10\}$) & 1,000 & $16k$ \\
	      & Logistic regression & 1,000 & 10,000 \\
	      & Least squares & 1,000 & 10,000 \\
	      & Multiclass logistic regression & 1,000 & 50,000 \\
	    \midrule
        ML training & ResNet + CIFAR-100 & 50,000 & 11,227,812 \\
        & ResNet + CIFAR-10 & 50,000 & 11,181,642 \\
        & RoBERTA-base + QNLI & 104,743 & 124,647,170 \\
        & RoBERTA-base + MRPC & 3,668 & 124,647,170 \\
        & RoBERTA-base + SST-2 & 67,349 & 124,647,170 \\
       \bottomrule
       \vspace{0.1in}
    \end{tabular}
    \caption{Summary of the models and data sets considered in this paper.}
    \label{tab:exp_settings}
\end{table}

\subsection{General details of experiments} 
\label{sec:ML_details}

All classical optimization and ML training experiments are performed 
on the UBC ARC Sockeye compute cluster.
For the ML training experiments, we use 16 GB and 32 GB NVIDIA Tesla V100 GPUs.
For the classical optimization experiments, we manage our experimental workflow 
with Nextflow (\url{https://www.nextflow.io/}) and use the CPU resources 
provided by UBC ARC Sockeye.
All ML and classical optimization experiments are run with 8 GB of RAM.

The estimated runtimes for each of the experiments are as follows: 
sum of quadratics (15 minutes), 
matrix factorization (1 hour), 
logistic regression (5 hours), 
least squares (4 hours), 
multiclass logistic regression (13 hours), 
CIFAR-100 (10 hours), 
CIFAR-10 (10 hours), 
QNLI (3 hours), 
MRPC (15 minutes), 
SST-2 (1 hour). 

For the ML experiments, train and test splits are created according to the standard 
HuggingFace dataset labels. 
The following batch sizes were used: 
SST-2 (16), QNLI (16), MRPC (16), CIFAR-10 (32), CIFAR-100 (32).
For all classical optimization experiments, we used a batch size of 1. 
There were no test splits in these settings.

The RoBERTa-base checkpoint is given by the \texttt{roberta-base} identifier in HuggingFace 
\cite{robertahuggingface}.
The ResNet18 checkpoint is given by \texttt{microsoft/resnet-18} \cite{resnet18huggingface}.
The learning rate grid used for the ML experiments is given by $\gamma \in \{10^{-1}, 10^{-2}, 10^{-3}\}$.
The solid and dashed lines for each of the five settings in 
\cref{fig:ML_training,fig:ML_validation} are given in \cref{tab:ML_legend}.

\begin{table}
    \centering
    \begin{tabular}{lrrrr}
        \toprule
        Dataset & 
        SGD constant & 
        SGD invsqrt & 
        SFSGD & 
        AutoSGD \\
        & (best/worst) & (best/worst)
        & (best/worst) & (best/worst) \\
        \midrule
        CIFAR-100 & $10^{-1}$ / $10^{-3}$ & $10^{-1}$ / $10^{-3}$ & $10^{-1}$ / $10^{-3}$ & $10^{-3}$ / $10^{-1}$ \\
        CIFAR-10  & $10^{-1}$ / $10^{-3}$ & $10^{-1}$ / $10^{-3}$ & $10^{-1}$ / $10^{-3}$ & $10^{-3}$ / $10^{-1}$ \\
        QNLI      & $10^{-2}$ / $10^{-1}$ & $10^{-1}$ / $10^{-2}$ & $10^{-1}$ / $10^{-3}$ & $10^{-2}$ / $10^{-3}$ \\
        MRPC      & $10^{-2}$ / $10^{-3}$ & $10^{-1}$ / $10^{-3}$ & $10^{-1}$ / $10^{-3}$ & $10^{-2}$ / $10^{-1}$ \\
        SST-2     & $10^{-3}$ / $10^{-1}$ & $10^{-1}$ / $10^{-3}$ & $10^{-1}$ / $10^{-3}$ & $10^{-2}$ / $10^{-3}$ \\
        \bottomrule
    \end{tabular}
    \caption{Learning rates for various optimizers in \cref{fig:ML_training,fig:ML_validation}.}
    \label{tab:ML_legend}
\end{table}

The licenses for the models and datasets used in the experiments are as follows: 
\begin{itemize}
  \item ResNet18: Apache License (Version 2.0)
  \item RoBERTa-base: MIT License 
  \item SST-2, QNLI, MRPC, CIFAR-10, CIFAR-100: From HuggingFace Datasets (unknown license). 
    No license was specified at the source.
\end{itemize}

In our experiments we used $C = 1/c = 2$ for all settings in the implementation 
of AutoSGD. For iterations where 
a restart occurs ($R_t = 1$), we decreased the learning rates in the grid by a 
factor of $c^2$ for the classical optimization experiments. 
For the ML training experiments, we decreased the learning rates by a factor of $c$ 
as we found that a decrease of $c^2$ could be overly aggressive.

For optimizers with additional hyperparameters, we set the parameters to reasonable 
defaults based on standard implementations (e.g., in PyTorch), recommendations 
from papers, and personal experience from which parameter settings performed well.
For instance, for schedule-free SGD we set $\beta = 0.9$. 
For DoG, we set $r_\epsilon = 10^{-6}$ and $\epsilon = 10^{-8}$. 
For non-monotone line search, we set $c = 0.1$, $\delta = 0.5$, $\xi = 0.5$, 
and the maximum number of backtracks to 20.

\subsection{Details of classical optimization experiments} 
\label{sec:classical_details} 

The definitions of each of the five classical optimization experiments is given below. 
All settings, except for the sum of quadratics problem, are (modifications of) those considered in 
\cite{vaswani2019painlesssgd} and \cite{nutini2022block}.

\subsubsection{Sum of quadratics}
The sum of quadratics optimization problem is given by
\[
  f(x) = \frac{1}{n} \sum_{i=1}^n \|X_i - x \|^2,
\]
where $X_i$ is the $i^\text{th}$ row of the matrix $X \in \reals^{n \times d}$ 
drawn from $X \sim \Norm_{n \times d}(0, I)$ with $n=100$ and $d=10$. 
We initialize $x$ according to $x \sim \Norm(0, 100 I_d)$.
For these experiments, we consider learning rates in the grid $\{10^{-1}, 10^{-2}, 10^{-3}, 10^{-5}\}$.

\subsubsection{Matrix factorization} 
The matrix factorization optimization problems for $k=1,4,10$ are given by 
\[
  f(W_1, W_2) = \frac{1}{n} \sum_{i=1}^n \|W_2 W_1 x_i - A x_i \|^2.
\]
Here, each $x_i \sim \Norm(0, I_6)$ with $n=1,000$. 
We have $W_1 \in \reals^{k \times 6}$ and $W_2 \in \reals^{10 \times k}$ 
for $k \in \{1, 4, 10\}$.
The matrix $A \in \reals^{10 \times 6}$ is created by:
\begin{enumerate}
  \item Creating a $10 \times 6$ matrix $Z$ with each entry distributed according 
  to an independent $U(0, 1)$ distribution.
  \item Multiplying each row of $Z$ by a linearly-spaced sequence from 1 to $\kappa$ with 
  10 entries.
\end{enumerate}
We set $\kappa = 10^5$. 
We initialize according to $W_1 \sim \Norm_{k \times 6}(0, I)$ and 
$W_2 \sim \Norm_{10 \times k}(0, I)$.
For these experiments, we consider learning rates in the grid $\{10^{-7}, 10^{-8}, 10^{-12}\}$.

\subsubsection{Least squares}
The least squares optimization problem is given by
\[
  f(x) = \frac{1}{2} \|Ax - b\|^2,
\]
where $A \in \reals^{n \times d}$, $b \in \reals^n$ with $n=1,000$ and $d=10,000$. 
The matrix $A$ is generated as follows: 
\begin{enumerate}
  \item Each entry is drawn according to $\Norm(1,1)$ 
  \item Each column is multiplied by an independent $\Norm(0, 10^2)$ draw. 
  \item Each entry of $A$ is non-zero with probability $10 \log(n)/n$.
\end{enumerate}
We then set $b = Ax + e$, where $e \sim \Norm(0,I_n)$.
To initialize $x$, we set ~50\% of entries to zero and the remaining entries are drawn 
from $\Norm(0,1)$.
For these experiments, we consider learning rates in the grid $\{10^{-8}, 10^{-9}, 10^{-11}, 10^{-12}\}$.

\subsubsection{Logistic regression}
The logistic regression optimization problem is given by 
\[
  f(x) = \sum_{i=1}^n \log(1 + \exp(-b_i x^\top a_i)),
\]
where $a_i$ is the $i^\text{th}$ row of $A$ as specified for the least squares 
objective and $x$ is initialized as in the setting. We set $b_i = \text{sign}(x^\top a_i)$ 
but flip each entry with probability 0.1.
Here, $n=1,000$ and $d=10,000$.
For these experiments, we consider learning rates in the grid $\{10^{-6}, 10^{-7}, 10^{-9}\}$.

\subsubsection{Multiclass logistic regression}
The multiclass logistic regression optimization problem is given by
\[
  f(x) = \sum_{i=1}^n \left[-x_{b_i}^\top a_i + \log\left( \sum_{c=1}^k \exp(x^\top_c a_i)\right)\right].
\]
Here, we initialize the matrix $X \sim \Norm_{d \times k}(0, I)$ where there are 
$k=50$ labels. The $b_i$ are generated as the maximum index of $(AX + E)_i$, 
where $A$ is as in the least squares setting, and $E \sim \Norm_{n \times k}(0, I)$.
We have $n=1,000$ and $d = k \times 1,000 = 50,000$.
For these experiments, we consider learning rates in the grid $\{10^{-6}, 10^{-7}, 10^{-8}\}$.

%%%%%%%%%%%%%%%%%%%%%%%%%%%%%%%%%%%%%%%%%%%%%%%%%%%%%%%%%%%%%%%%%%%%%%%%%%%%%%%%

\subsection{Additional results}
\label{sec:exp_additional_results}

\begin{figure*}
  \centering
  \begin{subfigure}{0.32\textwidth}
    \centering
    \includegraphics[width=\textwidth]{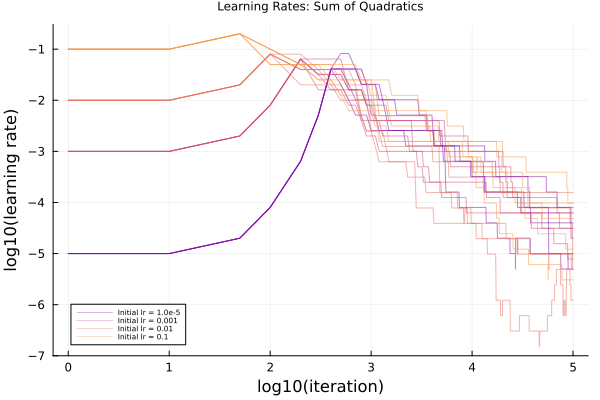}
  \end{subfigure}
  \begin{subfigure}{0.32\textwidth}
    \centering
    \includegraphics[width=\textwidth]{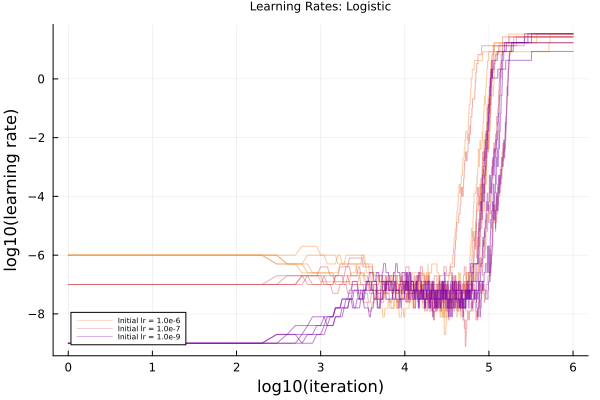}
  \end{subfigure} \\
  \begin{subfigure}{0.32\textwidth}
    \centering
    \includegraphics[width=\textwidth]{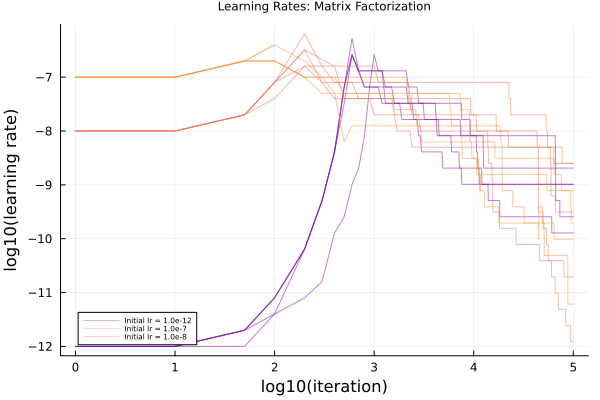}
  \end{subfigure} 
  \begin{subfigure}{0.32\textwidth}
    \centering
    \includegraphics[width=\textwidth]{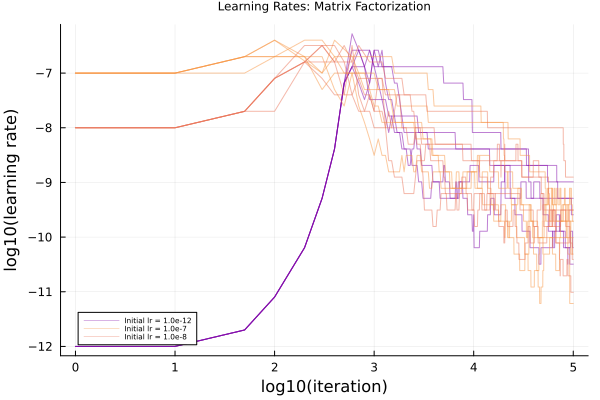}
  \end{subfigure} 
  \begin{subfigure}{0.32\textwidth}
    \centering
    \includegraphics[width=\textwidth]{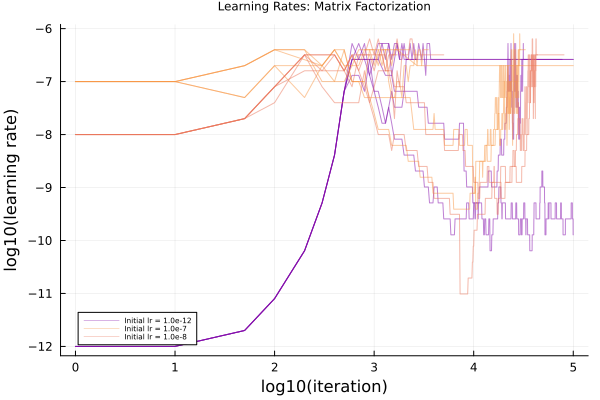}
  \end{subfigure} \\
  \begin{subfigure}{0.32\textwidth}
    \centering
    \includegraphics[width=\textwidth]{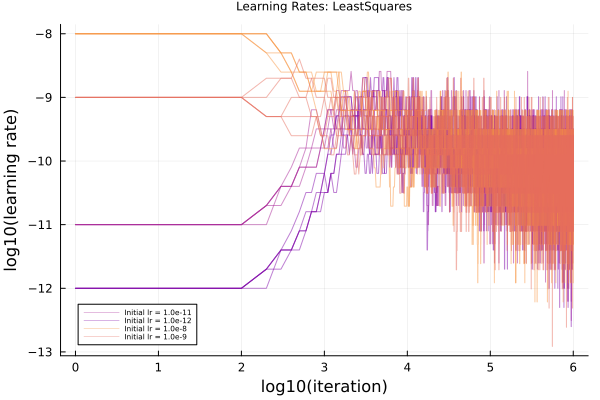}
  \end{subfigure}
  \begin{subfigure}{0.32\textwidth}
    \centering
    \includegraphics[width=\textwidth]{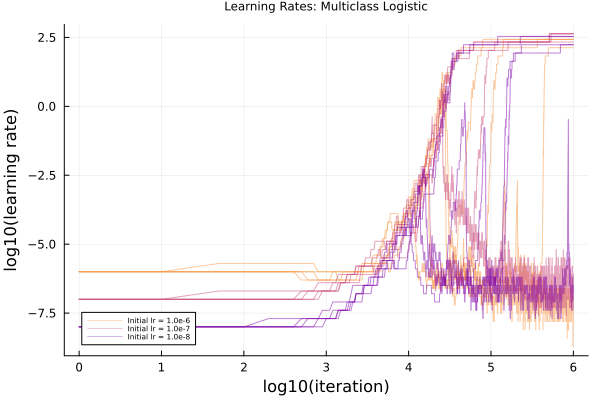}
  \end{subfigure}
  \caption{AutoSGD learning rates for each of the classical optimization problems. 
  In the second row, the three matrix factorization settings correspond to $k=1,4,10$, from left to right.}
  \label{fig:lr_experiments_classical}
\end{figure*}

\begin{figure*}
  \centering
  \begin{subfigure}{0.32\textwidth}
    \centering
    \includegraphics[width=\textwidth]{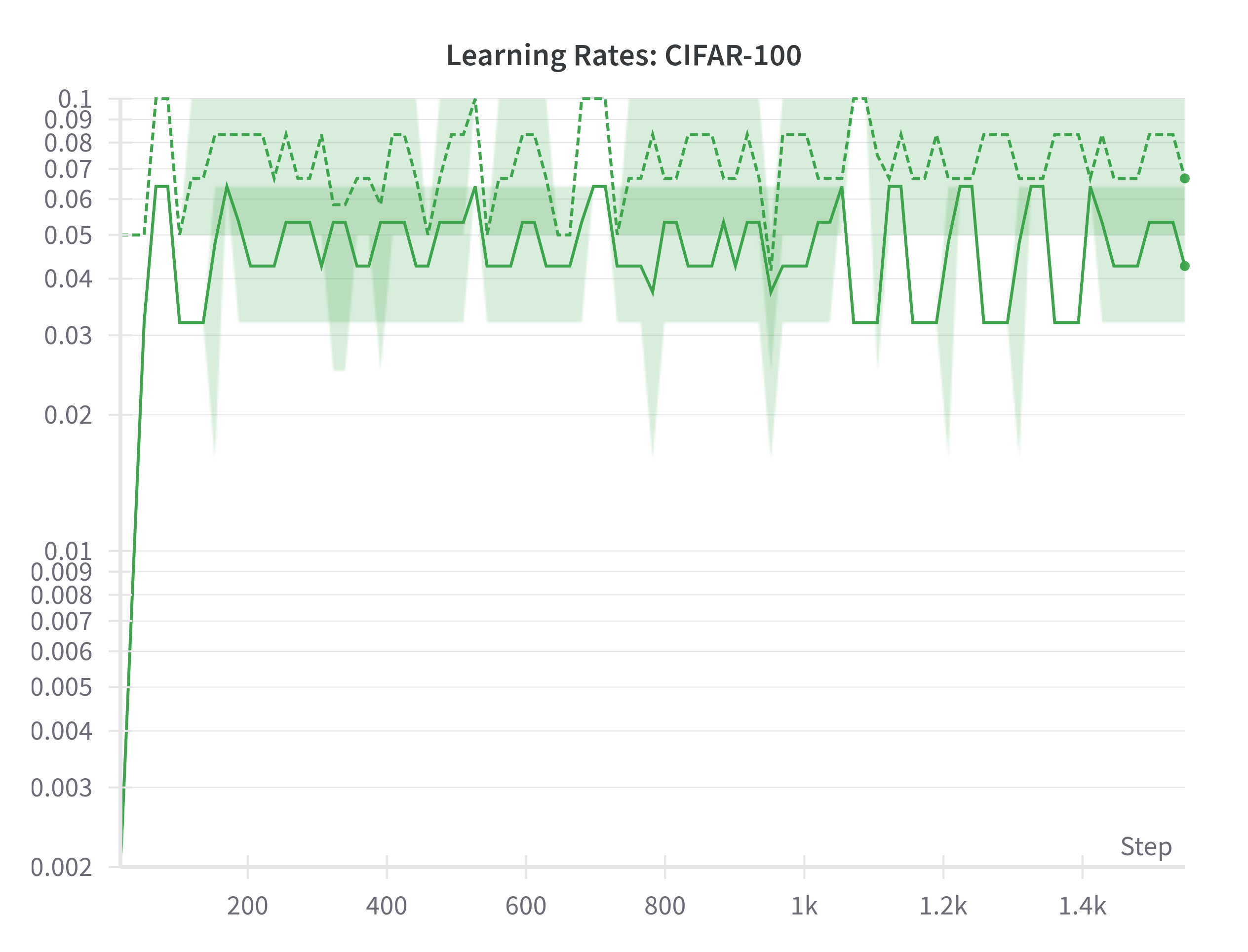}
  \end{subfigure}
  \begin{subfigure}{0.32\textwidth}
    \centering
    \includegraphics[width=\textwidth]{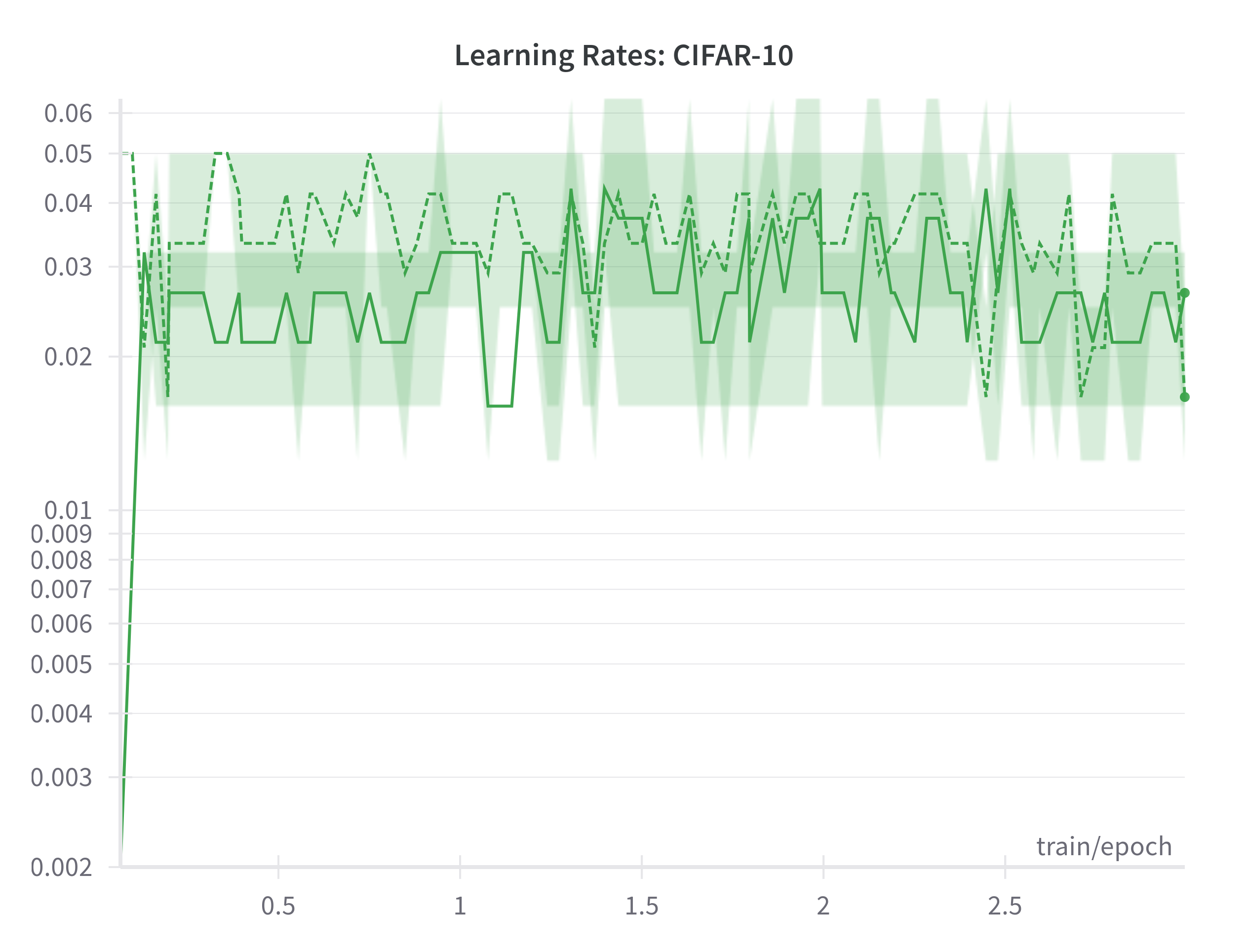}
  \end{subfigure} \\
  \begin{subfigure}{0.32\textwidth}
    \centering
    \includegraphics[width=\textwidth]{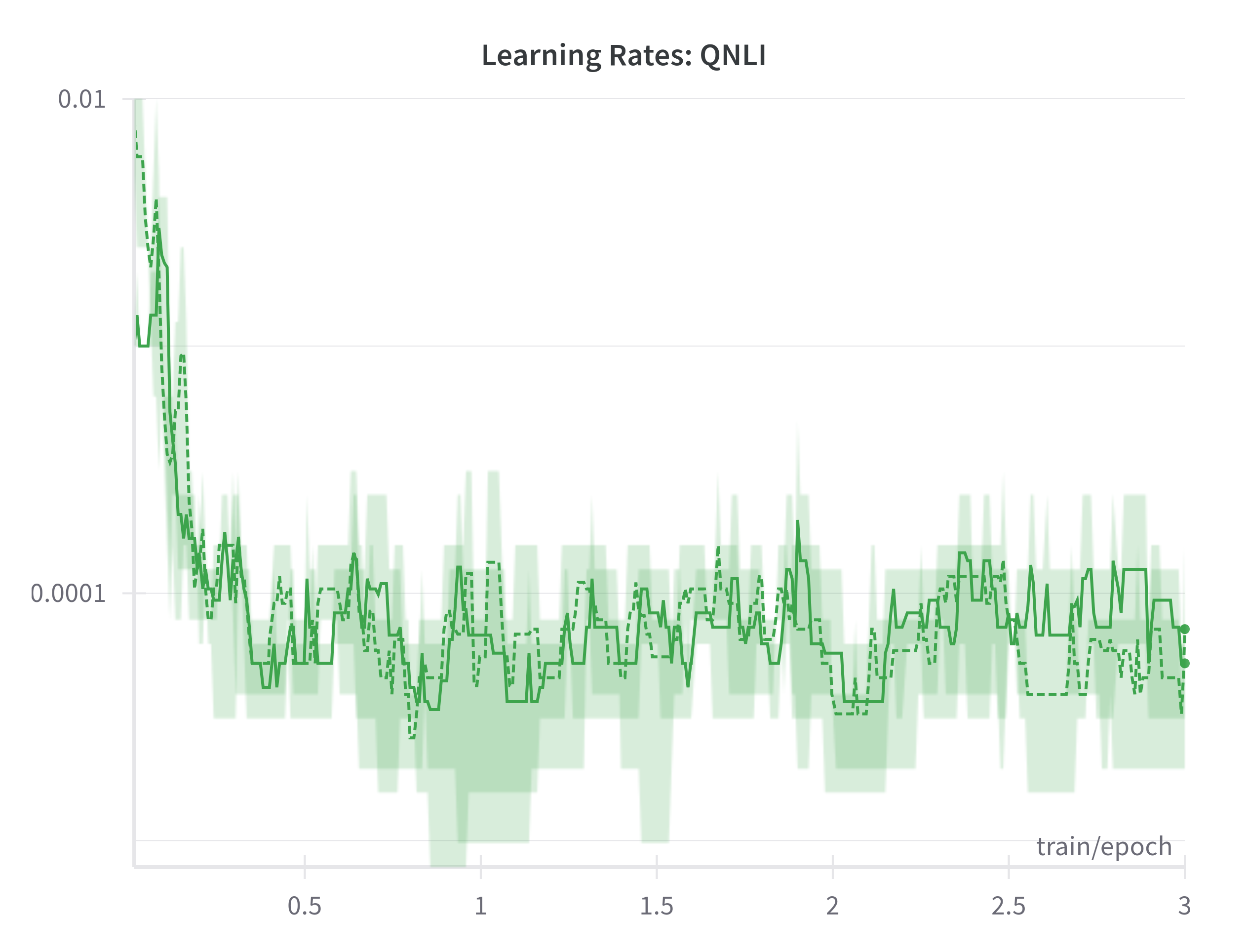}
  \end{subfigure} 
  \begin{subfigure}{0.32\textwidth}
    \centering
    \includegraphics[width=\textwidth]{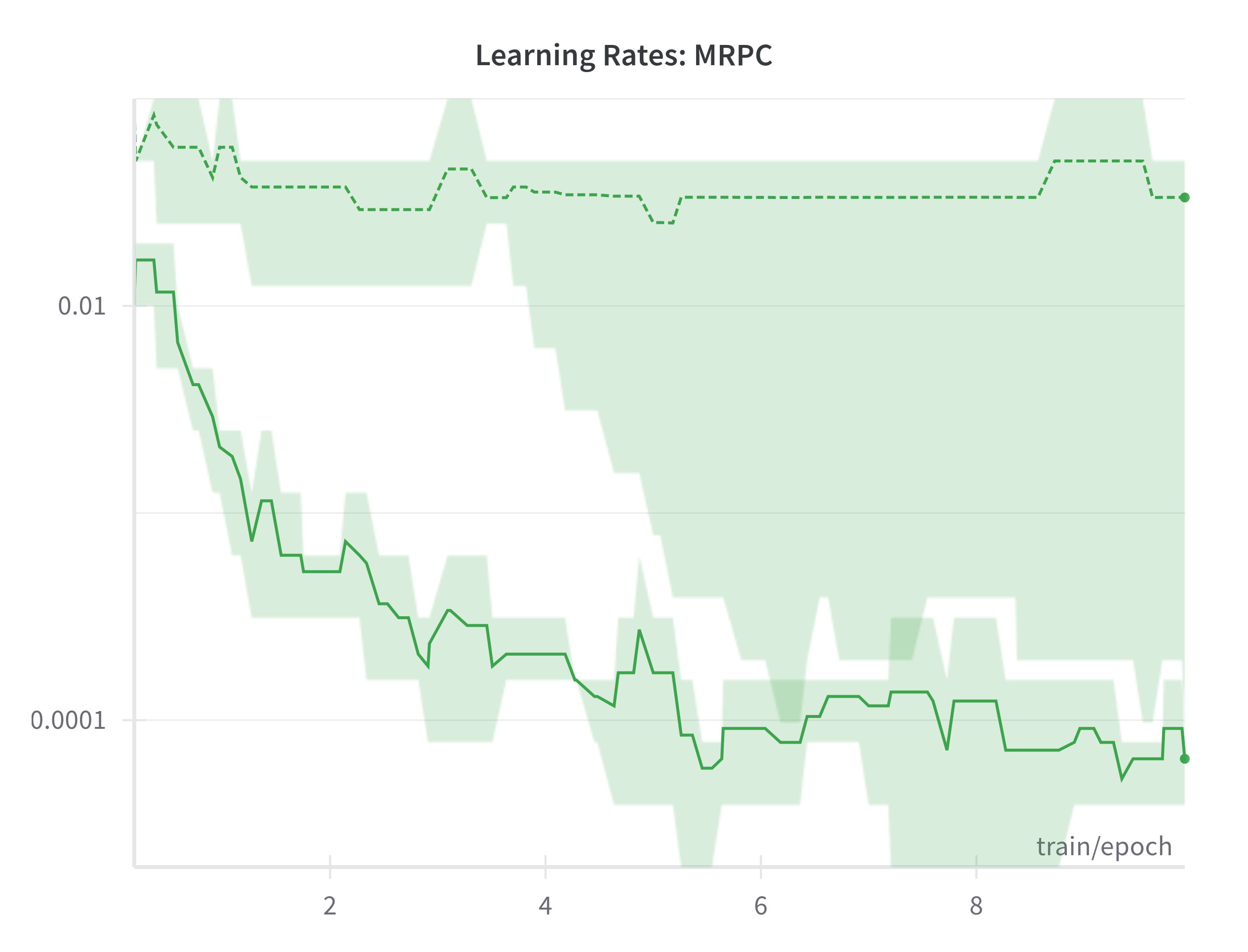}
  \end{subfigure} 
  \begin{subfigure}{0.32\textwidth}
    \centering
    \includegraphics[width=\textwidth]{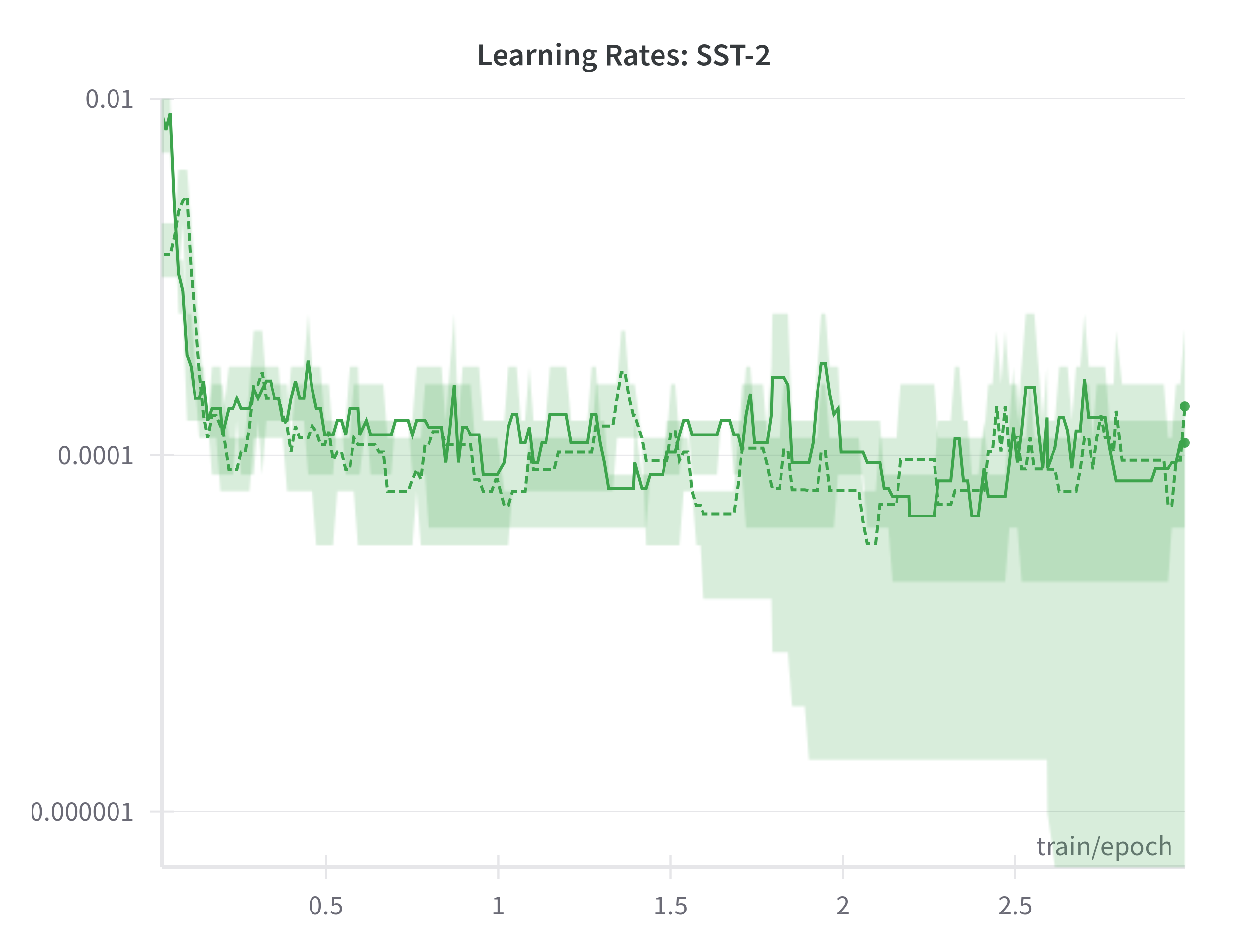}
  \end{subfigure} \\
  \caption{AutoSGD learning rates for each of the ML training optimization problems.}
  \label{fig:lr_experiments_ML}
\end{figure*}

\begin{figure*}[!t]
  \centering
  \begin{subfigure}{0.48\textwidth}
    \centering
    \includegraphics[width=\textwidth]{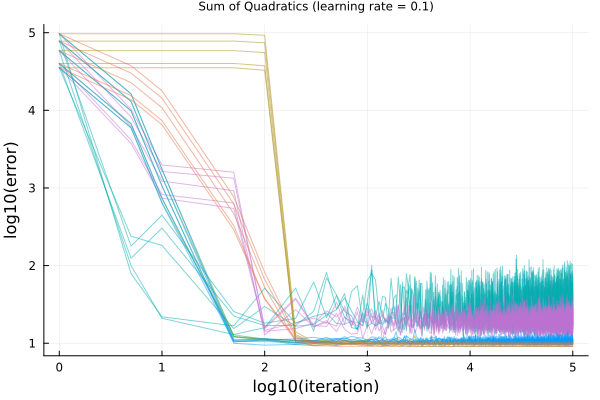}
  \end{subfigure}
  \begin{subfigure}{0.48\textwidth}
    \centering
    \includegraphics[width=\textwidth]{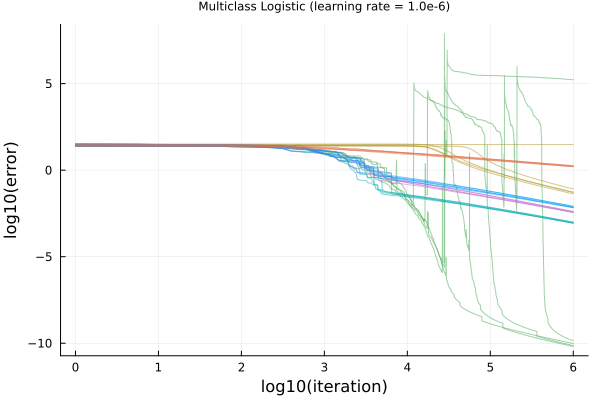}
  \end{subfigure}
  \caption{Classical optimization settings with learning rates chosen that are favourable 
  to the majority of optimizers (excluding AutoSGD and DoG). 
  Five different seeds are presented for each optimizer.
  The optimizers are labeled with the following colours:
  \textcolor{julia3}{AutoSGD}, 
  \textcolor{julia5}{DoG},
  \textcolor{julia4}{SFSGD},
  \textcolor{julia1}{SGD (constant)},
  \textcolor{julia2}{SGD (invsqrt)}, 
  \textcolor{julia6}{NMLS}. 
  In the sum of quadratics example, AutoSGD aligns closely with constant learning rate SGD 
  and so is less visible.}
  \label{fig:classical_tuned_additional}
\end{figure*}

\begin{figure*}[!t]
  \centering
  \begin{subfigure}{0.48\textwidth}
    \centering
    \includegraphics[width=\textwidth]{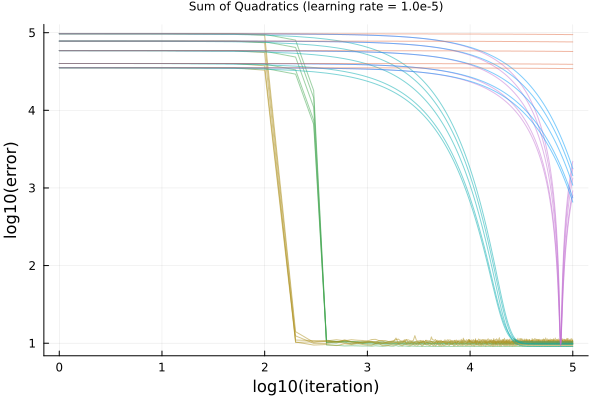}
  \end{subfigure}
  \begin{subfigure}{0.48\textwidth}
    \centering
    \includegraphics[width=\textwidth]{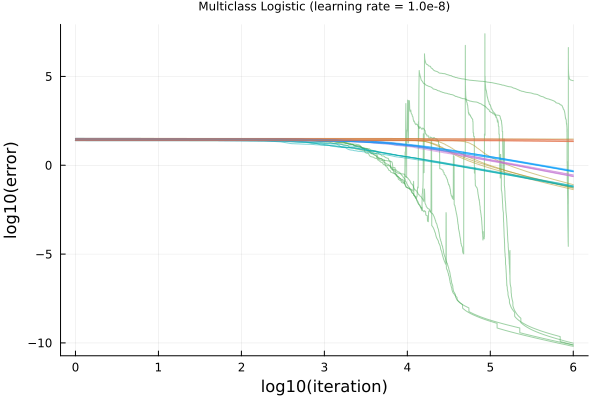}
  \end{subfigure}
  \caption{Classical optimization settings with initially small learning rates for the different optimizers 
  in order to assess robustness.}
  \label{fig:classical_small_additional}
\end{figure*}

\begin{figure*}
  \centering
  \begin{subfigure}{0.48\textwidth}
    \centering
    \includegraphics[width=\textwidth]{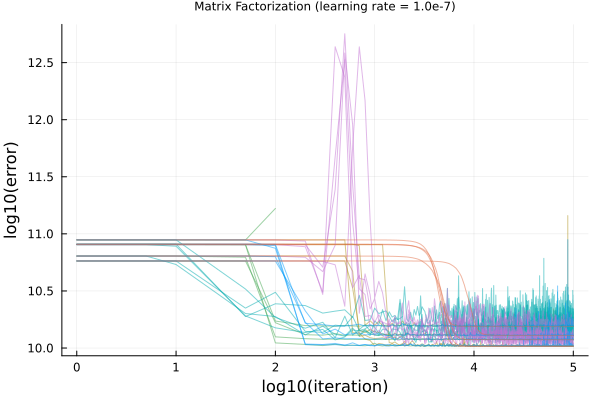}
  \end{subfigure}
  \begin{subfigure}{0.48\textwidth}
    \centering
    \includegraphics[width=\textwidth]{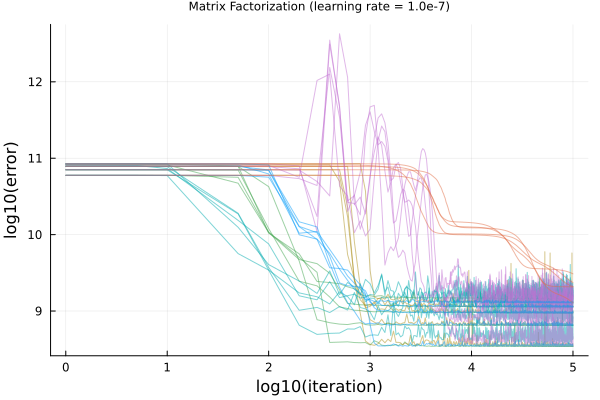}
  \end{subfigure} \\ 
  \begin{subfigure}{0.48\textwidth}
    \centering
    \includegraphics[width=\textwidth]{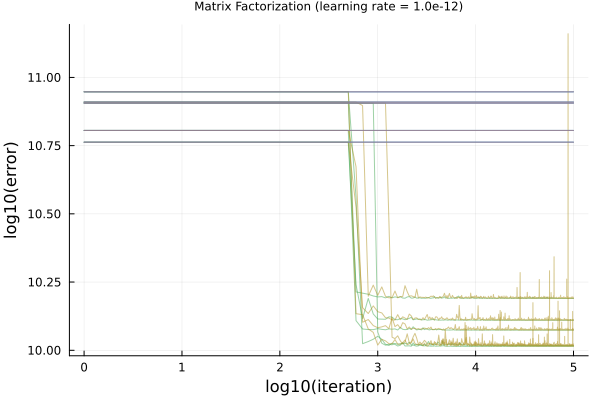}
  \end{subfigure} 
  \begin{subfigure}{0.48\textwidth}
    \centering
    \includegraphics[width=\textwidth]{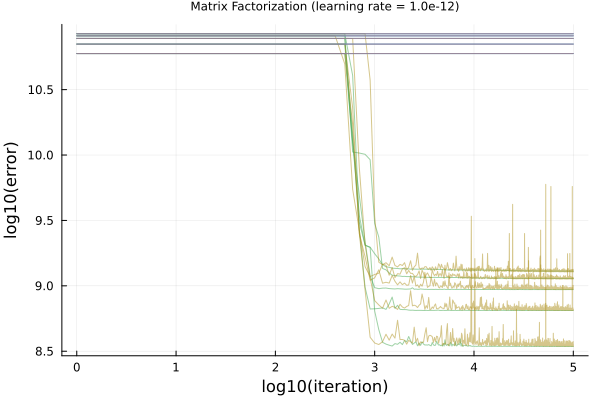}
  \end{subfigure} 
  \caption{Additional matrix factorization results with $k=1$ (left column) and $k=4$ 
  (right column). \textbf{Top:} Results with a common learning rate approximately optimal 
  for methods excluding AutoSGD and DoG. \textbf{Bottom:} Results with a learning rate 
  chosen to be small.}
  \label{fig:matrixfactor_additional}
\end{figure*}

\begin{figure*}[!t]
  \centering
  \begin{subfigure}{0.48\textwidth}
    \centering
    \includegraphics[width=\textwidth]{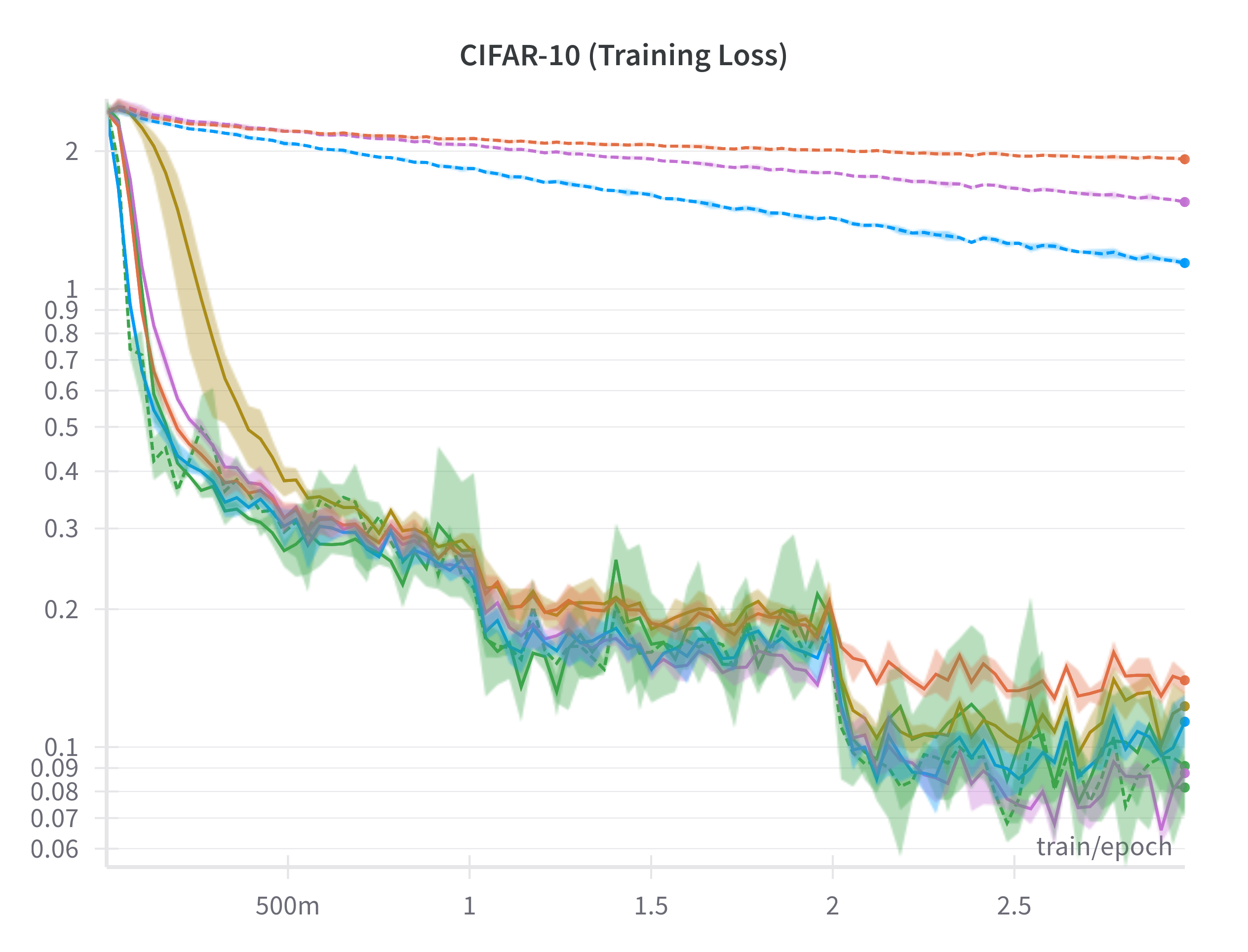}
  \end{subfigure}
  \begin{subfigure}{0.48\textwidth}
    \centering
    \includegraphics[width=\textwidth]{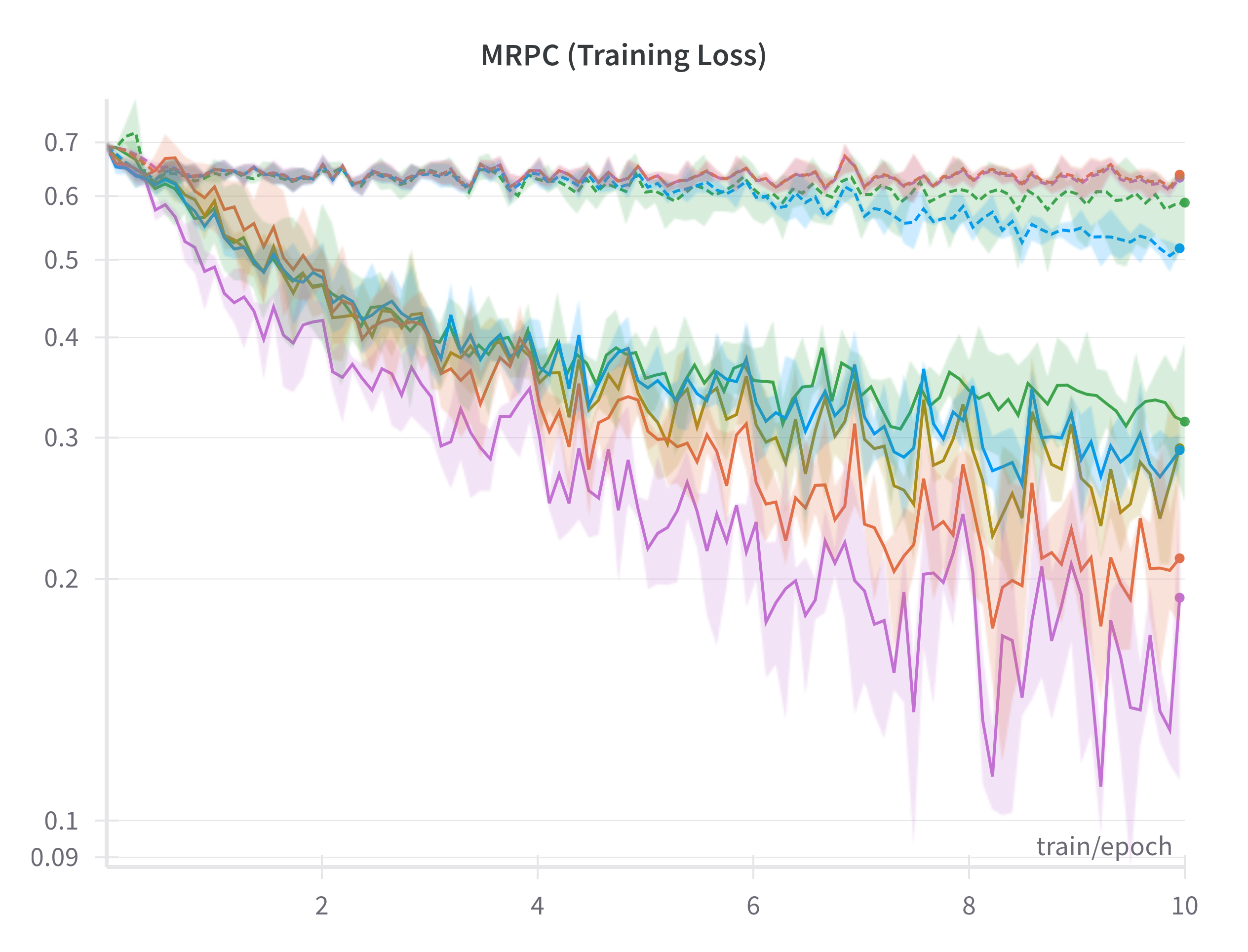}
  \end{subfigure}
  \caption{Training loss for various optimizers on additional ML data sets. Solid lines represent averages across seeds, 
  and shaded regions are the maximum and minimum across three seeds for a given optimizer. 
  For each data set and optimizer configuration, we present the best (solid line) 
  and worst (dashed line) learning rate 
  selections from a predefined grid. The colour scheme is the same as 
  in \cref{fig:classical_tuned}: 
  \textcolor{julia3}{AutoSGD}, 
  \textcolor{julia5}{DoG},
  \textcolor{julia4}{SFSGD},
  \textcolor{julia1}{SGD (constant)},
  \textcolor{julia2}{SGD (invsqrt)}.}
  \label{fig:ML_training_additional}
\end{figure*}

\begin{figure*}[!t]
  \centering
  \begin{subfigure}{0.48\textwidth}
    \centering
    \includegraphics[width=\textwidth]{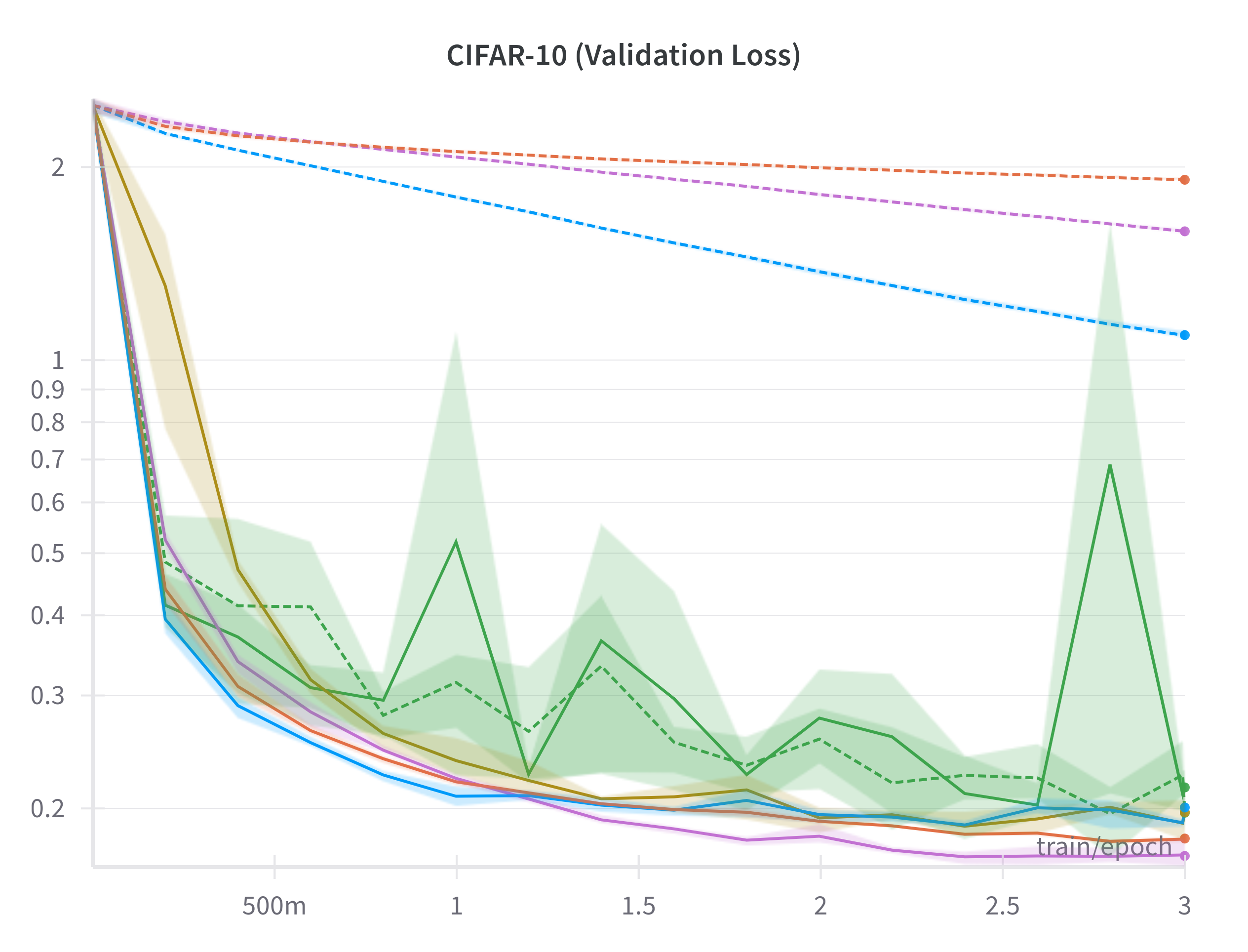}
  \end{subfigure} 
  \begin{subfigure}{0.48\textwidth}
    \centering
    \includegraphics[width=\textwidth]{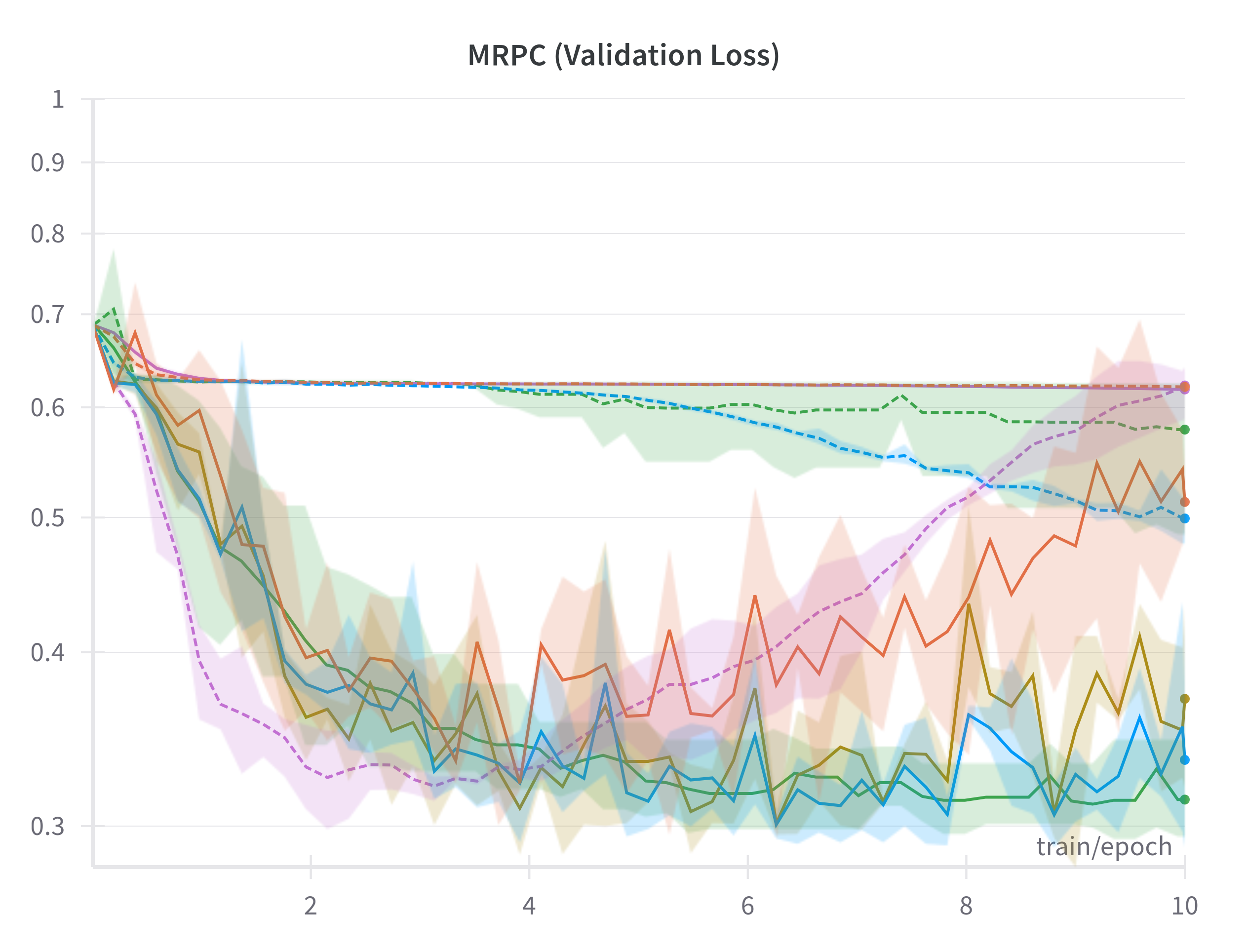}
  \end{subfigure}
  \caption{Validation loss for various optimizers on additional ML data sets.}
  \label{fig:ML_validation_additional}
\end{figure*}

\cref{fig:lr_experiments_classical,fig:lr_experiments_ML} present the 
selected AutoSGD learning rates results for each of the classical optimization and 
ML training experiments.
Interestingly, in \cref{fig:lr_experiments_classical} for the logistic regression 
examples, we find that AutoSGD tends to select larger learning rates as the iterations 
progress. This seems to be in line with recent findings for logistic regression \cite{axiotis2023logistic} 
and could be investigated further.
\cref{fig:classical_tuned_additional,fig:classical_small_additional} present 
experimental results for the classical optimization setting not presented 
in the main text.
\cref{fig:matrixfactor_additional} also presents additional results for the matrix factorization 
optimization problem with $k=1, 4$ as only $k=10$ was reported in the main text.
\cref{fig:ML_training_additional,fig:ML_validation_additional} present experimental 
results for the ML experiments that were not included in the main text.

\subsection{AutoGD experiments} 
\label{sec:autogd_experiments}

\begin{figure*}
  \centering
  \begin{subfigure}{0.32\textwidth}
    \centering
    \includegraphics[width=\textwidth]{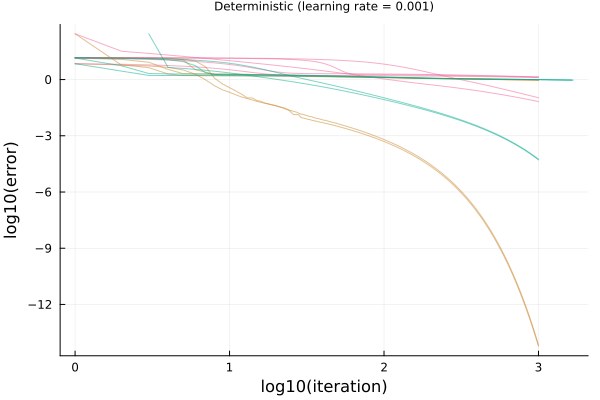}
  \end{subfigure}
  \begin{subfigure}{0.32\textwidth}
    \centering
    \includegraphics[width=\textwidth]{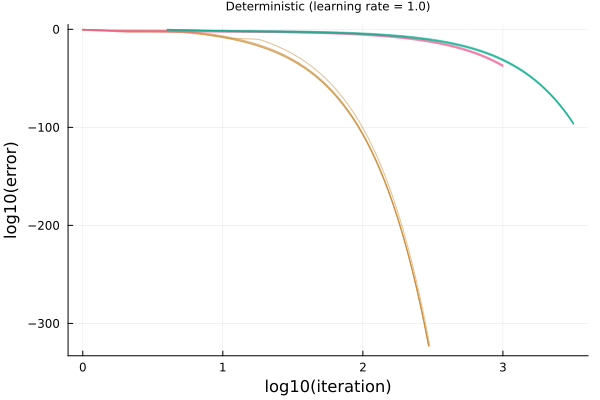}
  \end{subfigure}  
  \begin{subfigure}{0.32\textwidth}
    \centering
    \includegraphics[width=\textwidth]{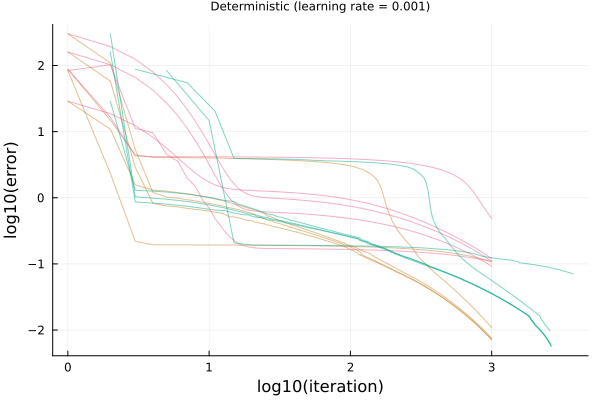}
  \end{subfigure} 
  \begin{subfigure}{0.32\textwidth}
    \centering
    \includegraphics[width=\textwidth]{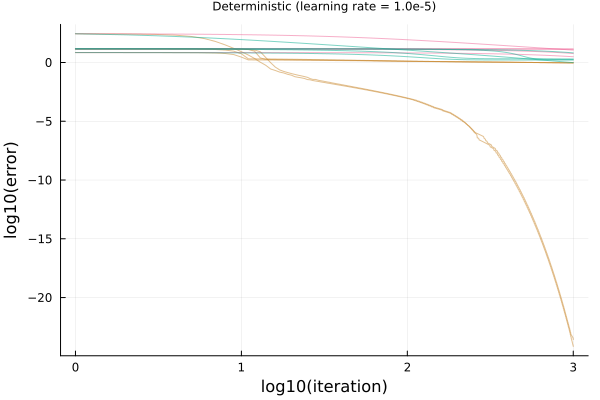}
  \end{subfigure} 
  \begin{subfigure}{0.32\textwidth}
    \centering
    \includegraphics[width=\textwidth]{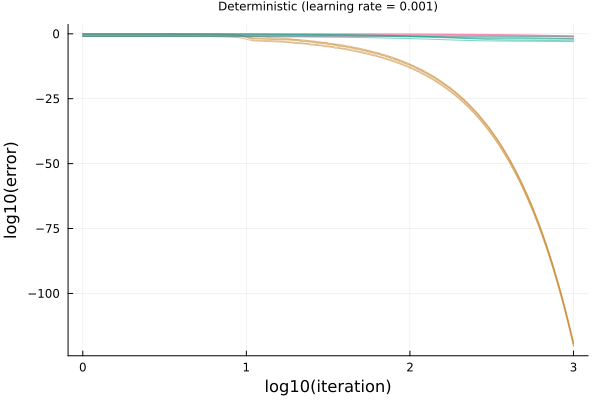}
  \end{subfigure}
  \begin{subfigure}{0.32\textwidth}
    \centering
    \includegraphics[width=\textwidth]{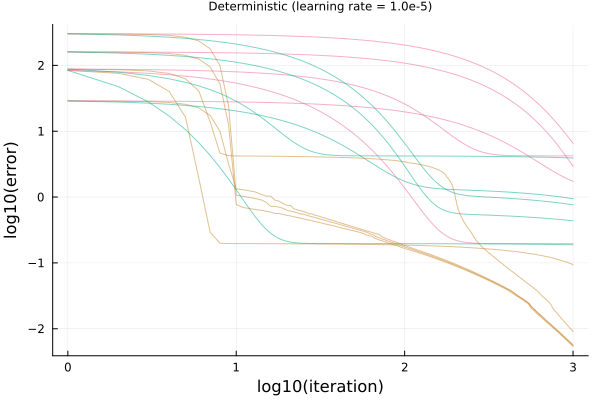}
  \end{subfigure}
  \caption{Simulation results for the Beale (left column), Matyas (middle column), 
  and Rosenbrock (right column) objectives with \textcolor{julia7}{GD}, 
  \textcolor{julia8}{AutoGD}, and \textcolor{julia9}{LineSearch}.
  For the line search algorithm, the x-axis represents the cumulative 
  number of backtracks performed.
  \textbf{Top:} Learning rates selected for good performance for GD and line search. 
  \textbf{Bottom:} Small initial learning rates.}
  \label{fig:autogd_additional_1}
\end{figure*}

\begin{figure*}
  \centering
  \begin{subfigure}{0.48\textwidth}
    \centering
    \includegraphics[width=\textwidth]{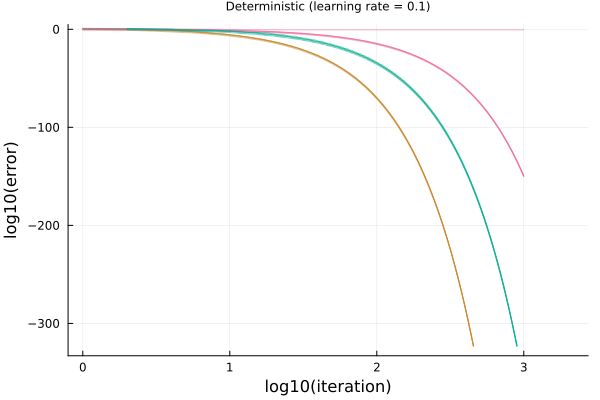}
  \end{subfigure}
  \begin{subfigure}{0.48\textwidth}
    \centering
    \includegraphics[width=\textwidth]{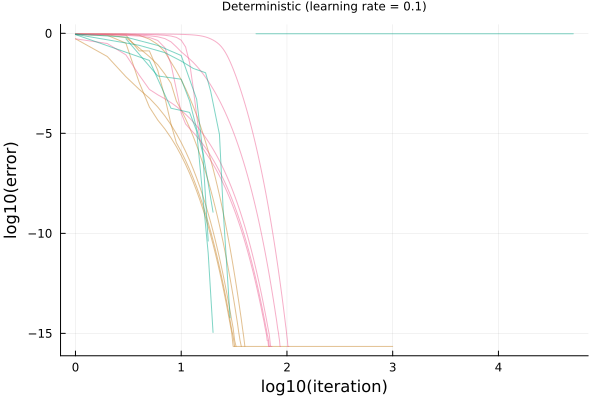}
  \end{subfigure} \\
  \begin{subfigure}{0.48\textwidth}
    \centering
    \includegraphics[width=\textwidth]{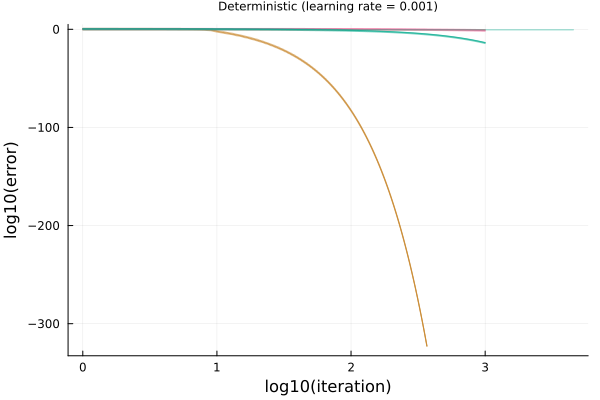}
  \end{subfigure} 
  \begin{subfigure}{0.48\textwidth}
    \centering
    \includegraphics[width=\textwidth]{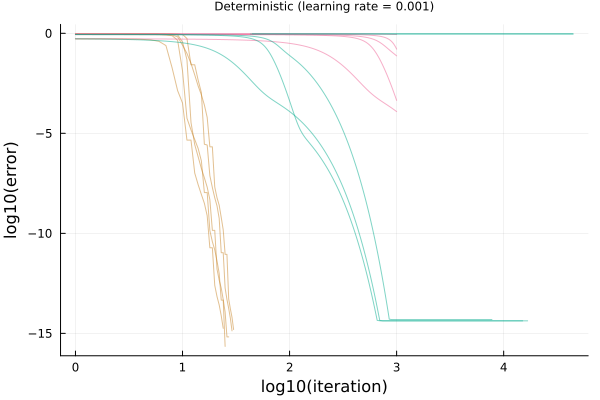}
  \end{subfigure} 
  \caption{Simulation results for the three-hump camel (left column) and 
  valley (right column) objectives with \textcolor{julia7}{GD}, 
  \textcolor{julia8}{AutoGD}, and \textcolor{julia9}{LineSearch}.
  For the line search algorithm, the x-axis represents the cumulative 
  number of backtracks performed.
  \textbf{Top:} Learning rates selected for good performance for GD and line search. 
  \textbf{Bottom:} Small initial learning rates.}
  \label{fig:autogd_additional_2}
\end{figure*}

\begin{figure*}
  \centering
  \begin{subfigure}{0.32\textwidth}
    \centering
    \includegraphics[width=\textwidth]{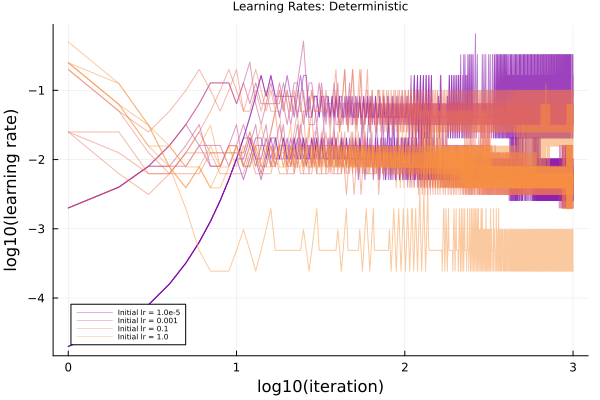}
  \end{subfigure}
  \begin{subfigure}{0.32\textwidth}
    \centering
    \includegraphics[width=\textwidth]{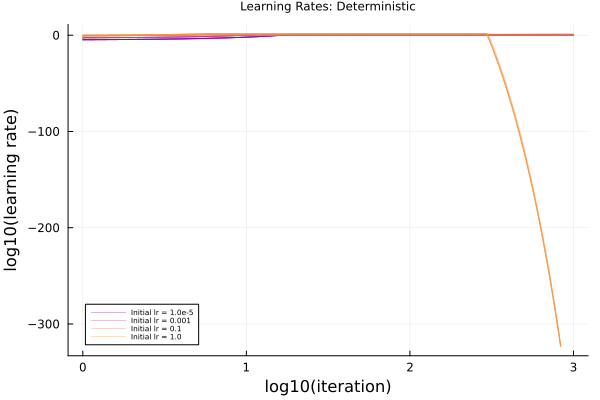}
  \end{subfigure} \\
  \begin{subfigure}{0.32\textwidth}
    \centering
    \includegraphics[width=\textwidth]{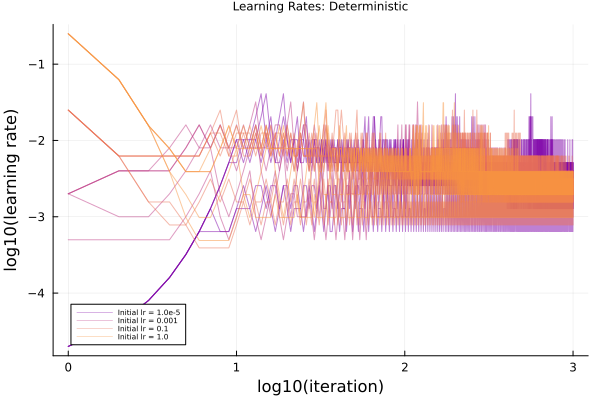}
  \end{subfigure} 
  \begin{subfigure}{0.32\textwidth}
    \centering
    \includegraphics[width=\textwidth]{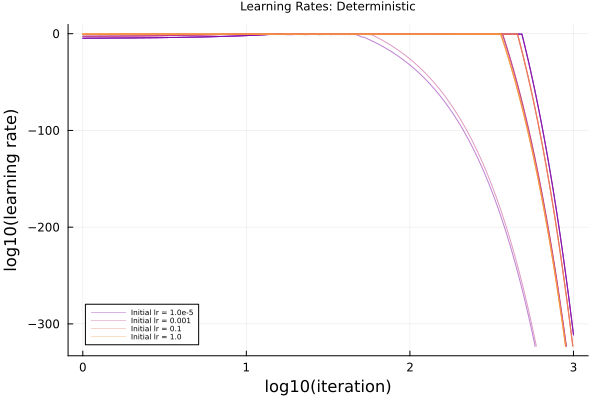}
  \end{subfigure} 
  \begin{subfigure}{0.32\textwidth}
    \centering
    \includegraphics[width=\textwidth]{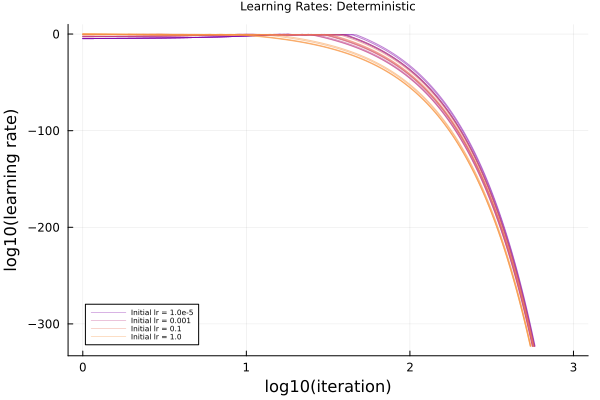}
  \end{subfigure} 
  \caption{AutoGD learning rates for each of the deterministic optimization problems.}
  \label{fig:autogd_stepsizes}
\end{figure*}

\cref{fig:autogd_additional_1,fig:autogd_additional_2} present results for 
AutoGD on various targets, including: 
the Beale, Matyas, Rosenbrock, three-hump camel, and valley functions \cite{surjanovic2013virtual}.
All functions are two-dimensional and specified below. 
\cref{fig:autogd_stepsizes} presents the chosen AutoGD learning rates for each of the 
five deterministic objectives.

The Beal function is given by 
\[
  f(x, y) = (1.5 - x + xy)^2 + (2.25 - x + xy^2)^2 + (2.625 - x + xy^3)^2.
\]
The Matyas function is given by 
\[
  f(x, y) = 0.26 (x^2 + y^2) - 0.48 xy.
\]
The Rosenbrock function is given by 
\[
  f(x, y) = (1-x)^2 + 100(y-x^2)^2.
\]
The three-hump camel function is given by 
\[
  f(x, y) = 2x^2 - 1.05x^4 + \frac{x^6}{6} + xy + y^2.
\]
The valley function is given by 
\[
  f(x, y) = 1 - 1/(1+x^2 + 4y^2).
\]

\clearpage

\end{document}